\documentclass[11pt]{article} 
\usepackage{amsmath,amssymb,mathrsfs,amsthm}
\usepackage{graphicx}
\usepackage{epstopdf}
\usepackage{stmaryrd}
\usepackage{wrapfig}
\usepackage{bm,cases,multirow,booktabs,enumitem}
\usepackage{tikz}
\usetikzlibrary{decorations.markings}
\usepackage{quiver}
\newtheorem{remark}{Remark}
\newtheorem{example}{Example}
\newtheorem{defi}{Definition}
\newtheorem{lem}{Lemma}
\newtheorem{theorem}{Theorem}
\usepackage{natbib}
\newtheorem{assumption}{Assumption}
\newtheorem{cor}{Corollary}
\newtheorem{prop}{Proposition}
\newcommand{\RR}{\mathbb{R}}
\newcommand{\EE}{\mathbb{E}}
\newcommand{\lp}{\left(}
\newcommand{\rp}{\right)}

\usepackage{authblk}

\usepackage{xcolor}
\usepackage[backref]{hyperref} 

\hypersetup{
    colorlinks=false,       
    linkbordercolor={1 0 0}, 
    pdfborder={0 0 1},      
    citebordercolor={0 1 0}, 
    urlbordercolor={0 1 1}   
}

\allowdisplaybreaks
\title{Diffusion Score Estimation with Shallow Neural Networks: Regularity, Approximation, and Finite-Sample Rates}

\author[1]{Fan Liu\thanks{fliu370@gatech.edu}} 
\author[2]{Yahong Yang}
\author[1]{Wei Zhu}
\affil[1]{School of Mathematics, Georgia Institute of Technology, USA}
\affil[2]{Department of Mathematics and Statistics, Binghamton University, USA}
\date{October 19, 2026}

\begin{document}

\maketitle
\begin{abstract}
We study the sample complexity of estimating Ornstein--Uhlenbeck diffusion
scores in $\mathbb R^d$ using denoising score matching (DSM) with
finite-width shallow ReLU networks, and its implications for
reverse-process sampling. Under weak differentiability and
polynomial-growth conditions on normalized derivatives of the initial
density, we combine complementary estimates from the Gaussian transition
kernel and the initial density to obtain explicit Sobolev bounds that
are uniform in time on each fixed ball. Combining these bounds with
existing shallow-network approximation theory gives explicit control
of the approximation error and a sufficient path-norm budget. We then
use truncation and localized Rademacher complexity to establish
high-probability risk bounds for path-norm-constrained empirical DSM
minimizers with unbounded inputs and labels. Under additional
bounded-density and sub-Gaussian assumptions, suitable choices of
network width and path-norm budget yield squared $L^2(P_t)$ score risk
$\widetilde O(n^{-(d+3)/(2d+3)})$ at each fixed time $t>0$, where $n$
is the number of independent training samples and $P_t$ is the forward
marginal. The nonasymptotic bounds retain explicit dependence on
diffusion time and the regularity and tail parameters of the initial
distribution. Aggregating the score errors over a time grid and applying
existing sampling theory together with an early-stopping bound yields
a high-probability total-variation error
$\widetilde O(n^{-(d+3)/(4d+6)})$ between the generated and original data
distributions under suitable sampling parameters. Here $\widetilde O$
suppresses logarithmic factors for fixed dimension, confidence level,
and initial distribution.
\end{abstract}

\section{Introduction}

Diffusion models generate samples by reversing a stochastic process that gradually transforms a data
distribution into a tractable reference distribution~\cite{ho2020denoising,song2021scorebased}. We
consider the Ornstein--Uhlenbeck (OU) forward diffusion, a particular variance-preserving
process~\cite{song2021scorebased},
\begin{equation}
\label{eq:forward_diffusion_intro}
    \mathrm dX_t=-X_t\,\mathrm dt+\sqrt{2}\,\mathrm dW_t,
    \qquad X_0\sim P_0.
\end{equation}
It admits the representation $X_t=e^{-t}X_0+\sigma_t Z$, where $\sigma_t^2=1-e^{-2t}$ and
$Z\sim\mathcal N(0,I_d)$ is independent of $X_0$. Thus the distribution $P_t$ of $X_t$ approaches
the standard Gaussian distribution as $t$ increases. For $t>0$, let $p_t$ be the density of $P_t$
and $s_t=\nabla\log p_t$ its \emph{score function}.

The reverse-time process makes the role of the score explicit. If the forward diffusion is run until
time $T$, the exact reverse process, initialized from $P_T$,
satisfies~\cite{anderson1982reverse,song2021scorebased}
\begin{equation}
\label{eq:reverse_diffusion_intro}
    \mathrm d\bar X_u
    =\bigl[\bar X_u+2s_{T-u}(\bar X_u)\bigr] \,\mathrm du
    +\sqrt{2}\,\mathrm d\bar W_u,
    \qquad \bar X_0\sim P_T.
\end{equation}
For any $\gamma\in(0,T)$, its law at reverse time $T-\gamma$ is $P_\gamma$, converging weakly to
$P_0$ as $\gamma\downarrow0$. In practice, one starts from an approximately matching Gaussian
distribution, discretizes the reverse dynamics, and replaces the unknown score $s_t$ with an
estimate learned from data. The accuracy of the generated distribution therefore depends critically on how accurately the score can be estimated from finitely many training samples.

In this work, we study the sample complexity of score estimation using
\emph{denoising score matching} (DSM) \cite{Vin11}.
Given $n$ independent samples $X_{0,i}\sim P_0$, we generate
Gaussian perturbations $Z_i\sim\mathcal N(0,I_d)$ and construct
the noisy observations
$X_{t,i}=e^{-t}X_{0,i}+\sigma_t Z_i$.
For a candidate score function $f:\mathbb R^d\to\mathbb R^d$,
the empirical DSM loss is
\begin{equation*}
    \widehat{\mathcal J}_t(f)
    :=\frac{1}{n}\sum_{i=1}^n
    \left\|f(X_{t,i})+\frac{Z_i}{\sigma_t}\right\|_2^2.
\end{equation*}
By Tweedie's identity,
\begin{equation*}
    \mathbb E\left[-\frac{Z}{\sigma_t}\,\middle|\,X_t\right]
    =s_t(X_t),
\end{equation*}
the true score minimizes the population DSM loss
$\mathcal J_t(f):=\mathbb E\|f(X_t)+Z/\sigma_t\|_2^2$.
Moreover, its excess loss satisfies
\begin{equation*}
    \mathcal J_t(f)-\mathcal J_t(s_t)
    =\mathbb E_{X_t\sim P_t}
    \|f(X_t)-s_t(X_t)\|_2^2.
\end{equation*}
Thus, minimizing the DSM loss provides a regression-based
approach to score estimation, with excess population loss
equal to the squared score-estimation error.
We investigate the finite-sample accuracy of empirical DSM
minimizers over finite-width shallow ReLU networks with
path-norm constraints, as defined in Section~\ref{sec:setup}.

Our goal is twofold: \emph{first}, to quantify the sample size needed to estimate $s_t$
accurately, including the effects of diffusion time and properties of the data distribution $P_0$; and \emph{second}, to
translate these score-estimation bounds into an error guarantee for the distribution generated by
a discretized reverse diffusion. In particular, we seek explicit bounds that connect the number of
training data to the discrepancy between the generated and target distributions.

\subsection{Main Results and Contributions}

\paragraph{Score regularity and shallow-network approximation.}
The first challenge is to quantify how accurately the diffusion score can be approximated by shallow ReLU networks. The existing result on shallow ReLU approximation of Sobolev functions
~\cite{liu2025integral} establishes a squared $L^2$ approximation rate of order $m^{-(d+3)/d}$ for width-$m$ networks, provided that the target function belongs to $H^{(d+3)/2}(\Omega_R)$ and the network path-norm budget is chosen accordingly. Here, $\Omega_R$ denotes the ball of radius $R$ centered at the origin. To apply this result to diffusion scores, we need explicit bounds on their Sobolev norms that quantify the dependence on diffusion time $t$, spatial radius $R$, ambient dimension $d$, and the regularity and tail properties of the initial distribution. Although Gaussian smoothing induced by the forward diffusion~\eqref{eq:forward_diffusion_intro} makes $s_t$ smooth for every $t>0$, smoothness alone does not provide such quantitative control, particularly a bound that remains uniform over diffusion time.

To address this challenge, we derive two complementary bounds on the Sobolev norm of $s_t$ over bounded balls. The first, a \emph{kernel-induced bound}, is obtained by differentiating the Gaussian transition kernel and applies to arbitrary initial distributions, but deteriorates as $t\downarrow0$. The second, a \emph{data-induced bound}, is obtained by transferring derivatives to the initial density. Under suitable weak differentiability and polynomial-growth assumptions on the normalized density derivatives, this bound remains controlled near $t=0$, although it deteriorates at larger times. Taking the minimum of these two bounds yields a Sobolev envelope that remains uniformly bounded for all $t>0$ on each fixed ball. This time-uniform regularity estimate, combined with the shallow-ReLU approximation theorem, provides explicit control of the score approximation error and a sufficient path-norm budget that can be chosen independently of diffusion time. The precise assumptions and estimates are presented in Section~\ref{sec:reg}.

\paragraph{Finite-sample score estimation.}

The second challenge is statistical. Both the noisy inputs $X_t$ and DSM labels
$Y_t=-Z/\sigma_t$ are unbounded, and the regression residual
$Y_t-s_t(X_t)$ need not be independent of $X_t$. To address these
difficulties, we combine truncation with localized Rademacher complexity
analysis, while accounting for the bias introduced by jointly truncating
the inputs and labels. This leads to a high-probability bound on the
squared score-estimation error consisting of an approximation term of
order $m^{-(d+3)/d}$, an estimation term of order $m/n$, and explicit
tail and confidence terms, where $m$ is the network width per output
coordinate and $n$ is the number of independent training samples
from $P_0$. The stated orders suppress logarithmic and
distribution-dependent factors.

Furthermore, under the regularity and normalized-derivative growth
conditions introduced in Section~\ref{sec:reg}, together with boundedness of the initial density and a sub-Gaussian tail assumption, we show that the tail, truncation-bias, and confidence terms can be controlled without affecting the leading convergence rate. Balancing the approximation and estimation terms then leads to a network width $m=\widetilde O(n^{d/(2d+3)})$ and yields, for each fixed $t>0$ and all sufficiently large $n$, the high-probability rate
\begin{equation*}
    \EE_{X\sim P_t}\!\left[
    \|\widehat s_{t,m}(X)-s_t(X)\|_2^2
    \right]
    =\widetilde O\!\left(n^{-\beta_d}\right),
    \qquad
    \beta_d:=\frac{d+3}{2d+3}.
\end{equation*}
Here $\widetilde O$ suppresses logarithmic factors, with constants
depending on the fixed dimension, diffusion time, confidence level,
and initial distribution. In particular, for fixed values of these
parameters, a sample size of order
$\widetilde O(\varepsilon^{-(2d+3)/(d+3)})$ is sufficient to achieve
a squared score-estimation error of at most $\varepsilon$.
Importantly, our nonasymptotic bounds make the dependence on diffusion
time and the regularity and tail parameters of the initial distribution
explicit. This quantitative dependence is essential for controlling
the accumulated score-estimation error in the subsequent reverse-sampling
analysis. The detailed results are given in
Section~\ref{sec:stats shallow}.

\paragraph{Implications for reverse sampling.}
By combining score estimates at a grid of diffusion times with an existing score-to-sampling
theorem~\cite{benton2024nearly}, we bound the discrepancy between the generated distribution and an
early-stopped target distribution. Controlling the additional early-stopping error gives, with high
probability and suitable choices of the sampling parameters,
\begin{equation*}
    \operatorname{TV}(p_0,\widehat p_{\mathrm{out}})
    =\widetilde O\!\left(n^{-\frac{d+3}{4d+6}}\right),
\end{equation*}
where $\widehat p_{\mathrm{out}}$ is the output density of the discretized reverse sampler. Thus the
fixed-time statistical guarantee leads to an end-to-end bound in total variation
(Section~\ref{sec:sampling}).

Notably, the score-estimation and sampling rates have exponents that approach $1/2$ and $1/4$, respectively, as the ambient dimension $d$ increases. Thus, for distributions satisfying the stated regularity and tail assumptions, the convergence-rate exponents remain bounded away from zero even in high dimensions. This demonstrates that shallow ReLU networks trained by DSM can mitigate the curse of dimensionality in terms of sample-size convergence rates.

\subsection{Related Work}

\paragraph{Regularity of Diffusion Scores.}
De Bortoli et al.~\cite{de2021diffusion} establish time-uniform polynomial-growth bounds for higher-order derivatives of OU log-densities under assumptions including smoothness and a dissipativity condition on the initial score. Their analysis combines derivative estimates suited to different diffusion-time regimes. St{\'e}phanovitch \cite{stephanovitch2025regularity} studies Lipschitz and higher-order H{\"o}lder regularity of diffusion scores, while St{\'e}phanovitch et al.~\cite{stephanovitch2025generalization} connect score regularity with neural approximation, statistical estimation, and sampling. Our work develops explicit local Sobolev bounds under weak differentiability and polynomial-growth conditions on normalized density derivatives, without requiring dissipativity. These bounds quantify the dependence on diffusion time, spatial radius, and properties of the initial distribution, and provide the regularity control needed for shallow-ReLU approximation.

\paragraph{Shallow-Network Approximation and Statistical Learning.}
Approximation and complexity theory for shallow neural networks under norm and sparsity constraints has been developed in~\cite{barron1993universal,bach2017breaking,ma2022barron,klusowski2018approximation,siegel2024sharp}. Our approximation analysis uses the Sobolev-to-shallow-ReLU theorem of Liu et al.~\cite{liu2025integral}, together with our quantitative regularity bounds for diffusion scores, to control the approximation error and sufficient path-norm budget.

Statistical analyses of two-layer networks include~\cite{e2019apriori,parhi2022near,yang2024nonparametric,li2026shallow}, with recent shallow-network regression results achieving rates of order $n^{-(d+3)/(2d+3)}$, up to logarithmic factors, for suitable function classes. These results are established under bounded-design or other regression assumptions that do not directly cover DSM. In the DSM setting, both inputs and labels are unbounded, and the regression residual generally depends on the input.

Our statistical analysis builds on localized Rademacher complexity and excess-risk theory~\cite{bartlett2005local,koltchinskii2006local}. Related work has studied square-loss learning with unbounded or heavy-tailed observations~\cite{mendelson2014learning,liang2015learning,kuchibhotla2022least}. We combine localization with tail control for DSM, accounting in particular for the bias introduced by jointly truncating the inputs and labels, which need not preserve their conditional-mean identity.

\paragraph{Neural Score Estimation for Diffusion Models.}
Neural score-estimation guarantees have been developed  using empirical-process method~\cite{block2020generative}, under low-dimensional structure~\cite{chen2023score}, and under smoothness or other structural assumptions on the data distribution~\cite{oko2023diffusion,cole2024score,mei2025deep}. Fu and Lee~\cite{fu2026approximation} study approximation and score estimation using deep ReLU networks under sub-Gaussian assumptions.

Other work considers alternative score-accuracy criteria~\cite{gupta2024improved}, time-conditioned estimation~\cite{kumar2025dimensionfree}, optimization-aware sample complexity~\cite{gaur2025improved}, and random-feature models~\cite{george2026denoising}. Particularly relevant is Han et al.~\cite{han2024neural}, who analyze time-conditioned DSM using truncation and localized complexity arguments. In contrast, we study empirical DSM minimizers over finite-width, path-norm-constrained shallow ReLU networks at fixed diffusion times, and derive global $L^2(P_t)$ risk bounds that track the regularity and tail parameters of the initial distribution. The explicit time dependence also supports our subsequent sampling analysis.

\paragraph{Statistical Benchmarks and Sampling Guarantees.}
Kernel-based and empirical-Bayes approaches provide complementary statistical benchmarks for score estimation and distribution recovery~\cite{zhang2024minimax,wibisono2024optimal,dou2024optimal}. Our analysis instead focuses on the finite-sample performance of a specified shallow-network DSM estimator, including the capacity needed to approximate the score.

A separate line of work translates score-estimation error into distributional guarantees for reverse-diffusion sampling~\cite{chen2022sampling,lee2023convergence,benton2024nearly}. Conforti et al.~\cite{conforti2025kl} obtain KL convergence guarantees without early stopping under suitable assumptions. We apply the score-to-sampling theorem of Benton et al.~\cite{benton2024nearly} and combine it with an early-stopping estimate to obtain a total-variation guarantee relative to the initial data distribution.

\subsection{Organization}
Section~\ref{sec:setup} introduces the OU diffusion, the DSM regression formulation, and the shallow-network estimator.
Section~\ref{sec:reg} establishes the two complementary Sobolev bounds for diffusion scores and their time-uniform envelope.
Section~\ref{sec:shallow approx} uses these bounds to derive shallow-network approximation guarantees.
Section~\ref{sec:stats} develops a general finite-sample analysis of DSM, which is specialized to shallow ReLU networks in Section~\ref{sec:stats shallow} to obtain explicit convergence rates.
Section~\ref{sec:sampling} applies these score-estimation guarantees to derive error bounds for reverse-process sampling.
Technical proofs are deferred to the appendices.

\section{Problem Setup}
\label{sec:setup}

\subsection{Ornstein--Uhlenbeck Diffusion and Score Functions}

Consider the OU diffusion introduced in \eqref{eq:forward_diffusion_intro}, with
initial distribution $P_0$. Its explicit solution is
\begin{equation}
    X_t
    =
    e^{-t}X_0+\sigma_t Z,
    \qquad
    \sigma_t^2:=1-e^{-2t},
    \qquad
    Z\sim\mathcal N(0,I_d),
    \label{eq:ou solution}
\end{equation}
where $Z$ is independent of $X_0$. As in the introduction, we write
$P_t:=\operatorname{Law}(X_t)$. For every $t>0$, the Gaussian
perturbation in \eqref{eq:ou solution} ensures that $P_t$ admits a
smooth, strictly positive density $p_t$, whose score function is
$s_t:=\nabla\log p_t$. We write
$s_t=(s_{t,1},\ldots,s_{t,d})^\top$ for its coordinate representation.
When $P_0$ admits a density, we denote it by $p_0$.

\subsection{Denoising Score Matching}

For a fixed time $t>0$, let $Y_t:=-Z/\sigma_t$ denote the
denoising label. As discussed in the introduction, Tweedie's
identity gives
\begin{equation}
    \mathbb E[Y_t\mid X_t]=s_t(X_t),
    \label{eq:tweedie}
\end{equation}
which identifies the score as the regression function of
$Y_t$ given $X_t$.

For a measurable predictor $f:\mathbb R^d\to\mathbb R^d$,
we define the population DSM objective and the squared
score-estimation risk, respectively, by
\begin{equation*}
    \mathcal J_t(f)
    :=\mathbb E_{(X_t,Y_t)}
    \left[\|f(X_t)-Y_t\|_2^2\right],
    \qquad
    \mathcal R_t(f)
    :=\mathbb E_{X\sim P_t}
    \left[\|f(X)-s_t(X)\|_2^2\right].
\end{equation*}
Conditioning on $X_t$ and using \eqref{eq:tweedie}, we obtain
\begin{equation*}
    \mathcal J_t(f)
    =\mathcal R_t(f)
    +\mathbb E\left[\|Y_t-s_t(X_t)\|_2^2\right].
\end{equation*}
Since the second term is independent of $f$, it follows that
\begin{equation}
    \mathcal J_t(f)-\mathcal J_t(s_t)
    =\mathcal R_t(f).
    \label{eq:excess risk identity}
\end{equation}
Thus, DSM can be viewed as a vector-valued least-squares
regression problem, with population excess loss equal to
the squared $L^2(P_t)$ score-estimation error.
This identity provides the basis for our finite-sample analysis.

An important feature of this regression problem is that
both $X_t$ and $Y_t$ are unbounded, and the regression
residual $Y_t-s_t(X_t)$ is generally dependent on $X_t$.
These features motivate the statistical analysis developed
in Sections~\ref{sec:stats} and~\ref{sec:stats shallow}.

\subsection{Shallow  Neural-Network Estimator}

Let $\sigma(u)=\max\{u,0\}$ denote the ReLU activation, and write $\widetilde x=(x^\top,1)^\top\in\mathbb R^{d+1}$.
For an integer $m\geq 1$ and a path-norm budget $M>0$,
define the scalar shallow-network class
\begin{equation*}
    \Sigma_{m,M}^{1}
    :=\left\{
        x\mapsto\sum_{\ell=1}^{m}
        a_\ell\sigma(\theta_\ell^\top\widetilde x)
        :\ \theta_\ell\in\mathbb S^d,\quad
        \sum_{\ell=1}^{m}|a_\ell|\leq M
    \right\},
\end{equation*}
where $\mathbb S^d$ is the unit sphere in $\mathbb R^{d+1}$.
Normalizing the hidden parameters in this way expresses the path-norm
constraint through the coefficient budget $M$.
For the vector-valued score, we use
\begin{equation*}
    \mathcal F_{m,M}
    :=
    \left(\Sigma_{m,M}^{1}\right)^d. 
\end{equation*}
Thus, each score coordinate is represented by a network of width $m$,
with at most $dm$ hidden neurons in total.


Let $X_{0,1},\ldots,X_{0,n}\overset{\mathrm{i.i.d.}}{\sim}P_0$
be the training samples, and let
$Z_1,\ldots,Z_n\overset{\mathrm{i.i.d.}}{\sim}\mathcal N(0,I_d)$
be independent Gaussian perturbations. At a fixed time $t>0$,
we construct the input--label pairs
\begin{equation*}
    X_{t,i}:=e^{-t}X_{0,i}+\sigma_t Z_i,
    \qquad
    Y_{t,i}:=-\frac{Z_i}{\sigma_t},
    \qquad i=1,\ldots,n.
\end{equation*}
We consider a measurable empirical DSM minimizer over the
shallow-network class $\mathcal F_{m,M}$:
\begin{equation}
    \widehat s_{t,m}
    \in\operatorname*{arg\,min}_{f\in\mathcal F_{m,M}}
    \frac{1}{n}\sum_{i=1}^{n}
    \|f(X_{t,i})-Y_{t,i}\|_2^2.
    \label{eq:dsm-erm}
\end{equation}
Here, the dependence on the path-norm budget $M$ is suppressed
in the notation. Our objective is to establish high-probability,
nonasymptotic bounds on the score-estimation risk
$\mathcal R_t(\widehat s_{t,m})$, with explicit dependence on
the sample size $n$, network width $m$, diffusion time $t$,
and the regularity and tail properties of $P_0$.

\subsection{From Score Estimation to Reverse Sampling}

We next connect fixed-time score estimation to reverse-process sampling.
Fix a terminal time $T\geq 1$ and an early-stopping time
$\gamma\in(0,1)$. As discussed in the introduction, the exact
reverse OU process~\eqref{eq:reverse_diffusion_intro}, initialized
from $P_T$, satisfies
\begin{equation*}
    \bar X_u\sim P_{T-u},
    \qquad 0\leq u\leq T-\gamma,
\end{equation*}
and hence $\bar X_{T-\gamma}\sim P_\gamma$.
Thus, the reverse dynamics require the score functions
$s_t$ over the time interval $[\gamma,T]$.

In practice, we replace the true scores with estimates trained
at a finite grid of diffusion times and initialize the discretized
reverse sampler from $\mathcal N(0,I_d)$ instead of $P_T$.
Denote the resulting output density by $\widehat p_{\mathrm{out}}$.
In Section~\ref{sec:sampling}, we first bound the Kullback–Leibler (KL) divergence
$\operatorname{KL}(p_\gamma\,\|\,\widehat p_{\mathrm{out}})$
by accounting for accumulated score-estimation error,
time-discretization error, and initialization error.
Combining this KL bound with an estimate of the early-stopping error
between $P_\gamma$ and $P_0$ yields a bound on the total-variation
distance between the generated distribution and the original data
distribution $P_0$.

\section{Regularity of Diffusion Scores} 
\label{sec:reg}

The shallow-network approximation theorem used in
Section~\ref{sec:shallow approx} requires quantitative, coordinatewise
Sobolev bounds on a bounded domain. For $R>0$, write
\begin{equation*}
    \Omega_R:=\{x\in\mathbb R^d:\|x\|_2<R\},
    \qquad s_\star:=\frac{d+3}{2}.
\end{equation*}
The relevant approximation estimate depends on the
$H^{s_\star}(\Omega_R)$ norm of each score component.
{\begin{remark}

The order $s_\star=(d+3)/2$ is motivated by the embedding
$H^{s_\star}(\Omega_R)\hookrightarrow\mathcal B^1(\Omega_R)$,
where $\mathcal B^1$ denotes the ReLU Barron space. Within the
$L^2$-based Sobolev scale, $s_\star$ is the critical smoothness
threshold for this embedding~\cite{liu2025integral,mao2026approximation}.
Although $H^{s_\star}(\Omega_R)$ is a proper subspace of
$\mathcal B^1(\Omega_R)$, their unit balls have the same asymptotic
order of metric entropy in
$L^2(\Omega_R)$~\cite{liu2025integral,siegel2024sharp}.

The associated shallow ReLU approximation rate is
$m^{-1/2-3/(2d)}$ in $L^2$, with a decay exponent that remains
bounded away from zero as $d$ increases. Applying Barron-space
approximation theory directly, however, requires quantitative
control of the target's Barron norm, which is difficult to obtain
because the norm is defined through an infimum over representing
measures. We instead work with Sobolev norms, which can be
bounded explicitly using derivative estimates for the OU density.
Combining these estimates with the shallow ReLU approximation
theorem for Sobolev functions (Lemma~\ref{lem:shallow approx})
yields quantitative bounds on the approximation error and the
required path-norm budget, with explicit dependence on diffusion
time, spatial radius, and the regularity and tail properties
of the initial distribution.

\end{remark}}

We first derive bounds at integer Sobolev orders and then use
interpolation to obtain the required $H^{s_\star}$ estimates.
For an integer $r\geq0$, we use the norm
\begin{equation*}
    \|f\|_{H^r(\Omega_R)}^2
    :=\sum_{|\alpha|\leq r}
    \int_{\Omega_R}|\partial^\alpha f(x)|^2\,\mathrm dx.
\end{equation*}
Our objective is to obtain explicit bounds on
\begin{equation*}
    \max_{1\leq i\leq d}\|s_{t,i}\|_{H^r(\Omega_R)}^2
\end{equation*}
in terms of the diffusion time $t$, spatial radius $R$, and properties
of the initial distribution, with constants depending only on $d$ and $r$.
We derive two complementary estimates by differentiating either the
Gaussian transition kernel or, under suitable regularity assumptions,
the initial density. Taking the pointwise minimum of these estimates
yields a time-uniform Sobolev bound that will be used for shallow-network
approximation.

To develop these estimates, let $P_0$ be an arbitrary probability
distribution on $\mathbb R^d$, with $X_0\sim P_0$, and let $g_t$
denote the density of $\mathcal N(0,\sigma_t^2I_d)$.
The OU representation gives, for every $t>0$,
\begin{equation}
    p_t(x)
    =\int_{\mathbb R^d}g_t(x-e^{-t}y)\,P_0(\mathrm dy).
    \label{eq:ou convolution}
\end{equation}
Since $p_t(x)>0$ for every $x\in\mathbb R^d$, we define the
posterior probability measure
\begin{equation}
    \Pi_{t,x}(\mathrm dy)
    :=\frac{g_t(x-e^{-t}y)\,P_0(\mathrm dy)}{p_t(x)}.
    \label{eq:ou posterior measure}
\end{equation}
By Bayes' rule, $\Pi_{t,x}$ is a version of the conditional
distribution of $X_0$ given $X_t=x$.

For every nonzero multi-index $\nu$, define the normalized density
derivative
\begin{equation*}
    Q_{\nu,t}(x):=\frac{\partial^\nu p_t(x)}{p_t(x)}.
\end{equation*}
The kernel-induced and data-induced approaches developed in
Sections~\ref{sec:kernel-induced-bound} and~\ref{sec:data-induced-bound},
respectively, express $Q_{\nu,t}$ as posterior expectations. For related identities connecting Gaussian
posterior moments, conditional-mean derivatives, and score functions,
see Dytso et al.~\cite{dytso2022conditional}.

Both approaches rely on the radial $\rho$-quantile radius of the
initial distribution, defined by
\begin{equation}
    B_\rho
    :=\inf\left\{
        B\geq0:\mathbb P(\|X_0\|_2\leq B)\geq\rho
    \right\},
    \qquad \rho\in(0,1).
    \label{eq:quantile radius}
\end{equation}
Since $P_0$ is a probability distribution, $B_\rho<\infty$
for every $\rho\in(0,1)$.

\subsection{Kernel-Induced Bounds without Smoothness Assumptions}
\label{sec:kernel-induced-bound}

For a multi-index $\nu\in\mathbb N_0^d$, define the multivariate
probabilists' Hermite polynomial by
\begin{equation*}
    H_\nu(z)
    :=(-1)^{|\nu|}e^{\|z\|_2^2/2}
    \partial^\nu e^{-\|z\|_2^2/2}.
\end{equation*}
Differentiating the Gaussian kernel in \eqref{eq:ou convolution}
gives
\begin{equation}
    Q_{\nu,t}(x)
    =(-1)^{|\nu|}\sigma_t^{-|\nu|}
    \int_{\mathbb R^d}
    H_\nu\!\left(\frac{x-e^{-t}y}{\sigma_t}\right)
    \Pi_{t,x}(\mathrm dy).
    \label{eq:kernel posterior representation}
\end{equation}
This representation requires neither a density nor any smoothness
assumption on $P_0$.

\begin{theorem}[Kernel-induced Sobolev regularity]
\label{thm:kernel-induced-regularity}
Let $r\in\mathbb N$, $R>0$, $t>0$, and $\rho\in(0,1)$.  For every initial probability distribution $P_0$,
\begin{equation}\label{eq:kernel-sobolev-main}
     \max_{i\in[d]}
    \|s_{t,i}\|_{H^r(\Omega_R)}^2
    \le
    C_{d,r}R^d\sigma_t^{-2(r+1)}
    \left(
        1+\sqrt{2\log(\rho^{-1})}
        +\frac{R+e^{-t}B_\rho}{\sigma_t}
    \right)^{2(r+1)}.
\end{equation}
In particular, since $e^{-t}\leq 1$ and $\sigma_t\leq1$,
\begin{equation}
\begin{aligned}
    \max_{i\in[d]}
    \|s_{t,i}\|_{H^r(\Omega_R)}^2
    \le{}&
    C_{d,r}\sigma_t^{-4(r+1)} 
    \Bigl[
        \bigl(
            1+(2\log\rho^{-1})^{r+1}
            +B_\rho^{2(r+1)}
        \bigr)R^d
        +R^{d+2r+2}
    \Bigr].
\end{aligned}
\label{eq:kernel-sobolev-expanded}
\end{equation}
Here, $C_{d,r}>0$ depends only on $d$ and $r$.
\end{theorem}

\begin{proof}[Sketch of proof]
For any multi-index $\gamma$, the multivariate Fa\`a di Bruno formula~\cite{constantine1996multivariate} gives
\begin{equation*}
\partial^\gamma s_{t,i}
=
\sum_{\beta\le\gamma}
\sum_{\pi\in\mathfrak P(\gamma-\beta)}
\binom{\gamma}{\beta}
(-1)^{|\pi|}|\pi|!\,
\frac{\partial^{\beta+e_i}p_t}{p_t}
\prod_{B\in\pi}
\frac{\partial^Bp_t}{p_t}.
\end{equation*}
Here, $\mathfrak P(\eta)$ denotes the partitions of the labeled
derivative indices associated with $\eta$, and $\partial^B$ denotes
the derivative associated with a block $B$. We use
$\mathfrak P(0)=\{\varnothing\}$ and the empty-product convention.
It therefore suffices to control $Q_{\nu,t}$ for
$1\leq|\nu|\leq r+1$.

By \eqref{eq:kernel posterior representation},
\begin{equation*}
Q_{\nu,t}(x)
=
(-1)^{|\nu|}\sigma_t^{-|\nu|}
\mathbb E
\left[
H_\nu\left(
\frac{x-e^{-t}X_0}{\sigma_t}
\right)
\Bigm| X_t=x
\right].
\end{equation*}
The central-mass condition  \eqref{eq:quantile radius}, together with the Gaussian decay of the transition kernel, provides uniform control of the relevant posterior moments over $\Omega_R$. Combined with the polynomial growth of the Hermite polynomials, this yields the required bounds on $Q_\nu$ for all $|\nu|\le r+1$.

Finally, substituting the bounds on $Q_\nu$ into the Fa\`a di Bruno expansion, summing over $|\gamma|\le r$, and integrating over $\Omega_R$ yields \eqref{eq:kernel-sobolev-main}. The expanded estimate
\eqref{eq:kernel-sobolev-expanded} follows using $e^{-t}\leq1$
and $\sigma_t\leq1$.
\end{proof}

\subsection{Data-Induced Bounds from Normalized Weak Derivatives}
\label{sec:data-induced-bound}

To obtain an estimate that remains controlled as $t\downarrow0$,
we instead transfer derivatives from the Gaussian kernel to the
initial density. This requires regularity of $P_0$, but avoids
the negative powers of $\sigma_t$ in the kernel-induced estimate.
Suppose that $P_0(\mathrm dy)=p_0(y)\,\mathrm dy$.

\begin{assumption}[Weak differentiability through order $r+1$]
\label{asm:p0 reg}
The initial density satisfies $p_0\in W^{r+1,1}(\mathbb R^d)$. 
\end{assumption}

Under Assumption~\ref{asm:p0 reg}, integration by parts in the
weak sense gives, for $1\leq|\nu|\leq r+1$,
\begin{equation}
    \partial^\nu p_t(x)
    =e^{t|\nu|}\int_{\mathbb R^d}
    \partial^\nu p_0(y)g_t(x-e^{-t}y)\,\mathrm dy.
    \label{eq:transfer derivatives to p0}
\end{equation}
The following proposition allows us to normalize these weak
derivatives even when $p_0$ vanishes.

\begin{prop}[Absolute continuity of weak derivatives]
\label{prop:absolute continuity}
Suppose that $p_0$ satisfies Assumption~\ref{asm:p0 reg}.
For every multi-index $\nu$ with $1\leq|\nu|\leq r+1$,
the signed measure $\partial^\nu p_0(x)\,\mathrm dx$ satisfies
\begin{equation*}
    \partial^\nu p_0(x)\,\mathrm dx\ll p_0(x)\mathrm{d}x 
\end{equation*}
Consequently, there exists $h_\nu\in L^1(p_0)$, unique $P_0$-almost surely,  such that
\begin{equation}\label{eq:normalized weak derivative}
    \partial^\nu p_0(x)
    =
    h_\nu(x)p_0(x)
    \qquad\text{for Lebesgue-a.e. }x\in\RR^d.
\end{equation}
Moreover,
\begin{equation*}
    \|h_\nu\|_{L^1(p_0)}
    =
    \|\partial^\nu p_0\|_{L^1(\RR^d)}
    <\infty.
\end{equation*}
\end{prop}

Combining \eqref{eq:ou posterior measure},
\eqref{eq:transfer derivatives to p0}, and
\eqref{eq:normalized weak derivative} yields the data-induced
posterior representation
\begin{equation}\label{eq:data-side posterior representation}
     Q_{\nu,t}(x)
     =
    e^{t|\nu|}
    \int_{\mathbb R^d}h_\nu(y)\,\Pi_{t,x}(\mathrm dy)  =
    e^{t|\nu|}
    \mathbb E[h_\nu(X_0)\mid X_t=x].
\end{equation} 
Assumption~\ref{asm:p0 reg} justifies the transfer of derivatives
from the Gaussian kernel to the initial density and the posterior
representation~\eqref{eq:data-side posterior representation}.
However, obtaining quantitative bounds on
$\|s_{t,i}\|_{H^r(\Omega_R)}$ requires additional control of the
normalized derivatives $h_\nu$. We therefore impose the following
growth condition.

\begin{assumption}[Polynomial growth of normalized weak derivatives]
\label{asm:poly growth}
There exist $H\geq1$ and $\kappa\geq0$ such that, for every $1\leq|\nu|\leq r+1$,
\begin{equation*}
    |h_\nu(y)|
    \leq
    H^{|\nu|}(1+\|y\|_2)^{\kappa|\nu|},
    \qquad P_0\text{-a.e. }y.
\end{equation*}
\end{assumption}

\begin{remark}[Motivation for the polynomial-growth condition]
\label{rmk:motivation-poly-growth}
The normalized density derivatives $h_\nu$ are closely related to
the score and its higher-order derivatives. In the diffusion-model
literature, the $m$-th order score tensor is commonly defined as
$\nabla^{(m)}\log p_t$ for $m\geq1$; see, for example,
\cite{lu2022maximum}.

To clarify this relationship, suppose temporarily that $p_0$ is
smooth and strictly positive. Then
\begin{equation*}
    h_\nu(y)=\frac{\partial^\nu p_0(y)}{p_0(y)}.
\end{equation*}
At first order, these normalized derivatives coincide with the
components of the initial score:
\begin{equation*}
    h_{e_j}=\bigl[\nabla\log p_0\bigr]_j.
\end{equation*}
At second order, however,
\begin{equation*}
    h_{e_i+e_j}
    =\bigl[\nabla^2\log p_0\bigr]_{ij}
    +\bigl[\nabla\log p_0\bigr]_i
     \bigl[\nabla\log p_0\bigr]_j.
\end{equation*}
More generally, for $|\nu|=m$, $h_\nu$ can be expressed as a
polynomial in the components of
\[
\nabla\log p_0,\nabla^{(2)}\log p_0,\ldots,
\nabla^{(m)}\log p_0.
\]
Thus, higher-order normalized density derivatives are related
to, but generally differ from, higher-order score tensors.

Growth conditions on scores and their derivatives are common in
the theoretical analysis of diffusion models. For example,
Song et al.~\cite{song2021maximum} assume a linear-growth bound
of the form
\begin{equation*}
    \|s_t(x)\|_2\leq C(1+\|x\|_2),
\end{equation*}
which, at the initial time, corresponds to the first-order case
of Assumption~\ref{asm:poly growth} with $\kappa=1$, up to
dimension-dependent constants. Chang et al.~\cite{chang2025provable}
assume polynomial growth of the initial score, while De Bortoli
et al.~\cite{de2021diffusion} impose polynomial-growth conditions
on higher-order derivatives of the initial log-density.

Our assumption instead imposes growth control directly on the
normalized density derivatives $h_\nu$ appearing in the posterior
representation~\eqref{eq:data-side posterior representation},
while allowing these derivatives to be defined in the weak sense.
\end{remark}

The following example shows that the growth condition is satisfied
by Gaussian distributions and finite Gaussian mixtures. In particular,
the latter demonstrates that the condition also accommodates
multimodal and non-log-concave distributions.

\begin{example}[Gaussian distributions and finite Gaussian mixtures]
\label{ex:finite-gaussian-mixture}
Fix $r\in\mathbb N$, and let $\phi_{\mu,\Sigma}$ be the density
of $\mathcal N(\mu,\Sigma)$ with $\Sigma\succ0$. Set
\begin{equation*}
    A:=\Sigma^{-1},
    \qquad
    P_\nu^{\mu,\Sigma}(y)
    :=
    \frac{\partial^\nu\phi_{\mu,\Sigma}(y)}
         {\phi_{\mu,\Sigma}(y)}.
\end{equation*}
Since
\begin{equation*}
    \nabla\log\phi_{\mu,\Sigma}(y)
    =
    -A(y-\mu),
\end{equation*}
the normalized derivatives satisfy the recurrence
\begin{equation*}
    P_{\nu+e_j}^{\mu,\Sigma}(y)
    =
    \partial_jP_\nu^{\mu,\Sigma}(y)
    -
    \bigl[A(y-\mu)\bigr]_j
    P_\nu^{\mu,\Sigma}(y),
    \qquad
    P_0^{\mu,\Sigma}(y)=1.
\end{equation*}
For example,
\begin{equation*}
    P_{e_j}^{\mu,\Sigma}(y)
    =
    -\bigl[A(y-\mu)\bigr]_j
\end{equation*}
and
\begin{equation*}
    P_{e_i+e_j}^{\mu,\Sigma}(y)
    =
    \bigl[A(y-\mu)\bigr]_i
    \bigl[A(y-\mu)\bigr]_j
    -
    A_{ij}.
\end{equation*}
It follows inductively that $P_\nu^{\mu,\Sigma}$ is a polynomial of degree $|\nu|$. 
Consequently, for every fixed $r$, there exists a constant $H_{\mu,\Sigma,r}\geq1$ such that
\begin{equation*}
    \left|
        P_\nu^{\mu,\Sigma}(y)
    \right|
    \leq
    H_{\mu,\Sigma,r}^{|\nu|}
    (1+\|y\|_2)^{|\nu|},
    \qquad
    1\leq|\nu|\leq r+1.
\end{equation*}

Now consider the finite mixture
\begin{equation*}
    p_0(y)
    =
    \sum_{\ell=1}^K w_\ell\phi_\ell(y),
\end{equation*}
where $w_\ell>0$, $\sum_{\ell=1}^K w_\ell=1$, and $\phi_\ell$ is the
density of $\mathcal N(\mu_\ell,\Sigma_\ell)$ with
$\Sigma_\ell\succ0$. Define the  weights
\begin{equation*}
    \rho_\ell(y)
    :=
    \frac{w_\ell\phi_\ell(y)}{p_0(y)}.
\end{equation*}
Then
\begin{equation*}
    h_\nu(y)
    =
    \sum_{\ell=1}^K
    \rho_\ell(y)
    \frac{\partial^\nu\phi_\ell(y)}{\phi_\ell(y)}.
\end{equation*}
Since $\rho_\ell(y)\geq0$ and
$\sum_{\ell=1}^K\rho_\ell(y)=1$, we obtain
\begin{align*}
    |h_\nu(y)|
    &\leq
    \max_{1\leq\ell\leq K}
    \left|
        \frac{\partial^\nu\phi_\ell(y)}{\phi_\ell(y)}
    \right|  \leq
    H^{|\nu|}(1+\|y\|)^{|\nu|},
    \qquad
    1\leq|\nu|\leq r+1,
\end{align*}
for some $H\geq1$ depending on $r$, $d$, and the mixture parameters.
Thus, Assumption~\ref{asm:poly growth} holds with $\kappa=1$.
\end{example}

We now state the data-induced counterpart of
Theorem~\ref{thm:kernel-induced-regularity}.

\begin{theorem}[Data-induced Sobolev bound on a prescribed ball]
\label{thm:data-induced-regularity}
Let $r\in\mathbb N$, $R>0$, $t>0$, and $\rho\in(0,1)$.  Suppose that Assumptions~\ref{asm:p0 reg} and~\ref{asm:poly growth} hold at order $r$, meaning that the required weak derivatives and growth bounds are available through order $r+1$.  Define
\begin{equation*}
    \mathcal M_{t,\rho,R}
    :=
    1+e^t\left[
        2R+e^{-t}B_\rho
        +\sigma_t\left(
            \sqrt{2\log(\rho^{-1})}
            +2\sqrt{1+\kappa(r+1)}
        \right)
    \right]. 
\end{equation*}
Then
\begin{equation}
    \max_{i\in[d]}
    \|s_{t,i}\|_{H^r(\Omega_R)}^2
    \leq
    C_{d,r}R^d(e^tH)^{2(r+1)}
    \mathcal M_{t,\rho,R}^{2\kappa(r+1)},
    \label{eq:data-induced-regularity}
\end{equation}
where $C_{d,r}>0$ depends only on $d$ and $r$.
\end{theorem}

\begin{proof}[Sketch of Proof]
Assumption~\ref{asm:poly growth} and \eqref{eq:data-side posterior representation} give
\begin{equation*}
    |Q_{\nu,t}(x)|\leq e^{t|\nu|}H^{|\nu|}\mathbb E[(1+\|X_0\|_2)^{\kappa|\nu|}\mid X_t=x].
\end{equation*}
The same central-mass likelihood comparison used in the
kernel-induced argument controls the required conditional moments.
For $1\leq|\nu|\leq r+1$, this yields
\begin{equation*}
    |Q_{\nu,t}(x)|
    \leq
    C_{d,r}^{|\nu|}(e^tH)^{|\nu|}
    \mathcal M_{t,\rho,R}^{\kappa|\nu|},
    \qquad x\in\Omega_R.
\end{equation*}
Substituting these estimates into the Fa\`a di Bruno expansion
and integrating over $\Omega_R$ gives
\eqref{eq:data-induced-regularity}.
\end{proof}

The parameters $H$ and $\kappa$ quantify the permitted growth of
the normalized weak derivatives. Larger admissible values enlarge
the regularity bound. This dependence is carried into the
approximation and statistical estimates in
Sections~\ref{sec:shallow approx} and~\ref{sec:stats shallow}.

\subsection{A Complementary Sobolev Envelope}

Under the assumptions of
Theorem~\ref{thm:data-induced-regularity}, the kernel-induced and
data-induced estimates hold simultaneously. Their complementary
time dependence allows us to obtain a uniform bound by taking
their pointwise minimum. 

For $r\in\mathbb N$, $\rho\in(0,1)$, and $R>0$, define the kernel-induced bound by
\begin{equation*}
    \mathcal K_{r,\rho,R}(t)
    :=
    R^d\sigma_t^{-2(r+1)}
    \left(
        1+\sqrt{2\log(\rho^{-1})}
        +
        \frac{R+e^{-t}B_\rho}{\sigma_t}
    \right)^{2(r+1)}
\end{equation*}
and the data-induced bound by
\begin{equation*}
    \mathcal D_{r,\rho,R}(t)
    :=
    R^d
    (e^tH)^{2(r+1)}
        \mathcal M_{t,\rho,R}^{2\kappa(r+1)}.
\end{equation*}
Their pointwise minimum is denoted by
\begin{equation*}
    \mathcal E_{r,\rho,R}(t)
    :=
    \mathcal K_{r,\rho,R}(t)
    \wedge
    \mathcal D_{r,\rho,R}(t).
\end{equation*}
Let
\begin{equation*}
    \overline{\mathcal E}_{r,\rho,R}
    :=
    \sup_{t>0}\mathcal E_{r,\rho,R}(t).
\end{equation*}

\begin{cor}[Complementary Sobolev envelope]
\label{cor:regularity envelope}
Let $r\in\mathbb N$, $R>0$, and $\rho\in(0,1)$.  Suppose that Assumptions~\ref{asm:p0 reg} and~\ref{asm:poly growth} hold at order $r$.  
Then, for every $t>0$,
\begin{equation}
    \max_{i\in[d]}
    \|s_{t,i}\|_{H^r(\Omega_R)}^2
    \leq
    C_{d,r}\mathcal E_{r,\rho,R}(t).
    \label{eq:regularity-envelope}
\end{equation}
Moreover, the two explicit upper bounds have a unique cross-over time $t^\ast_{r,\rho,R}>0$ satisfying
\begin{equation*}
    \mathcal K_{r,\rho,R}(t^\ast_{r,\rho,R})
    =
    \mathcal D_{r,\rho,R}(t^\ast_{r,\rho,R}),
\end{equation*}
and
\begin{equation}\label{eq:uniform envelope value}
    \overline{\mathcal E}_{r,\rho,R} =
    \mathcal E_{r,\rho,R}(t^\ast_{r,\rho,R})  =
    \mathcal K_{r,\rho,R}(t^\ast_{r,\rho,R})
    =
    \mathcal D_{r,\rho,R}(t^\ast_{r,\rho,R})<\infty.
\end{equation}

Consequently,
\begin{equation}
    \sup_{t>0}\max_{i\in[d]}
    \|s_{t,i}\|_{H^r(\Omega_R)}^2
    \leq
    C_{d,r}\overline{\mathcal E}_{r,\rho,R}.
    \label{eq:uniform-regularity-envelope}
\end{equation}
\end{cor}

\begin{proof}
Taking the smaller bound in
Theorems~\ref{thm:kernel-induced-regularity}
and~\ref{thm:data-induced-regularity} gives
\eqref{eq:regularity-envelope}.
The function $\mathcal K_{r,\rho,R}$ is continuous and strictly
decreasing, with
\begin{equation*}
\begin{aligned}
    \lim_{t\downarrow0}\mathcal K_{r,\rho,R}(t)&=\infty,\\
    \lim_{t\to\infty}\mathcal K_{r,\rho,R}(t)
    &=R^d\bigl(1+\sqrt{2\log(\rho^{-1})}+R\bigr)^{2(r+1)}.
\end{aligned}
\end{equation*}
On the other hand,
\begin{equation*}
    \mathcal M_{t,\rho,R}
    =
    1+B_\rho+2Re^t
    +\left(
        \sqrt{2\log(\rho^{-1})}
        +2\sqrt{1+\kappa(r+1)}
    \right)\sqrt{e^{2t}-1}.
\end{equation*}
Thus, $\mathcal D_{r,\rho,R}$ is continuous and strictly increasing,
with a finite limit as $t\downarrow0$ and an infinite limit as
$t\to\infty$. The intermediate value theorem and strict
monotonicity therefore give a unique crossover time.
Before this time, the minimum equals the increasing function
$\mathcal D_{r,\rho,R}$; after it, the minimum equals the
decreasing function $\mathcal K_{r,\rho,R}$. Hence the envelope
attains its maximum at the crossover, proving
\eqref{eq:uniform envelope value} and
\eqref{eq:uniform-regularity-envelope}.
\end{proof}

\begin{figure}[h]
    \centering
    \includegraphics[width=0.87\linewidth]{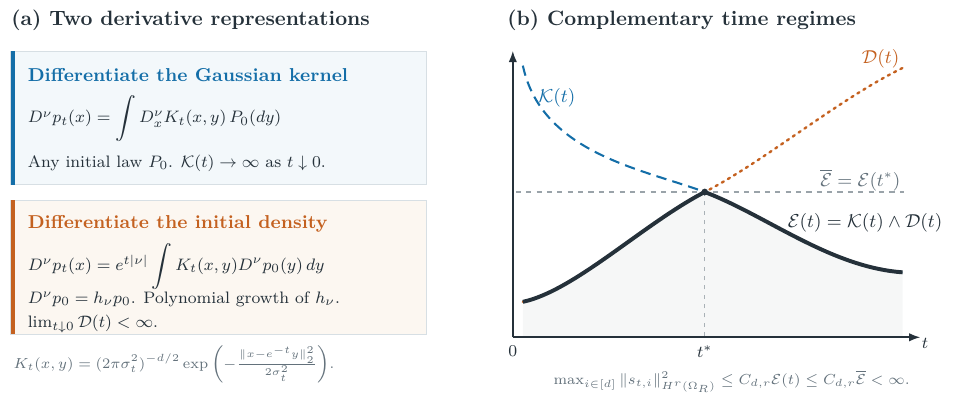}
    \caption{Schematic illustration of the complementary data-induced and kernel-induced Sobolev estimates for the score. The data-induced estimate provides control near $t=0$, while the kernel-induced estimate dominates at larger times. Their pointwise minimum yields a uniform-in-time Sobolev envelope on each fixed ball.}
    \label{fig:ou_sobolev_envelope}
\end{figure}

\begin{remark}[Interpretation of the envelope]
The data-induced estimate controls the score near the initial
time, where the kernel-induced estimate diverges. At later times,
the kernel-induced estimate compensates for the growth of the
data-induced bound. Their minimum therefore remains bounded over
all $t>0$ on each fixed ball, as illustrated in Figure~\ref{fig:ou_sobolev_envelope}.
\end{remark}

\begin{remark}[Extension to fractional Sobolev orders]
\label{rmk:fractional extension}
Let $s>0$ and set
\begin{equation*}
    r_s:=\lceil s\rceil.
\end{equation*}
With the interpolation-space
norm convention for fractional Sobolev spaces
(see, e.g., \cite{bergh1976interpolation}),
\begin{equation*}
    \|f\|_{H^s(\Omega_R)}
    \leq\|f\|_{H^{r_s}(\Omega_R)}.
\end{equation*}
Consequently, if Assumptions~\ref{asm:p0 reg} and~\ref{asm:poly growth} hold at the integer order $r_s$, then Corollary~\ref{cor:regularity envelope} implies
\begin{equation*}
\begin{aligned}
    \max_{1\leq i\leq d}
    \|s_{t,i}\|_{H^s(\Omega_R)}^2
    &\leq
    C_{d,r_s}\mathcal E_{r_s,\rho,R}(t),
    \qquad t>0,
    \\
    \sup_{t>0}
    \max_{1\leq i\leq d}
    \|s_{t,i}\|_{H^s(\Omega_R)}^2
    &\leq
    C_{d,r_s}
    \overline{\mathcal E}_{r_s,\rho,R}.
\end{aligned}
\end{equation*}
\end{remark}

For the approximation and statistical analyses, we also need
explicit control of the uniform envelope as the spatial radius
grows. The next result gives a polynomial bound in $R$.
\begin{cor}
\label{cor:polynomial regularity envelope}
Let $r\in\mathbb N$ and $\rho\in(0,1)$. Suppose that Assumptions~\ref{asm:p0 reg} and~\ref{asm:poly growth} hold through order $r$. Define the distributional regularity coefficient
\begin{align}
    \mathcal C_{\mathrm{reg}}(H,\kappa;r,\rho)
    :={}&
    2^{r+1}
    \left(
        1\vee H^{2(r+1)}
    \right)
    \notag\\
    &\times
    \left(
        1+\sqrt{2\log(\rho^{-1})}
        +2\sqrt2+B_\rho
        +2\sqrt{1+\kappa(r+1)}
    \right)^{
        2(r+1)(1\vee\kappa)
    }.
    \label{eq:distributional regularity coefficient}
\end{align}
Then, for every $R\geq1$,
\begin{equation}
    \overline{\mathcal E}_{r,\rho,R}
    \leq
    \mathcal C_{\mathrm{reg}}(H,\kappa;r,\rho)
    R^{d+2(r+1)(1\vee\kappa)}.
    \label{eq:polynomial regularity envelope}
\end{equation}
\end{cor}

Finally, Sobolev embedding converts the time-uniform regularity
estimate into a pointwise bound on the score.
\begin{cor}[Time-uniform pointwise bound for the diffusion score]
\label{cor:uniform pointwise score bound}
Let $r\in\mathbb N$ satisfy
\begin{equation*}
    r>\frac d2.
\end{equation*}
Suppose that Assumptions~\ref{asm:p0 reg} and~\ref{asm:poly growth} hold at order $r$.
Then, for every $\rho\in(0,1)$ and $R\geq1$,
\begin{equation}
    \sup_{t>0}
\|s_t\|_{L^\infty(\Omega_R;\mathbb R^d)}
\leq
C_{d,r}R^{r-d/2}
\overline{\mathcal E}_{r,\rho,R}^{1/2}.
    \label{eq:uniform pointwise score bound}
\end{equation}
\end{cor}


\section{Shallow Network Approximation of Scores}\label{sec:shallow approx}

Section~\ref{sec:reg} establishes explicit local Sobolev bounds
for diffusion scores, including time-uniform bounds under regularity
and growth assumptions on the initial density. We now use these
estimates to obtain shallow ReLU approximation guarantees with
quantitative control of the network width and path-norm budget.

We begin with a consequence of
\cite[Theorem~2.4]{liu2025integral} for scalar functions on $\Omega_R$.

\begin{lem}[Shallow ReLU approximation of Sobolev functions]
    \label{lem:shallow approx}
    Let $R\geq1$ and $f\in H^{(d+3)/2}(\Omega_R)$. If the path-norm budget $M$ is chosen such that $M=C_{\mathrm{cap},d} R^{\frac{3}{2}} \|f\|_{H^{\frac{d+3}{2}}(\Omega_R)}$ 
    , then
    \begin{equation*}
        \inf_{f_m\in \Sigma_{m,M}^1}\|f-f_m\|_{L^2(\Omega_R)}\lesssim_d R^{\frac{d+3}{2}}m^{-\frac{1}{2}-\frac{3}{2d}}\|f\|_{H^{\frac{d+3}{2}}(\Omega_R)}.
    \end{equation*}
    Here, $C_{\mathrm{cap},d}$ is a constant that depends only on $d$.
\end{lem}

\begin{remark}
The constants in \cite[Theorem~2.4]{liu2025integral} may depend
on the underlying bounded domain.
Lemma~\ref{lem:shallow approx} makes this dependence explicit
for $\Omega_R$ by rescaling to the unit ball and using the
positive homogeneity of ReLU. The details are given in
Appendix~\ref{appendix: shallow approx}.
\end{remark}

The preceding approximation error is measured with respect to
Lebesgue measure. To obtain an $L^2(P_t)$ estimate on the same
ball, we first establish a uniform bound on the marginal
densities $p_t$.

\begin{assumption}\label{asm:density infty}
    The initial distribution admits a density $p_0\in L^\infty(\mathbb R^d)$.
\end{assumption}
\begin{lem}\label{lem:density infty}
Under Assumption~\ref{asm:density infty},
    \begin{equation*}
        \sup_{t>0}\|p_t\|_{\infty}\leq \left(
\|p_0\|_\infty^{2/d}
+\frac{1}{2\pi}
\right)^{d/2}:=P_{\infty}.
    \end{equation*}
\end{lem}
\begin{proof}[Sketch of proof]
    The OU convolution representation gives
    \begin{equation*}
        \|p_t\|_{\infty}\leq \lp 2\pi\sigma_t^2\rp^{-d/2}
    \end{equation*}
    and 
    \begin{equation*}
         \|p_t\|_{\infty}\leq e^{dt}\|p_0\|_\infty. 
    \end{equation*}
    The proof is completed by an envelope argument.
\end{proof}

With these two results, we are able to derive the approximation error of the score function using a shallow network.
\begin{theorem}[Local Shallow-Network Approximation of Diffusion Scores]
\label{thm:score approx}
Let $R\geq1$ and $t>0$. Suppose that
Assumption~\ref{asm:density infty} holds and
\begin{equation*}
    s_{t,i}\in H^{\frac{d+3}{2}}(\Omega_R),
    \qquad i\in[d].
\end{equation*}
Let
\begin{equation*}
    \mathcal F_{m,M_t}:=
    \left(\Sigma_{m,M_t}^1\right)^d,
\end{equation*}
and choose a path-norm budget satisfying
\begin{equation*}
    M_t
    =C_{\mathrm{cap},d}R^{\frac{3}{2}}
    \max_{1\leq i\leq d}
    \|s_{t,i}\|_{H^{\frac{d+3}{2}}(\Omega_R)}.
\end{equation*}
Then
\begin{align*}
    &\inf_{s_m\in\mathcal F_{m,M_t}}
    \left(
    \mathbb E_{X_t\sim p_t}
    \left[
        \|s_m(X_t)-s_t(X_t)\|_2^2
        \mathbf 1_{\{X_t\in\Omega_R\}}
    \right]
    \right)^{1/2}
    \lesssim_d
    P_\infty^{1/2}
    R^{\frac{d+3}{2}}
    m^{-\frac12-\frac{3}{2d}}
    \max_{1\leq i\leq d}
    \|s_{t,i}\|_{H^{\frac{d+3}{2}}(\Omega_R)}.
\end{align*}
Moreover, if $I\subset(0,\infty)$ satisfies
\begin{equation*}
    \sup_{t\in I}\max_{1\leq i\leq d}
    \|s_{t,i}\|_{H^{(d+3)/2}(\Omega_R)}<\infty,
\end{equation*}
then a common finite budget $M$, independent of $t\in I$,
can be used in place of $M_t$.
\end{theorem}

Set $q_d:=\lceil(d+3)/2\rceil$. Under
Assumptions~\ref{asm:p0 reg} and~\ref{asm:poly growth} at
order $q_d$, Corollary~\ref{cor:regularity envelope} and
Remark~\ref{rmk:fractional extension} provide the time-uniform
Sobolev bound required by Theorem~\ref{thm:score approx}.
Thus, on each fixed ball, separate score approximants at all
$t>0$ can be chosen with a common path-norm budget and a
time-uniform approximation bound.

The approximation guarantee is local in space. To control
the full $L^2(P_t)$ score-estimation risk, the statistical
analysis below also bounds the contribution from outside
$\Omega_R$.


\section{Statistical Estimation via Denoising Score Matching}
\label{sec:stats}

Throughout this section, we fix a diffusion time $t>0$ and a sample
size $n\geq2$. Recall the excess-risk identity
\eqref{eq:excess risk identity}:
\begin{equation*}
    \mathcal J_t(f)-\mathcal J_t(s_t)=\mathcal R_t(f).
\end{equation*}
We use this identity to derive an oracle inequality for empirical
DSM over a general function class. The bound separates approximation
error, statistical complexity, tail contributions, and confidence
terms.

\paragraph{Empirical Risk Minimization.}
For notational convenience, write
\begin{equation*}
    X:=X_t,
    \qquad
    Y:=Y_t=-\frac{Z}{\sigma_t}.
\end{equation*}
Let $P$ denote the joint law of $(X,Y)$ and $P_X$ the marginal
law of $X$. Given an i.i.d. sample
\begin{equation*}
    S:=\bigl((X_i,Y_i)\bigr)_{i=1}^n\sim P^n,
\end{equation*}
we write
\begin{equation*}
    Pg:=\mathbb E[g(X,Y)],
    \qquad
    P_ng:=\frac1n\sum_{i=1}^n g(X_i,Y_i)
\end{equation*}
for every integrable function
$g:\mathbb R^d\times\mathbb R^d\to\mathbb R$.

{We first work with a general class $\mathcal F$ of measurable functions from $\mathbb R^d$ to $\mathbb R^d$.
In the next section, we specialize to the shallow-network class $\mathcal F_{m,M}$.} For $f\in\mathcal F$, define
the pointwise DSM loss by
\begin{equation*}
    \ell_f(x,y):=\|f(x)-y\|_2^2,
\end{equation*}
and let $\widehat f$ be an empirical DSM minimizer:
\begin{equation}
    \widehat f
    \in
    \operatorname*{arg\,min}_{f\in\mathcal F}P_n\ell_f
    =
    \operatorname*{arg\,min}_{f\in\mathcal F}
    \frac1n\sum_{i=1}^n\|f(X_i)-Y_i\|_2^2.
    \label{eq:general dsm erm}
\end{equation}
The best approximation error within $\mathcal F$ is
\begin{equation}
    \varepsilon_{\mathrm{app}}
    :=
    \inf_{f\in\mathcal F}\mathcal R_t(f)
    =
    \inf_{f\in\mathcal F}
    \mathbb E_{X\sim P_X}
    \|f(X)-s_t(X)\|_2^2.
    \label{eq:approx error}
\end{equation}


For $f\in\mathcal F$, define the excess DSM loss relative to the true
score by
\begin{equation*}
    h_f(x,y)
    :=
    \|f(x)-y\|_2^2-\|s_t(x)-y\|_2^2.
\end{equation*}
Its expectation equals the score-estimation risk:
\begin{equation*}
    Ph_f
    =
    \mathbb E_{X\sim P_X}\|f(X)-s_t(X)\|_2^2
    =
    \mathcal R_t(f).
\end{equation*}
Taking $\mathcal F=\mathcal F_{m,M}$ in
\eqref{eq:general dsm erm} recovers the shallow-network estimator
$\widehat s_{t,m}$ defined in \eqref{eq:dsm-erm}.

\paragraph{Control on a Bounded Set and Tail Terms.}
The main difficulty in the statistical analysis of DSM is that both $X$ and $Y$ are unbounded. Consequently, the loss $\ell_f(X,Y)$ is not uniformly bounded on the sample space, so standard bounded empirical-process inequalities do not apply directly \cite{bartlett2005local,srebro2010optimistic,rakhlin2017empirical}.  
We therefore analyze the excess loss on a bounded input--label set
and control its contribution outside that set separately. This
truncation is used only in the analysis: the estimator
\eqref{eq:general dsm erm} still minimizes the original,
unmodified DSM objective.

{For $R=(R_X,R_Y)$ with $R_X,R_Y>0$, define the truncation set
and its tail probability by
\begin{equation}
    \mathcal A_R
    :=
    \left\{
        (x,y)\in\mathbb R^d\times\mathbb R^d:
        \|x\|_2\leq R_X,\quad \|y\|_2\leq R_Y
    \right\},
    \qquad
    \eta_R
    :=
    \mathbb P\bigl((X,Y)\notin\mathcal A_R\bigr).\label{eq:eta_R}
\end{equation}
Assume that there exists $M_R>0$ such that both the predictors
and the true score are bounded on the spatial ball:
\begin{equation}
    \sup_{f\in\mathcal F}
    \sup_{\|x\|_2\leq R_X}\|f(x)\|_2
    \leq M_R,
    \qquad
    \sup_{\|x\|_2\leq R_X}\|s_t(x)\|_2
    \leq M_R.\label{eq:M_R requirements}
\end{equation}

To control the population contribution outside $\mathcal A_R$,
define
\begin{equation}\label{eq:tail bias}
    \Delta_R
    :=
    \sup_{f\in\mathcal F}
    \left|
        \mathbb E\left[
            h_f(X,Y)\mathbf 1_{\mathcal A_R^c}(X,Y)
        \right]
    \right|,
\end{equation}
and assume that $\Delta_R<\infty$.
The quantity $\eta_R$ is the probability that a training observation
falls outside $\mathcal A_R$, whereas $\Delta_R$ bounds the
population excess-loss contribution from this region uniformly
over $\mathcal F$. In Section~\ref{sec:stats shallow}, we choose
the truncation radii to control both quantities.

Set
\begin{equation}
    U_R:=4M_R(M_R+R_Y),
    \qquad
    B_R:=4(M_R+R_Y)^2.\label{eq:truncation const}
\end{equation}
For every $f\in\mathcal F$ and $(x,y)\in\mathcal A_R$, we have
\begin{align*}
    |h_f(x,y)|
    &\leq
    2(M_R+R_Y)\|f(x)-s_t(x)\|_2
    \leq U_R.
\end{align*}
Squaring the first inequality also yields
\begin{equation*}
    h_f(x,y)^2
    \leq
    B_R\|f(x)-s_t(x)\|_2^2.
\end{equation*}
Thus, $U_R$ bounds the absolute excess loss on $\mathcal A_R$,
while $B_R$ enters the Bernstein-type variance condition used
in the proof.}

\paragraph{Complexity Bound and Oracle Inequality.}
We next bound the statistical complexity using local Rademacher analysis.
For each coordinate $j\in\{1,\ldots,d\}$, define
\begin{equation*}
    \mathcal G_{j,R}
    :=
    \left\{
        (x,y)\mapsto
        \bigl(f_j(x)-s_{t,j}(x)\bigr)
        \mathbf 1_{\mathcal A_R}(x,y):
        f\in\mathcal F
    \right\}
\end{equation*}
and its star hull
\begin{equation*}
    \mathcal G_{j,R}'
    :=
    \operatorname{star}(\mathcal G_{j,R}),
    \qquad
    \operatorname{star}(\mathcal G)
    :=
    \{\lambda g:g\in\mathcal G,\ \lambda\in[0,1]\}.
\end{equation*}
For a scalar function class $\mathcal G$, let
$\mathcal N_2(\varepsilon,\mathcal G,S)$ denote its external
covering number under the empirical $L^2$ metric
\begin{equation*}
    d_S(g,g')
    :=
    \left(
        \frac1n\sum_{i=1}^n
        |g(X_i,Y_i)-g'(X_i,Y_i)|^2
    \right)^{1/2}.
\end{equation*}

Define
\begin{equation*}
    Q_R:= 22M_R+2R_Y.
\end{equation*}
For a localization parameter $r\geq0$,
define the coordinatewise entropy integral
\begin{align}\label{eq:D_j,R}
    \mathfrak D_{j,R}(r)
    :=\;&
    10\mathbb E_S
        \int_0^{\sqrt{2r/B_R}}
        \sqrt{
            \frac{
                \log\mathcal N_2(
                    \varepsilon,\mathcal G_{j,R}',S
                )
            }{n}
        }
        \,\mathrm d\varepsilon,
\end{align}
where the expectation is over $S\sim P^n$. We combine these
coordinatewise quantities in
\begin{equation}
    \Phi_R(r)
    :=B_RQ_R\sum_{j=1}^d\mathfrak D_{j,R}(r)
    +\frac{2M_RB_RQ_R}{n}
    +\frac{44B_RM_R^2\log(dn)}{n}.
    \label{eq:subroot function}
\end{equation}
The entropy integrals control the Rademacher averages appearing
in the proof. Following the local Rademacher complexity approach
of~\cite{bartlett2005local}, we express the complexity term in the
oracle inequality through the fixed point of a sub-root function
that majorizes $\Phi_R$.

\begin{defi}[Sub-root function]
\label{def: subroot}
A function $\psi:[0,\infty)\to[0,\infty)$ is called sub-root
if it is nonnegative and nondecreasing, and
$r\mapsto\psi(r)/\sqrt r$ is nonincreasing on $(0,\infty)$.
\end{defi}

We can now state the main statistical result of this section.

\begin{theorem}[Oracle Inequality for Denoising Score Matching]
\label{thm:local oracle}
Let $\psi_R$ be any nontrivial sub-root function in the sense of Definition~\ref{def: subroot}
satisfying
\begin{equation*}
    \psi_R(r)\geq\Phi_R(r)
    \qquad\text{for every }r\geq0,
\end{equation*}
and let $r_R^*>0$ denote its unique positive fixed point, so that
$\psi_R(r_R^*)=r_R^*$.
Then, for every $u>0$, with probability at least
\begin{equation*}
    1-2e^{-u}-n\eta_R,
\end{equation*}
the empirical risk minimizer $\widehat f$ in
\eqref{eq:general dsm erm} satisfies
\begin{align*}
    \mathcal{R}_t (\widehat f)
    \leq
    4\varepsilon_{\mathrm{app}}
    +8\Delta_R
    +\frac{1408}{B_R}r_R^* +
    \left(
        53B_R+\frac{70}{3}U_R
    \right)
    \frac{u}{n},
\end{align*}
where $\varepsilon_{\mathrm{app}}$ is defined in
\eqref{eq:approx error},
{$\eta_R$ is defined in \eqref{eq:eta_R}, and the constants $U_R, B_R$ are defined in \eqref{eq:truncation const}. } 
\end{theorem}

\begin{remark}
The bound separates four contributions: the best approximation
error within $\mathcal F$, the tail term $\Delta_R$, the localized
complexity term determined by $r_R^*/B_R$, and a confidence term
proportional to $u/n$. The additional failure probability
$n\eta_R$ accounts for the possibility that at least one training
observation lies outside $\mathcal A_R$.
\end{remark}


Section~\ref{sec:stats shallow} applies this oracle inequality
to the shallow-network class $\mathcal F_{m,M}$. We bound the
fixed point $r_R^*$ and the tail quantities $\eta_R$ and
$\Delta_R$, and combine these estimates with the approximation
result from Section~\ref{sec:shallow approx} to obtain explicit
finite-sample guarantees.

\section{Statistical Guarantees for Shallow-Network Score Estimation}
\label{sec:stats shallow}

We apply the oracle inequality from Section~\ref{sec:stats} to the
shallow-network class
\begin{equation*}
    \mathcal F_{m,M}=(\Sigma_{m,M}^1)^d.
\end{equation*}
Combining a network complexity bound with the approximation result
from Section~\ref{sec:shallow approx} yields an
approximation--estimation tradeoff. Under a sub-Gaussian assumption
on the initial distribution, we then choose the truncation radii,
path-norm budget, and network width to obtain an explicit
finite-sample guarantee. The resulting bound quantifies the effects
of distributional regularity, tail behavior, and diffusion time.

\subsection{Fixed-Point Bound for Shallow Networks}
\label{sec:shallow-fixed-point}

The path-norm constraint gives the following growth bound,
which is independent of the network width.
\begin{lem}[Pointwise bound for shallow networks]
\label{lem:localized shallow bound}
For every $x\in\mathbb R^d$ and every $f\in\mathcal F_{m,M}$,
\begin{equation*}
    \|f(x)\|_2
    \leq
    \sqrt d\,M\sqrt{1+\|x\|_2^2}.
\end{equation*}
Consequently,
\begin{equation*}
    \sup_{f\in\mathcal F_{m,M}}
    \sup_{\|x\|_2\leq R_X}
    \|f(x)\|_2
    \leq
    \sqrt d\,M\sqrt{1+R_X^2}.
\end{equation*}
\end{lem}

Fix $t>0$ and $R=(R_X,R_Y)$, choose
\begin{equation} \label{eq:shallow truncation constants}
    M_R
    :=
        \sqrt d\,M\sqrt{1+R_X^2}+
        \sup_{\|x\|_2\leq R_X}\|s_t(x)\|_2
    .
\end{equation}
This choice bounds both the network outputs and the true score on
$\{x:\|x\|_2\leq R_X\}$, as required by
\eqref{eq:M_R requirements}. The supremum is finite for each
$t>0$ because $s_t$ is smooth. Under
Assumptions~\ref{asm:p0 reg} and~\ref{asm:poly growth} with
$r>d/2$, Corollary~\ref{cor:uniform pointwise score bound}
also gives a bound that is uniform over $t>0$.

With the chosen $M_R$, the corresponding constants are
\begin{equation*}
    U_R=4M_R(M_R+R_Y),
    \qquad
    B_R=4(M_R+R_Y)^2.
\end{equation*}
In particular, $M_R$, $U_R$, and $B_R$ do not depend on $m$
when $t$, $R$, and $M$ are fixed.

The next proposition bounds the fixed point that determines the
complexity term in Theorem~\ref{thm:local oracle}.

\begin{prop}[Fixed-point bound for shallow networks]
\label{prop:shallow fixed point}
Let $\Phi_R$ be defined in \eqref{eq:subroot function} for the function class $\mathcal F_{m,M}$. Assume that
$n\geq2$ and $m\leq n$. There exists a nontrivial sub-root function
$\psi_{m,M,R}$ such that
\begin{equation*}
    \psi_{m,M,R}(r)\geq\Phi_R(r),
    \qquad r\geq0,
\end{equation*}
whose fixed point $r_{m,M,R}^*$ satisfies
\begin{align*}
    r_{m,M,R}^*
    \leq{}&
    \frac{200Q_R^2B_Rd^2(d+3)m}{n}\Lambda_{n,R}+
    \frac{4M_RB_RQ_R+88B_RM_R^2\log(dn)}{n},
\end{align*}
where
\begin{equation}\label{eq:log factor defi}
    \Lambda_{n,R}
    :=
    1+\log\left[
        \sqrt n\max\left\{
            4,
            1+\frac{2\sqrt{1+R_X^2}(n+M^2)}{M_R}
        \right\}
    \right].
\end{equation}
\end{prop}

\begin{remark}\label{rmk:Lambda_n,R log term}
For fixed $t$, $R$, and $M$, the factor $\Lambda_{n,R}=O(\log n)$. Thus, the width-dependent contribution to the oracle bound has order $m/n$, up to a logarithmic factor.
\end{remark}

\subsection{Approximation-Estimation Tradeoff}
Set $q_d:=\lceil(d+3)/2\rceil$. Throughout the remainder of this
section, suppose that Assumptions~\ref{asm:p0 reg}
and~\ref{asm:poly growth} hold with $r=q_d$, and that
Assumption~\ref{asm:density infty} holds. Fix an admissible pair
$(H,\kappa)$ in Assumption~\ref{asm:poly growth}, set $\rho=1/2$,
and write $P_\infty$ for the time-uniform density bound in
Lemma~\ref{lem:density infty}.

Fix $R_X\geq1$. The fractional-order extension in
Remark~\ref{rmk:fractional extension} gives
\begin{equation*}
    \sup_{t>0}
    \max_{1\leq i\leq d}
    \|s_{t,i}\|_{H^{\frac{d+3}{2}}(\Omega_{R_X})}^2
    \leq
    C_d\overline{\mathcal E}_{q_d,1/2,R_X}.
\end{equation*}
Consequently, by Theorem~\ref{thm:score approx}, it is sufficient to choose the time-independent path-norm budget
\begin{equation}\label{eq:cap choice}
    M=C_{\mathrm{cap},d}R_X^{3/2}\overline{\mathcal E}_{q_d,1/2,R_X}^{1/2},
\end{equation}
where $C_{\mathrm{cap},d}$ is a fixed, sufficiently large constant depending only on $d$. Theorem~\ref{thm:score approx} then yields
\begin{align*}
    &\inf_{f\in\mathcal F_{m,M}}
    \mathbb E_{X_t\sim p_t}
    \left[
        \|f(X_t)-s_t(X_t)\|_2^2
        \mathbf 1_{\{X_t\in\Omega_{R_X}\}}
    \right] \leq
    C_dP_\infty
    R_X^{d+3}\overline{\mathcal E}_{q_d,1/2,R_X}
    m^{-\frac{d+3}{d}}.
\end{align*}

To pass from local approximation to the full $L^2(P_t)$ score risk,
define the width-uniform tail bound
\begin{equation}
    \tau_{t,R_X}(M):=
    \sup_{k\geq1}\sup_{f\in\mathcal F_{k,M}}
    \mathbb E\left[
        \|f(X_t)-s_t(X_t)\|_2^2
        \mathbf 1_{\{X_t\notin\Omega_{R_X}\}}
    \right].
    \label{eq:approximation tail}
\end{equation}
Consequently,
\begin{equation}
    \varepsilon_{\mathrm{app}}
    \leq
    C_dP_\infty
    R_X^{d+3}\overline{\mathcal E}_{q_d,1/2,R_X}
    m^{-\frac{d+3}{d}}
    +
    \tau_{t,R_X}(M).
    \label{eq:approximation tail decomposition}
\end{equation}
The tail quantities $\tau_{t,R_X}(M)$ and $\Delta_R$ serve different
purposes: the former bounds the squared approximation error outside
$\Omega_{R_X}$, while the latter controls the truncation bias of the
excess DSM loss on the joint input--label space.

Define the approximation and estimation coefficients
\begin{equation}\label{eq:app est coeff}
\Gamma_{\mathrm{app}}(R_X):=P_\infty R_X^{d+3}\overline{\mathcal E}_{q_d,1/2,R_X},
\quad
\Gamma_{\mathrm{est}}(n,R):=281600Q_R^2d^2(d+3)\Lambda_{n,R},
\end{equation}
and the width-independent remainder
\begin{align}\label{eq:width-dependent remainder}
    \operatorname{Rem}_{n,R}(u)
    :=
    \frac{
        5632M_RQ_R
        +
        123904M_R^2\log(dn)
    }{n} +
    \left(
        53B_R+\frac{70}{3}U_R
    \right)
    \frac{u}{n}.
\end{align}

\begin{theorem}[Approximation--estimation tradeoff for shallow networks]
\label{thm:approx-est tradeoff}
Fix $t>0$, let $R=(R_X,R_Y)$ with $R_X\geq1$, and suppose that $n\geq2$ and the preceding assumptions hold. Let $M$ be the time-independent path-norm budget defined by \eqref{eq:cap choice}. For every integer
$1\leq m\leq n$, let $\widehat f_m$ be an empirical DSM minimizer over $\mathcal F_{m,M}$, as defined in
\eqref{eq:general dsm erm}. Then, for every $u>0$, with probability
at least
\begin{equation*}
    1-2e^{-u}-n\eta_R,
\end{equation*}
we have
\begin{align}
    \mathcal R_t(\widehat f_m)
    \leq{}&
    C_d\Gamma_{\mathrm{app}}(R_X)m^{-\frac{d+3}{d}}
    +
    \Gamma_{\mathrm{est}}(n,R)\frac{m}{n}
    +
    4\tau_{t,R_X}(M)
    +
    8\Delta_R
    +
    \operatorname{Rem}_{n,R}(u).
    \label{eq:pre balanced rate}
\end{align}
Here $\Delta_R$, defined in \eqref{eq:tail bias}, is evaluated over
$\mathcal F_{m,M}$.

To balance the two width-dependent terms, define
\begin{equation*}
    \widetilde m
    :=
    \left(
        \frac{\Gamma_{\mathrm{app}}(R_X)n}{\Gamma_{\mathrm{est}}(n,R)}
    \right)^{\frac{d}{2d+3}}.
\end{equation*}
If $1\leq\lceil\widetilde m\rceil\leq n$ and
$m=\lceil\widetilde m\rceil$, then, with the same probability bound,
\begin{equation}
    \begin{aligned}
    \mathcal R_t(\widehat f_m)
    \leq{}&
    C_d
    \Gamma_{\mathrm{app}}(R_X)^{\frac{d}{2d+3}}
    \left(
        \frac{\Gamma_{\mathrm{est}}(n,R)}{n}
    \right)^{\frac{d+3}{2d+3}}
    +
    4\tau_{t,R_X}(M)
    +
    8\Delta_R
    \\
    &+
    \operatorname{Rem}_{n,R}(u)
    +
    \frac{\Gamma_{\mathrm{est}}(n,R)}{n}.
    \label{eq:balanced rate}
\end{aligned}
\end{equation} 

\end{theorem}
\begin{remark}
For fixed truncation radii $R$, balancing the approximation and
estimation terms gives $m=\widetilde O(n^{d/(2d+3)})$ and a
width-dependent contribution of order
$\widetilde O(n^{-(d+3)/(2d+3)})$. To obtain convergence of the full
risk, we must also control $\tau_{t,R_X}(M)$ and $\Delta_R$, which
need not vanish for fixed $R$. The following subsections address
these terms by allowing the truncation radii to increase with $n$.
\end{remark}

\subsection{Tail Control under Sub-Gaussian Data}

We now impose a sub-Gaussian assumption on $X_0$ to control both the
probability $\eta_R$ that an observation lies outside the truncation
set $\mathcal A_R$ and the tail contributions $\tau_{t,R_X}(M), \Delta_R$ in \eqref{eq:balanced rate}.

\begin{assumption}[Sub-Gaussian data]
\label{asm:subg}
The initial random vector $X_0$ is sub-Gaussian in the sense that
\begin{equation*}
    K_0
    :=
    \sup_{u\in\mathbb S^{d-1}}
    \|\langle X_0,u\rangle\|_{\psi_2}
    <\infty,
\end{equation*}
where the sub-Gaussian norm is defined by
\begin{equation*}
    \|W\|_{\psi_2}
    :=
    \inf\left\{
        c>0:
        \mathbb E\exp(W^2/c^2)\leq2
    \right\}.
\end{equation*}
\end{assumption}
For $Z\sim\mathcal N(0,I_d)$, write
\begin{equation*}
    K_Z:=\sup_{u\in\mathbb S^{d-1}}
    \|\langle Z,u\rangle\|_{\psi_2},
    \qquad K:=K_0+K_Z,
\end{equation*}
where $K_Z$ is a universal constant independent of the data distribution.

\begin{prop}[Tail bounds for the inputs and DSM labels]
\label{prop:subG}
Under Assumption~\ref{asm:subg}, fix $t>0$. If
\begin{equation*}
    R_X\geq K\sqrt{8d\log 5},
    \qquad
    R_Y\geq\frac{2\sqrt d}{\sigma_t},
\end{equation*}
then
\begin{equation*}
    \mathbb P\bigl(\|X_t\|_2>R_X\bigr)
    \leq
    2\exp\left(-\frac{R_X^2}{8K^2}\right) \qquad \text{and}\qquad\mathbb P\bigl(\|Y\|_2>R_Y\bigr)
    \leq
    \exp\left(-\frac{\sigma_t^2R_Y^2}{8}\right).
\end{equation*}

Consequently,
\begin{equation*}
    \eta_R\leq 2\exp\left(-\frac{R_X^2}{8K^2}\right)
    +\exp\left(-\frac{\sigma_t^2R_Y^2}{8}\right).
\end{equation*}
\end{prop}
\begin{remark}[Sub-Gaussian quantile radius]
Recall that $B_\rho$, defined in \eqref{eq:quantile radius}, denotes the $\rho$-quantile radius of $X_0$.
The same $1/2$-net argument used in the proof of  Proposition~\ref{prop:subG} in Appendix~\ref{app:stats shallow} yields
\begin{equation*}
    B_\rho\leq2K_0\sqrt{d\log 5+\log\frac{2}{1-\rho}}, \qquad \rho\in(0,1).
\end{equation*}

Hence,
\begin{equation*}
B_{1/2}\leq 2K_0\sqrt{d\log 5+\log 4} \lesssim K_0\sqrt d.
\end{equation*}
\end{remark}

Finally, the next proposition controls the tail terms $\tau_{t,R_X}(M)$ and $\Delta_R$ under sub-Gaussian data.

\begin{prop}[Control of the tail contributions]
\label{prop:tail control}
Under Assumption~\ref{asm:subg}, fix $t>0$, $R_X\geq1$, $R_Y>0$,
and $M>0$. Uniformly over the network width,
\begin{align*}
    \tau_{t,R_X}(M)
    &\lesssim_d\left(M^2(1+K_0)^2
        +{\sigma_t^{-2}}\right)\eta_R^{1/2},\\
    \Delta_R
    &\lesssim_d\left(M^2(1+K_0)^2
        +{\sigma_t^{-2}}\right)\eta_R^{1/2}.
\end{align*}
Here $\Delta_R$ is evaluated over $\mathcal F_{m,M}$ for any
integer $m\geq1$. With the path-norm budget in \eqref{eq:cap choice},
\begin{equation*}
    \tau_{t,R_X}(M)\vee\Delta_R
    \lesssim_d\left(
        R_X^3\overline{\mathcal E}_{q_d,1/2,R_X}(1+K_0)^2
        +\sigma_t^{-2}\right)\eta_R^{1/2}.
\end{equation*} 
\end{prop}

\subsection{Explicit Rate under Sub-Gaussian Data}

Define
\begin{equation*}
    \gamma_d :=  \frac{d}{2d+3},
    \qquad
    \beta_d := \frac{d+3}{2d+3}.
\end{equation*} 

We now choose the truncation radii and network width to balance
approximation and estimation while controlling the tail and
confidence terms. The theorem below expresses the resulting risk
bound through the regularity envelope; the subsequent corollary
makes its dependence on the initial distribution explicit.

\begin{theorem}[Statistical rate under sub-Gaussian data]
\label{thm:concrete subgaussian rate}
Suppose that Assumptions~\ref{asm:p0 reg}  and~\ref{asm:poly growth} hold with $r=q_d$, and that Assumption~\ref{asm:density infty} and \ref{asm:subg} hold. Fix $t>0$, $\delta\in(0,1)$, and $n\geq2$. Choose the truncation radii
\begin{align*}
    R_{X,n}&:=1\vee K\sqrt{
        8\max\left\{d\log5,\log\frac{4n^2}{\delta}\right\}},\\
    R_{Y,n,t}&:=\sigma_t^{-1}
        \max\left\{2\sqrt d,\sqrt{8\log\frac{2n^2}{\delta}}\right\},
\end{align*}
and write
\begin{equation*}
    R_{n,t}:=(R_{X,n},R_{Y,n,t}),\qquad
    \ell_{n,\delta}:=1+\log\frac{4n^2}{\delta}.
\end{equation*} 
Define the regularity envelope and path-norm budget by
\begin{equation}\label{eq:cap choice with n}
     E_n:=\overline{\mathcal E}_{q_d,1/2,R_{X,n}},
    \qquad
    M_n:=C_{\mathrm{cap},d}R_{X,n}^{3/2}E_n^{1/2}.
\end{equation} 
Evaluate the constants in \eqref{eq:shallow truncation constants}
at $M=M_n$ and $R=R_{n,t}$, and write $\Lambda_n:=\Lambda_{n,R_{n,t}}$. Dependence of $\Lambda_n$ on the fixed time $t$ is suppressed in this notation.

Set
\begin{equation*}
     \widetilde m_n:=
        \left(\frac{P_\infty R_{X,n}^{d+3}E_n}{R_{X,n}^5E_n+R_{Y,n,t}^2}\frac{n}{\Lambda_n}\right)^{\gamma_d},
    \qquad m_n:=\lceil\widetilde m_n\rceil.
\end{equation*} 
Suppose that $m_n\leq n$, and let $\widehat f_{m_n}$ be a measurable empirical DSM minimizer over $\mathcal F_{m_n,M_n}$. Then, with probability at least $1-\delta$,
\begin{equation}
    \mathcal R_t(\widehat f_{m_n})
    \lesssim_d
    (P_\infty R_{X,n}^{d+3}E_n)^{\gamma_d}
    \left(\frac{R_{X,n}^5E_n+R_{Y,n,t}^2}{n}\Lambda_n\right)^{\beta_d}
    +\frac{(R_{X,n}^5E_n+R_{Y,n,t}^2)(\Lambda_n+\ell_{n,\delta})}{n}.
    \label{eq:compact subgaussian rate}
\end{equation}
\end{theorem}
\begin{proof}[Sketch of proof]
Proposition~\ref{prop:subG}  and Proposition~\ref{prop:tail control} control the truncation probability and tail contributions.
Substituting these bounds into Theorem~\ref{thm:approx-est tradeoff} with the
parameter choices yields the result. The details
are provided in Appendix~\ref{app:stats shallow}.
\end{proof}

\begin{remark}[The logarithmic factor]
\label{rmk:Lambda n log term}
The choice of path-norm budget in \eqref{eq:cap choice with n}, the target function bound in \eqref{eq:shallow truncation constants} and Corollary~\ref{cor:uniform pointwise score bound} give
\begin{equation*}
    M_n\asymp_dR_{X,n}^{3/2}E_n^{1/2},\qquad M_{R_{n,t}}\asymp_d R_{X,n}^{5/2}E_n^{1/2}.
\end{equation*}
Consequently, from the definition of $\Lambda_n=\Lambda_{n,R_{n,t}}$ in \eqref{eq:log factor defi}, we obtain
\begin{equation*}
    \Lambda_n\asymp_d 1+\log\left[ \sqrt n\max\{4,\,1+n/M_n+M_n\}\right].
\end{equation*} 

For fixed $d$, $\delta$,  and distributional parameters, $\Lambda_n=O(\log n)$. The dependence on the path-norm budget
remains inside this logarithmic factor.
\end{remark}

To make the dependence on the initial distribution explicit, set
\begin{equation*}
    e_{d,\kappa} := d+2(q_d+1)\max\{1,\kappa\},
    \qquad
    c_d := \gamma_d(d+3)+5\beta_d = \frac{(d+3)(d+5)}{2d+3}.
\end{equation*}
and abbreviate $\mathcal C_{\mathrm{reg}}(H,\kappa):= \mathcal C_{\mathrm{reg}}(H,\kappa;q_d,1/2)$ for the fixed pair $(H,\kappa)$. 
Corollary~\ref{cor:polynomial regularity envelope} gives
\begin{equation}
    \overline{\mathcal E}_{q_d,1/2,R_X}
    \leq\mathcal C_{\mathrm{reg}}(H,\kappa)R_X^{e_{d,\kappa}},
    \qquad R_X\geq1.
    \label{eq:regularity complexity explicit}
\end{equation}

\begin{cor}[Explicit dependence on the initial distribution]
\label{cor:explicit distributional rate}
Under the assumptions and parameter choices of Theorem~\ref{thm:concrete subgaussian rate}, define
\begin{equation}
     S_0:=1+K_0  ,\qquad  L_{n,\delta}:=6\sqrt dS_0\sqrt{\ell_{n,\delta}}.
    \label{eq:explicit radius envelope}
\end{equation}
Then $R_{X,n}\leq L_{n,\delta}$, and, with probability at least
$1-\delta$,
\begin{align}
    \mathcal R_t(\widehat f_{m_n})
    \lesssim_d{}&
    P_\infty^{\gamma_d}\left(\frac{\Lambda_n}{n}\right)^{\beta_d}
    \Bigl[
        \mathcal C_{\mathrm{reg}}(H,\kappa)L_{n,\delta}^{e_{d,\kappa}+c_d}
        \notag+\mathcal C_{\mathrm{reg}}(H,\kappa)^{\gamma_d}
        L_{n,\delta}^{\gamma_d(e_{d,\kappa}+d+3)}
        (\sigma_t^{-2}\ell_{n,\delta})^{\beta_d}
    \Bigr]\notag\\
    &+\frac{\Lambda_n+\ell_{n,\delta}}{n}
    \Bigl[
        \mathcal C_{\mathrm{reg}}(H,\kappa)L_{n,\delta}^{e_{d,\kappa}+5}
        +\sigma_t^{-2}\ell_{n,\delta}
    \Bigr].
    \label{eq:concrete subgaussian rate}
\end{align}
The implicit constant depends only on $d$. 
\end{cor}

\begin{remark}
    Under the assumptions and parameter choices of Theorem~\ref{thm:concrete subgaussian rate}, for every fixed $t>0$, with probability at least $1-\delta$,
    \begin{equation*}
        \mathcal R_t(\widehat f_{m_n})
    =
    \widetilde O\!\left(
        \frac{\sigma_t^{-2\beta_d}}{n^{\beta_d}}
        +
        \frac{\sigma_t^{-2}}{n}
    \right),
    \qquad
    \beta_d=\frac{d+3}{2d+3}.
    \end{equation*}
    Here, $\widetilde O$ suppresses logarithmic factors, with implicit
    constants depending on the fixed dimension, confidence level, and
    distributional parameters; the diffusion-time dependence is displayed
    explicitly. In particular, at any fixed $t>0$,
    \begin{equation*}
         \mathcal R_t(\widehat f_{m_n})
        =
        \widetilde O\!\left(n^{-\beta_d}\right).
    \end{equation*} 
\end{remark}

\subsection{Discussion: Dependence on the Data Distribution}

\paragraph{Distributional Dependence.}
For fixed $d$, $t>0$, $\delta$, and initial distribution $p_0$, together with a fixed admissible pair $(H,\kappa)$. The leading and remainder terms in \eqref{eq:concrete subgaussian rate} have orders $\widetilde O(n^{-\beta_d})$ and $\widetilde O(n^{-1})$, respectively.
The condition $m_n\leq n$ holds for all sufficiently large $n$ under these fixed-parameter asymptotics.

The distributional parameters enter the bound in distinct ways.
The density bound $P_\infty$ converts Lebesgue approximation error
to $L^2(P_t)$ error, the tail parameter $S_0$ controls the spatial
truncation radius, and $\mathcal C_{\mathrm{reg}}(H,\kappa)$ bounds
the growth of the Sobolev envelope with that radius. The exponent
$\kappa$ also enters through $e_{d,\kappa}$ and therefore affects
the powers of the logarithmic factors. These dependencies quantify
the derived upper bound; they do not imply an ordering of the
actual estimation risks across distributions.

\paragraph{Balancing Width and Regularity.}

The balanced width depends on the relative sizes of the two
contributions to the estimation coefficient: the network term
$R_{X,n}^5E_n$ and the label term $R_{Y,n,t}^2$.
\begin{itemize}
    \item When $R_{Y,n,t}^2\ll R_{X,n}^5E_n$, the network term dominates,
so $R_{X,n}^5E_n+R_{Y,n,t}^2\asymp R_{X,n}^5E_n$ and
\begin{equation*}
    \widetilde m_n\asymp_d
    \left(
        P_\infty R_{X,n}^{d-2}\frac{n}{\Lambda_n}
    \right)^{\gamma_d}.
\end{equation*}
The explicit factor $E_n$ cancels from the ratio of the
approximation and network-dependent estimation coefficients:
\begin{equation*}
    \frac{P_\infty R_{X,n}^{d+3}E_n}{R_{X,n}^5E_n}
    =P_\infty R_{X,n}^{d-2}.
\end{equation*}
The exact balanced width still depends on $E_n$ through the
relative size of the label term and the logarithmic factor
$\Lambda_n$.

The leading term in the risk bound satisfies
\begin{equation*}
    \left(P_\infty R_{X,n}^{d+3}E_n\right)^{\gamma_d}
    \left(
        \frac{(R_{X,n}^5E_n+R_{Y,n,t}^2)\Lambda_n}{n}
    \right)^{\beta_d} \asymp_d
    P_\infty^{\gamma_d}E_nR_{X,n}^{c_d}
    \left(\frac{\Lambda_n}{n}\right)^{\beta_d},
\end{equation*}
where we used $\gamma_d+\beta_d=1$ and
$c_d=\gamma_d(d+3)+5\beta_d$. Thus, the risk bound retains an
explicit factor $E_n$, and the required path-norm budget satisfies
$M_n\asymp_d R_{X,n}^{3/2}E_n^{1/2}$. The cancellation in the
width therefore does not remove the regularity dependence of
either the budget or the risk bound.
\item When $R_{Y,n,t}^2\gg R_{X,n}^5E_n$, the label term dominates,
so $R_{X,n}^5E_n+R_{Y,n,t}^2\asymp R_{Y,n,t}^2$ and
    \begin{equation*}
    \widetilde m_n\asymp_d
    \left(
        \frac{P_\infty R_{X,n}^{d+3}E_n}{R_{Y,n,t}^2}
        \frac{n}{\Lambda_n}
    \right)^{\gamma_d}.
    \end{equation*}
    The label term does not contain $E_n$, so no such cancellation
occurs, and the displayed width retains the explicit factor
$E_n^{\gamma_d}$.
\end{itemize}

\paragraph{Fixed-Time Asymptotics and Scope.}
Under the fixed-parameter asymptotics considered above, the prescribed radii satisfy $R_{X,n}\asymp(\log n)^{1/2}$ and $R_{Y,n,t}\asymp\sigma_t^{-1}(\log n)^{1/2}$.
Since the regularity envelope satisfies $E_n\geq R_{X,n}^d$,
\begin{equation*}
    \frac{R_{Y,n,t}^2}{R_{X,n}^5E_n}
    \leq
    \frac{R_{Y,n,t}^2}{R_{X,n}^{d+5}}
    \lesssim_d
    \frac{\sigma_t^{-2}}{(\log n)^{(d+3)/2}}
    \longrightarrow0
    \qquad\text{as }n\to\infty.
\end{equation*}
At any fixed diffusion time, the network term therefore dominates
for sufficiently large $n$ (the first case above). The label-dominated regime (the second case above), however, remains
relevant for finite samples or joint limits in which $t$ decreases
with $n$.

Although the path-norm budget $M_n$ is independent of $t$, the
statistical bound retains noise-dependent factors involving
$\sigma_t^{-2}$. A time-independent approximation budget therefore
does not make the score-risk bound uniform as $t\downarrow0$.
Finally, the choices of $M_n$ and $m_n$ depend on distributional parameters.

\section{Consequences for Reverse-Process Sampling}\label{sec:sampling}

We now translate the fixed-time score-estimation guarantees into
error bounds for a discretized reverse sampler. Applying the
score-to-sampling result of Benton et al.~\cite{benton2024nearly}
(Lemma~\ref{lem:score to sampling}) yields a KL divergence bound
between the generated distribution and the early-stopped target
$p_\gamma$. Combining this bound with an estimate of the
early-stopping error then establishes a total-variation guarantee
relative to the original data distribution.

Throughout this section, we fix the dimension $d$ and the initial
distribution $p_0$, and absorb the distributional
parameters from Section~\ref{sec:stats shallow} into the implicit
constants. The notation $\widetilde O_{d,p_0}$ suppresses
polylogarithmic factors in $nN/\delta$, while the dependence on the
terminal time $T$ and early-stopping time $\gamma$ is displayed
explicitly. We retain the notation $\beta_d=(d+3)/(2d+3)$ from
Section~\ref{sec:stats shallow}.

\paragraph{Sampler and Accumulated Score Error.}
Fix $T\geq1$ and $\gamma\in(0,1)$, and consider a reverse-time grid
\begin{equation*}
    0=u_0<u_1<\cdots<u_N=T-\gamma.
\end{equation*}
The corresponding forward times and step sizes are
\begin{equation*}
    \tau_k:=T-u_k,
    \qquad
    h_k:=u_{k+1}-u_k,
\end{equation*}
so that
\begin{equation*}
    T=\tau_0>\tau_1>\cdots>\tau_N=\gamma.
\end{equation*}

At each time $\tau_k$, $k=0,\ldots,N-1$, we train a separate fixed-time DSM estimator using the same $n$ original data
\begin{equation*}
    X_{0,1},\ldots,X_{0,n}
    \overset{\mathrm{i.i.d.}}{\sim}P_0.
\end{equation*}
We generate Gaussian corruptions
\begin{equation*}
    Z_{i,k}
    \overset{\mathrm{i.i.d.}}{\sim}
    \mathcal N(0,I_d),
    \qquad
    1\leq i\leq n,\quad 0\leq k\leq N-1,
\end{equation*}
independently of the original data, and construct the input--label
pairs
\begin{equation*}
    X_{\tau_k,i}
    :=
    e^{-\tau_k}X_{0,i}
    +
    \sigma_{\tau_k}Z_{i,k},
    \qquad
    Y_{\tau_k,i}
    :=
    -\frac{Z_{i,k}}{\sigma_{\tau_k}}.
\end{equation*}
Let $\widehat s_{\tau_k}$ denote the corresponding DSM estimator
of $s_{\tau_k}$. 

We initialize the approximate reverse process from the stationary
Gaussian distribution,
\begin{equation*}
    \widehat X_0\sim\mathcal N(0,I_d),
\end{equation*}
independently of the training data and Gaussian corruptions.
The mismatch between this initialization and the exact reverse
initial law $P_T$ is retained as a separate error term and
controlled under the moment assumption in the sampling theorem (Lemma~\ref{lem:score to sampling}) below.

On each interval $[u_k,u_{k+1}]$, the exponential-integrator sampler is defined by
\begin{equation*}
    \mathrm d\widehat X_u
    =
    \left[
        \widehat X_u
        +
        2\widehat s_{\tau_k}(\widehat X_{u_k})
    \right]\mathrm du
    +
    \sqrt{2}\,\mathrm d\widehat W_u,
    \qquad
    u\in[u_k,u_{k+1}],
\end{equation*}
where $(\widehat W_u)_{u\geq0}$ is a standard Brownian motion
independent of all training randomness. Equivalently,
\begin{equation*}
    \widehat X_{u_{k+1}}
    =
    e^{h_k}\widehat X_{u_k}
    +
    2(e^{h_k}-1)
    \widehat s_{\tau_k}(\widehat X_{u_k})
    +
    \sqrt{e^{2h_k}-1}\,\xi_k,
\end{equation*}
where $\xi_0,\ldots,\xi_{N-1}$ are independent standard Gaussian
vectors, independent of the initialization and training randomness.
We write $\widehat p_{\mathrm{out}}$ for the density of the output
$\widehat X_{T-\gamma}$.

At each time step $\tau_k$, define the accumulated score error by
\begin{equation*}
    \varepsilon_{\mathrm{score}}^2
    :=
    \sum_{k=0}^{N-1}
    h_k
    \mathbb E_{X\sim p_{\tau_k}}
    \left[
        \left\|
            \widehat s_{\tau_k}(X)
            -
            s_{\tau_k}(X)
        \right\|_2^2
    \right] =
    \sum_{k=0}^{N-1}
    h_k
    \mathcal R_{\tau_k}
    \left(
        \widehat s_{\tau_k}
    \right).
\end{equation*}

\paragraph{Score-to-sampling Bound.}

We first state the sampling result of \cite{benton2024nearly} in our notation.

\begin{lem}[Score-to-sampling bound;
{\cite[Theorem 2 and Corollary 1]{benton2024nearly}}]
\label{lem:score to sampling}
Suppose that the initial distribution has finite second moment
\begin{equation*}
    M_2
    :=
    \mathbb E\|X_0\|_2^2
    <
    \infty.
\end{equation*}
Let $T\geq1$, $\gamma\in(0,1)$, and $N\geq\log(\gamma^{-1})$. There exists a reverse-time grid $ 0=u_0<u_1<\cdots<u_N=T-\gamma$ and a parameter
\begin{equation*}
    \overline h_N
    \asymp
    \frac{T+\log(\gamma^{-1})}{N}
\end{equation*}
such that
\begin{equation*}
    h_k
    \leq
    \overline h_N
    \min\{1,\tau_{k+1}\},
    \qquad
    k=0,\ldots,N-1.
\end{equation*}
For this grid, the output density of the exponential-integrator
sampler satisfies
\begin{align*}
    \operatorname{KL}
    \left(
        p_\gamma
        \,\middle\|\,
        \widehat p_{\mathrm{out}}
    \right)
    \lesssim{}&
    \varepsilon_{\mathrm{score}}^2
    +
    \overline h_N^2dN
    +
    \overline h_N(dT+M_2)
    +
    (d+M_2)e^{-2T}. 
\end{align*}
\end{lem}

It remains to bound $\varepsilon_{\mathrm{score}}^2$ using the
fixed-time statistical guarantees from
Section~\ref{sec:stats shallow}.

\begin{prop} \label{prop:accumulated score error}
Suppose that the assumptions of Corollary~\ref{cor:explicit distributional rate} hold. 
At each time $\tau_k$, let $\widehat s_{\tau_k}$ be the estimator from Corollary~\ref{cor:explicit distributional rate},  with its parameter choices and confidence parameter $\delta/N$. 
Let $m_{n,k}$ denote the balanced network width at time $\tau_k$, and assume that $m_{n,k}\leq n$ for all $k=0,\ldots,N-1$. Then, with probability at least $1-\delta$,
\begin{align}
    \varepsilon_{\mathrm{score}}^2
    =
    \sum_{k=0}^{N-1}
    h_k
    \mathcal R_{\tau_k}
    \left(
        \widehat s_{\tau_k}
    \right)
    =\widetilde{O}_{d,p_0}\left(\frac{T}{n^{\beta_d}}+\frac{T+\log(\gamma^{-1})}{n}\right)
    \label{eq:accumulated score error}
\end{align}
\end{prop}
\begin{proof}
Applying Corollary~\ref{cor:explicit distributional rate} at each
$\tau_k$ with confidence parameter $\delta/N$. By the union bound, with probability at least $1-\delta$,
\begin{equation}
    \label{eq:fix time risk bound}
    \mathcal R_{\tau_k}(\widehat s_{\tau_k})
    \leq
    C\left(1+\log\frac{nN}{\delta}\right)^A
    \left(
        \frac{\sigma_{\tau_k}^{-2\beta_d}}{n^{\beta_d}}
        +\frac{\sigma_{\tau_k}^{-2}}{n}
    \right)
\end{equation}
holds simultaneously for every $k=0,\ldots,N-1$, where $C>0$
and $A\geq0$ depend only on $d$ and the fixed initial distribution.

For $a>0$, the function $t\mapsto\sigma_t^{-2a}$ is decreasing.
Therefore,
\[
    \sum_{k=0}^{N-1}h_k\sigma_{\tau_k}^{-2a}
    \leq
    \int_\gamma^T\sigma_t^{-2a}\,\mathrm dt
    =:J_a(\gamma,T).
\]
Using $0<\beta_d<1$ and $T\geq1$, we obtain
\[
    J_{\beta_d}(\gamma,T)\lesssim_d T,
    \qquad
    J_1(\gamma,T)
    =\frac12\log\left(
        \frac{e^{2T}-1}{e^{2\gamma}-1}
    \right)
    \lesssim T+\log(\gamma^{-1}).
\]
Multiplying \eqref{eq:fix time risk bound} by $h_k$, summing
over $k$, and applying these integral estimates proves
\eqref{eq:accumulated score error}.
\end{proof}

For fixed $T$, the time integral associated with the leading
statistical term remains bounded as $\gamma\downarrow0$, whereas
the integral associated with the $n^{-1}$ remainder grows only
logarithmically in $\gamma^{-1}$.
Substituting \eqref{eq:accumulated score error} into
Lemma~\ref{lem:score to sampling} yields the following KL bound.

\begin{cor}[Sampling consequence of statistical score estimation]
\label{cor:sampling consequence}
Under the assumptions of Lemma~\ref{lem:score to sampling} and
Proposition~\ref{prop:accumulated score error}, with probability at
least $1-\delta$,
\begin{align*}
    \operatorname{KL}
    \left(
        p_\gamma
        \,\middle\|\,
        \widehat p_{\mathrm{out}}
    \right)=\widetilde O_{d,p_0}\!\left(
\frac{ T}{n^{\beta_d}}
+{\frac{T+\log(\gamma^{-1})}{n}}
+\left(\frac{T+\log(\gamma^{-1})}{\sqrt N}\right)^2
+e^{-2T}
\right)
\end{align*}
\end{cor}

\paragraph{Early Stopping and End-to-end Guarantee.}
To compare the generated distribution with the original data
distribution, we also control the total-variation distance between
$p_\gamma$ and $p_0$.

\begin{prop}[Early-stopping error]
\label{prop: early stopping error}
Suppose that Assumptions~\ref{asm:p0 reg}
and~\ref{asm:poly growth} hold for some integer $r\geq1$,
and that Assumption~\ref{asm:subg} holds.
Then, for every $\gamma>0$,
\begin{equation*}
    \mathrm{TV}(p_0,p_\gamma)
    \leq
    \frac{\gamma}{2}
    \left\|
        \Delta p_0+\nabla\cdot(xp_0)
    \right\|_{L^1(\RR^d)}\lesssim_{p_0}\gamma.
\end{equation*}
\end{prop}

\begin{remark}
    Early-stopping bias has been controlled in Wasserstein
distance~\cite{debortoli2022convergence} and, under density
regularity assumptions, in total
variation~\cite{oko2023diffusion,zhang2024minimax}.
Under our assumptions, the $L^1$-contraction property of the
forward OU semigroup yields a bound linear in $\gamma$;
see \cite[Lemma~1.3]{engel2000one} and
\cite{kloeden1992numerical} for the semigroup and diffusion tools
used in the argument. The proof, including an explicit bound on
the coefficient, is given in Appendix~\ref{appendix:sampling}.
\end{remark}
The triangle inequality and Pinsker's inequality give
\begin{equation*}
    \operatorname{TV}(p_0,\widehat p_{\mathrm{out}})
    \leq\operatorname{TV}(p_0,p_\gamma)
    +\operatorname{TV}(p_\gamma,\widehat p_{\mathrm{out}})
    \leq\operatorname{TV}(p_0,p_\gamma)
    +\sqrt{
        \operatorname{KL}
        \left(p_\gamma\,\middle\|\,\widehat p_{\mathrm{out}}\right)
    }.
\end{equation*}
Combining this with
Corollary~\ref{cor:sampling consequence} and
Proposition~\ref{prop: early stopping error} yields the
end-to-end  total variation bound.
\begin{cor}\label{cor:endtoend TV bound}
    Suppose that the assumptions of Corollary~\ref{cor:sampling consequence} and Proposition~\ref{prop: early stopping error} hold.
    Then, with probability at least $1-\delta$,
    \begin{align*}
    \mathrm{TV}(p_0,\widehat p_{\mathrm{out}})
    =&\widetilde O\!\left(
\gamma
+\frac{\sqrt T}{n^{\beta_d/2}}
+\sqrt{\frac{T+\log(\gamma^{-1})}{n}}
+\frac{T+\log(\gamma^{-1})}{\sqrt N}
+e^{-T}
\right).
\end{align*}

In particular, for any fixed $\delta\in(0,1)$ and all sufficiently large $n$, choosing
\begin{equation*}
    \gamma =n^{-\beta_d/2}, 
    \qquad
    T =1+\frac{\beta_d}{2}\log n,
    \qquad
    N =\left\lceil n^{\beta_d}\log^2(en)\right\rceil
\end{equation*}
yields 
\begin{equation*}
     \operatorname{TV}(p_0,\widehat p_{\mathrm{out}})=\widetilde O\!\left(n^{-\beta_d/2}\right)=\widetilde O\!\left(n^{-\frac{d+3}{4d+6}}\right).
\end{equation*}
\end{cor}

\section{Conclusion}

We have established finite-sample guarantees for estimating
Ornstein--Uhlenbeck diffusion scores using finite-width,
path-norm-constrained shallow ReLU networks. The analysis connects
the regularity and tail properties of the initial distribution
to approximation error, sufficient network capacity, and
statistical accuracy.

The analytic foundation consists of two complementary local
Sobolev bounds: a kernel-induced estimate valid for arbitrary
initial laws and a data-induced estimate based on normalized
weak derivatives of the initial density. Under the stated
regularity and growth assumptions, their minimum yields a
time-uniform envelope on each fixed ball. This envelope provides
the quantitative norm control needed for shallow ReLU
approximation.

For direct-label DSM, we combine truncation with localized
empirical-process analysis to handle unbounded inputs and labels
and input-dependent regression residuals. The resulting
high-probability bounds track the effects of initial-density
regularity, density bounds, and tail behavior. For fixed
dimension $d$, positive diffusion time, confidence level, and initial
distribution, the squared score-estimation risk is
$\widetilde O(n^{-(d+3)/(2d+3)})$. The explicit time dependence of these bounds allows us to control the score-estimation error accumulated over a reverse-time grid. Combining this estimate with the sampling theorem and an early-stopping bound yields an end-to-end total-variation rate of $\widetilde O(n^{-(d+3)/(4d+6)})$ for suitable choices of the sampling parameters.

\section*{Acknowledgement}
The research of Y. Yang was supported by the Simons Foundation International through Grant No.~SFI-MPS-SEFF-00021457. The research of F. Liu and W. Zhu was supported by the U.S. National Science Foundation under Awards No.~DMS-2502900 and DMS-2540370, and by the Air Force Office of Scientific Research (AFOSR) under Grant No.~FA9550-25-1-0079.

\bibliographystyle{alpha} 
\bibliography{ref.bib}

\section*{Appendix}
\appendix
\section{Proofs for Section~\ref{sec:reg}}
\label{app:reg}
\paragraph{Proof of Theorem~\ref{thm:kernel-induced-regularity}:}
\begin{proof}
    Fix $x\in\Omega_R$ and a multi-index $\nu\in\mathbb N^d$ satisfying $1\leq|\nu|\leq r+1$.
    Then,
    \begin{align}
        Q_{\nu,t}(x)
        &=
        \frac{
            \int_{\mathbb R^d}
            \partial_x^\nu g_t(x-e^{-t}y)\,P_0(\mathrm dy)
        }{p_t(x)}
        \notag\\
        &=
        (-1)^{|\nu|}\sigma_t^{-|\nu|}
        \int_{\mathbb R^d}
        H_\nu\left(
            \frac{x-e^{-t}y}{\sigma_t}
        \right)
        \Pi_{t,x}(\mathrm dy).
        \label{eq:kir-hermite-representation}
    \end{align}
    Set
    \begin{equation*}
        W_{t,x}(y)
        :=
        \frac{x-e^{-t}y}{\sigma_t}.
    \end{equation*}

    Consequently, \eqref{eq:kir-hermite-representation} becomes
    \begin{equation}\label{eq:kir-posterior-hermite}
        Q_{\nu,t}(x)=(-1)^{|\nu|}\sigma_t^{-|\nu|}\mathbb E \left[H_\nu\left(W_{t,x}\right)|X_t=x\right].
    \end{equation} 

    For the Hermite polynomials, we have the standard polynomial growth bound
    \begin{equation*}
        \left|\mathrm{He}_n(x)\right|=\left|n!\sum_{r=0}^{\lfloor n/2 \rfloor}\frac{(-1)^r}{2^rr!(n-2r)!}x^{n-2r}\right|\leq C_n(1+|x|)^n.
    \end{equation*}
    Further, 
    \begin{equation*}
        |H_\nu(w)|\leq \prod_{j=1}^dC_{\nu_j}(1+|w_j|)^{\nu_j}\leq \prod_{j=1}^dC_{\nu_j}(1+\|w\|)^{\nu_j}\leq C_\nu(1+\|w\|^{|\nu|}).
    \end{equation*}

    Therefore, combining this estimation with \eqref{eq:kir-posterior-hermite}, we obtain
    \begin{equation}\label{eq:kir-Q-via-moment}
        |Q_{\nu,t}(x)|\leq C_\nu \sigma_t^{-|\nu|}\left(1+\mathbb E\left[\|W_{t,x}\|^{|\nu|}|X_t=x\right]\right).
    \end{equation}

    We next bound the posterior moments of $W_{t,x}$ uniformly over $x\in\Omega_R$.

    First, for every $u\geq 0$
    \begin{align*}
        &\int_{\{\|W_{t,x}(y)\|_2\geq u\}}
        g_t(x-e^{-t}y)\,P_0(\mathrm dy)
        \\
        &= (2\pi\sigma_t^2)^{-\frac{d}{2}}\int_{\{y:\|W_{t,x}(y)\|\geq u\}}\exp\left(-\frac{\|W_{t,x}(y)\|^2}{2}\right)P_0(\mathrm dy)\\
        &\leq
        (2\pi\sigma_t^2)^{-d/2}e^{-u^2/2}.
    \end{align*}
    
    On the other hand, using the definiton \eqref{eq:quantile radius} and $\|x\|_2\le R$, we obtain\begin{align*}
        p_t(x)&=\int_{\RR^d}g_t(x-e^{-t}y)P_0(\mathrm dy)\\
        &\geq
        \int_{\{\|y\|_2\leq B_\rho\}}
        g_t(x-e^{-t}y)\,P_0(\mathrm dy)
        \\
        &\geq
        \rho(2\pi\sigma_t^2)^{-d/2}
        \exp\left(
            -\frac{(R+e^{-t}B_\rho)^2}{2\sigma_t^2}
        \right).
    \end{align*}

   Combining these two bounds yields \begin{equation}\label{eq:kir-posterior-tail}
        \Pi_{t,x}\bigl(\|W_{t,x}\|_2\geq u\bigr)
        \leq
        \min\left\{
            1,\,
            \rho^{-1}
            \exp\left(
                \frac{(R+e^{-t}B_\rho)^2}{2\sigma_t^2}
                -\frac{u^2}{2}
            \right)
        \right\}.
    \end{equation}
    Define 
    $$
    u_0:=\sqrt{\frac{(R+e^{-t}B_\rho)^2}{\sigma_t^2}+2\log \rho^{-1}}\leq \frac{R+e^{-t}B_\rho}{\sigma_t}+\sqrt{2\log \rho^{-1}}.
    $$ 
    
    Then \eqref{eq:kir-posterior-tail} becomes
    \begin{equation}\label{eq:kir-posterior-tail-new}
        \Pi_{t,x}\left(\|W_{t,x}\|\geq u\right)\leq \min\left\{1, \exp\left(\frac{u_0^2-u^2}{2}\right)\right\}.
    \end{equation}
    It follows from \eqref{eq:kir-posterior-tail-new} that, for every $v>0$,
    \begin{equation*}
        \Pi_{t,x}\left(\left(\|W_{t,x}\|-u_0\right)_+\geq v\right)=\Pi_{t,x}\left(\|W_{t,x}\|\geq u_0+v\right)\leq \exp\left(-\frac{v^2}{2}\right).
    \end{equation*}
    Therefore, by the layer-cake identity, for every $q\ge 1$
    \begin{align*}
        \mathbb E\left[(\|W_{t,x}\|-u_0)_+^q|X_t=x\right]
        &=q\int_0^\infty v^{q-1}\Pi_{t,x}\left(\left(\|W_{t,x}\|-u_0\right)_+\geq v\right)\mathrm dv\\
        &\leq q\int_0^\infty v^{q-1}e^{-\frac{v^2}{2}}\mathrm dv\\
        &=q2^{\frac{q}{2}-1}\int_{0}^\infty u^{\frac{q}{2}-1}e^{-u}\mathrm du\\
        &=2^{\frac{q}{2}}\Gamma\left(\frac{q}{2}+1\right)\leq (2\sqrt{q})^q.
    \end{align*}

    Since $\|W_{t,x}\|\leq u_0+(\|W_{t,x}\|-u_0)_+$, by Minkowski's inequality, we have
    \begin{align*}
        \mathbb E^{1/q}\left[\|W_{t,x}\|^q|X_t=x\right]
        &\leq \mathbb E^{1/q}\left[(u_0+(\|W_{t,x}\|-u_0)_+)^q|X_t=x\right]\\
        &\leq u_0+\mathbb E^{1/q}\left[(\|W_{t,x}\|-u_0)_+^q|X_t=x\right]\\
        &\leq u_0+2\sqrt{q}.
    \end{align*}

    Then,
    \begin{align}
        \mathbb E\left[\|W_{t,x}\|^q|X_t=x\right] 
        &\leq \left(u_0+2\sqrt{q}\right)^q\notag\\
        &\leq \left( \frac{R+e^{-t}B_\rho}{\sigma_t}+\sqrt{2\log\rho^{-1}}+2\sqrt{q}\right)^q\label{eq:kir-posterior-moment}.
    \end{align}

    As a result, combining \eqref{eq:kir-Q-via-moment} with \eqref{eq:kir-posterior-moment} yields
    \begin{align*}
        |Q_{\nu,t}(x)|
        &\leq C_\nu \sigma_t^{-|\nu|}\left(1+\mathbb E\left[\|W_{t,x}\|^{|\nu|}|X_t=x\right]\right)\\
        &\leq C_\nu \sigma_t^{-|\nu|}\left(1+\left( \frac{R+e^{-t}B_\rho}{\sigma_t}+\sqrt{2\log\rho^{-1}}+2\sqrt{|\nu|}\right)^{|\nu|}\right)\\
        &\leq C_\nu \sigma_t^{-|\nu|}\left(1+ \frac{R+e^{-t}B_\rho}{\sigma_t}+\sqrt{2\log\rho^{-1}}+2\sqrt{|\nu|}\right)^{|\nu|}.
    \end{align*}
    
    For $k\geq |\nu|$, we define 
    \begin{equation*}
        \Lambda_k:=1+ \frac{R+e^{-t}B_\rho}{\sigma_t}+\sqrt{2\log\rho^{-1}}+2\sqrt{k},
    \end{equation*}
    then 
    \begin{equation}\label{eq:kir-normalized-derivative}
        |Q_{\nu,t}(x)|\leq C_\nu\sigma_t^{-|\nu|}\Lambda_k^{|\nu|}.
    \end{equation}

    We now pass from normalized density derivatives to derivatives of the score. For a multi-index $\delta$, let $\mathfrak P(\delta)$ denote the set of partitions of the labelled derivative slots associated with $\delta$. Each block $B$ induces a multi-index $\nu(B)$, and we write $Q_{B,t}:=Q_{\nu(B),t}$. We use the usual conventions $\mathfrak P(0)=\{\varnothing\}$ and $\prod_{B\in\varnothing}Q_{B,t}=1$.
    
    By definition of the Sobolev norm and the multivariate Faa di Bruno formula, we obtain
    \begin{align}
        \|s_{t,i}\|_{H^r(\Omega_R)}^2 &=\sum_{|\gamma|\leq r }\int_{\Omega_R}|\partial^\gamma s_{t,i}(x)|^2\mathrm d x\notag\\
        &= \sum_{|\gamma|\leq r }\int_{\Omega_R}\left| \sum_{\beta\leq \gamma }\sum_{\pi\in\mathfrak P(\gamma-\beta)}\binom{\gamma}{\beta}(-1)^{|\pi|}|\pi|!{Q_{\beta+e_i,t}(x)}\prod_{B\in\pi}{Q_{B,t}(x)}\right|^2\mathrm dx\notag\\
        &\leq C_r\sum_{|\gamma|\leq r }\sum_{\beta\leq \gamma }\sum_{\pi\in\mathfrak P(\gamma-\beta)}\int_{\Omega_R}\binom{\gamma}{\beta}^2(|\pi|!)^2\left|{Q_{\beta+e_i,t}(x)}\prod_{B\in\pi}{Q_{B,t}(x)}\right|^2\mathrm dx\label{eq:kir-score-expansion}.
    \end{align}

    For every summand in \eqref{eq:kir-score-expansion}, it follows from \eqref{eq:kir-normalized-derivative} that
    \begin{align}
        \left|{Q_{\beta+e_i,t}(x)}\prod_{B\in\pi}{Q_{B,t}(x)}\right|^2
        \leq \left(C_{\beta+e_i}^2\prod_{B\in\pi}C_{\nu(B)}^2\right) \sigma_t^{-2(|\gamma|+1)}\Lambda_{|\gamma|+1}^{2(|\gamma|+1)}\label{eq:kir-score-derivative}.
    \end{align}
    
    Using $|\Omega_R|=\omega_dR^d$ and \eqref{eq:kir-score-derivative}, we conclude that 
    \begin{align*}
        \|s_{t,i}\|_{H^r(\Omega_R)}^2
        &\leq C_{d,r}R^d\sum_{k=1}^{r+1}\sigma_t^{-2k}\Lambda_k^{2k}\\
        &\leq C_{d,r}R^d\sigma_t^{-2r-2}\Lambda_{r+1}^{2r+2}\\
        &\leq  C_{d,r}R^d\sigma_t^{-2r-2}\left(1+ \frac{R+e^{-t}B_\rho}{\sigma_t}+\sqrt{2\log\rho^{-1}}\right)^{2r+2}.
        \end{align*}
        This proves \eqref{eq:kernel-sobolev-main}.

        Indeed, we further have that  
        \begin{align*}
             \|s_{t,i}\|_{H^r(\Omega_R)}^2
        &\leq C_{d,r}R^d\sigma_t^{-2r-2}\left(1+ \frac{R+e^{-t}B_\rho}{\sigma_t}+\sqrt{2\log\rho^{-1}}\right)^{2r+2}.\\
        &\leq C_{d,r}R^d\sigma_t^{-2r-2}\left(1+(2\log\rho^{-1})^{r+1}+\frac{(R+e^{-t}B_\rho)^{2r+2}}{\sigma_t^{2r+2}}\right)\\
        &\leq C_{d,r}\sigma_t^{-4r-4}[(1+(2\log\rho^{-1})^{r+1}+B_\rho^{2r+2})R^d+R^{d+2r+2}],
        \end{align*}
        which proves \eqref{eq:kernel-sobolev-expanded}.
\end{proof}

\paragraph{Proof of Proposition~\ref{prop:absolute continuity}:}

\begin{lem}\label{lem:p0 locality}
   Let $k\in\mathbb N$. For every measurable set $E\subset\RR^d$ and every real-valued function $u\in W^{k,1}(\RR^d)$ satisfying $u=0$ almost everywhere on $E$, we have
\begin{equation*}
    \partial^\nu u=0
    \qquad\text{for a.e. }x\in E,
    \qquad 1\le|\nu|\le k.
\end{equation*}
\end{lem}
\begin{proof}
   We proceed by induction on $k$.

For $k=1$, the assumption that $u=0$ almost everywhere on $E$
means that
\begin{equation*}
  | E\setminus\{u=0\}|=0.
\end{equation*}
By \cite[Theorem~6.19]{lieb2001analysis},
$\nabla u=0$ almost everywhere on $\{u=0\}$. Therefore,
\begin{equation*}
    \nabla u=0
    \qquad\text{for a.e. }x\in E.
\end{equation*}
Thus, the claim holds for $k=1$.

Suppose that the claim holds for some $k\ge1$, and let
$u\in W^{k+1,1}(\RR^d)$ satisfy $u=0$ almost everywhere on $E$.
Since $u\in W^{k,1}(\RR^d)$, the induction hypothesis gives
\begin{equation*}
    \partial_i u=0
    \qquad\text{for a.e. }x\in E,
    \qquad i=1,\ldots,d.
\end{equation*}
Moreover,
\begin{equation*}
    \partial_i u\in W^{k,1}(\RR^d),
    \qquad i=1,\ldots,d.
\end{equation*}
 Therefore we can apply the induction hypothesis to the function
$\partial_i u$, which yields
\begin{equation*}
    \partial^\beta\partial_i u=0
    \qquad\text{for a.e. }x\in E,
    \qquad 1\le|\beta|\le k.
\end{equation*}

For every multi-index $\nu$ satisfying $|\nu|=k+1$, choose
$i$ such that $\nu_i\ge1$ and write
\begin{equation*}
    \nu=\beta+e_i,
    \qquad |\beta|=k.
\end{equation*}
It follows that
\begin{equation*}
    \partial^\nu u
    =
    \partial^\beta\partial_i u
    =
    0
    \qquad\text{for a.e. }x\in E.
\end{equation*}
The derivatives of order at most $k$ are already covered by the
induction hypothesis. Hence, the claim holds for $k+1$.
\end{proof}
\begin{proof}
    [Proof of Proposition~\ref{prop:absolute continuity}]

    Let
\begin{equation*}
    Z_0:=\{x\in\RR^d:p_0(x)=0\}.
\end{equation*}

    Under Assumption~\ref{asm:p0 reg}, we have  $p_0\in W^{r+1,1}(\RR^d)$ and $p_0=0$ on $Z_0$. Let
    \begin{equation*}
        u=p_0,
        \qquad
        E=Z_0,
        \qquad
        k=r+1.
    \end{equation*}
    Therefore by Lemma~\ref{lem:p0 locality}, we conclude that
    \begin{equation}\label{eq:p0 locality}
        \partial^\nu p_0(x)=0
        \qquad\text{for a.e. }x\in Z_0,
        \qquad 1\le|\nu|\le r+1.
    \end{equation}

    For any Borel set $A$ such that 
    \begin{equation*}
        P_0(A)=\int_A p_0(x)\mathrm dx= 0,
    \end{equation*}
    we have that 
    \begin{equation*}
        p_0(x)= 0 \qquad a.e.~ x\in A.
    \end{equation*}

   By \eqref{eq:p0 locality}, we have 
   \begin{equation*}
       \partial^\nu p_0(x)=0
        \qquad\text{for a.e. }x\in A,
        \qquad 1\le|\nu|\le r+1.
   \end{equation*}
    Moreover,
    \begin{equation*}
        \int_A\partial^\nu p_0(x)\mathrm dx=0.
    \end{equation*}

    Thus 
    \begin{equation*}
        \partial^\nu p_0(x)\mathrm dx\ll p_0(x)\mathrm dx.
    \end{equation*}
    By Radon-Nikodym Theorem, there exist $h_\nu\in L^1(p_0)$ such that 
    \begin{equation*}
        \partial^\nu p_0(x) =h_\nu(x) p_0(x)\qquad a.e. ~x\in \RR^d.
    \end{equation*}
    
    Indeed, we can define 
    \begin{equation*}
        h_\nu(x):=
        \begin{cases} 
        \dfrac{\partial^\nu p_0(x)}{p_0(x)}, & p_0(x)>0,\\
        0, & p_0(x)=0.
        \end{cases}
    \end{equation*}
    Then 
     \begin{equation*}
        \partial^\nu p_0(x) =h_\nu(x) p_0(x)\qquad a.e. ~x\in \RR^d.
    \end{equation*}
    Moreover,
    \begin{equation*}
        \|h_\nu\|_{L^1(p_0)}=\int_{\RR^d}\left|\frac{\partial^\nu p_0}{p_0}\right|p_0\mathrm dx= \|\partial^\nu p_0\|_{L^1(\RR^d)}<\infty.
    \end{equation*}

    This completes the proof.
\end{proof}

\paragraph{Proof of Theorem~\ref{thm:data-induced-regularity}:}
\begin{proof}
    [proof of Theorem~\ref{thm:data-induced-regularity}]
    Recall that 
    \begin{equation*}
        W_{t,x}(X_0)=\frac{x-e^{-t}X_0}{\sigma_t},
    \end{equation*}
    then 
    \begin{equation*}
        X_0=e^t(x-\sigma_tW_{t,x}(X_0)).
    \end{equation*}
     For any $r\geq 1$, by Minkowski's inequality and previous estimate \eqref{eq:kir-posterior-moment}, we have 
    \begin{align}
        \mathbb E^{1/r}[\|X_0\|^r\mid X_t=x]
        &=e^t\mathbb E^{1/r}\left[\|x-\sigma_tW_{t,x}(X_0)\|^r\mid X_t=x\right]\notag\\
        &\leq e^t\left(\|x\|+\sigma_t\mathbb E^{1/r}[\|W_{t,x}\|^r\mid X_t=x]\right)\notag\\
        &\leq e^t\left( 2R+e^{-t}B_\rho+\sigma_t(\sqrt{2\log \rho^{-1}}+2\sqrt{r})\right).\label{eq:dir moment est}
    \end{align}

    Let $r_\nu:=\kappa|\nu|\vee 1$.
    For $1\leq |\nu|\leq r+1$, by Assumption~\ref{asm:poly growth},  \eqref{eq:data-side posterior representation}, and \eqref{eq:dir moment est}, we have the following estimate
    \begin{align}
        |Q_{\nu,t}(x)|
        &\leq e^{t|\nu|}H^{|\nu|}\mathbb E[(1+\|X_0\|)^{\kappa|\nu|}\mid X_t=x]\notag\\
        &\leq e^{t|\nu|}H^{|\nu|}\left(\mathbb E[(1+\|X_0\|)^{r_\nu}\mid X_t=x]\right)^{\kappa|\nu|/r_\nu}\notag\\
        &\leq e^{t|\nu|}H^{|\nu|}\left(1+\mathbb E^{1/r_\nu}[\|X_0\|^{r_\nu}\mid X_t=x]\right)^{\kappa|\nu|}\notag\\
        &\leq e^{t|\nu|}H^{|\nu|}\left( 1+2e^tR+B_\rho+e^t\sigma_t(\sqrt{2\log\rho^{-1}}+2\sqrt{r_\nu})\right)^{\kappa|\nu|}\notag\\
        &\leq  e^{t|\nu|}H^{|\nu|}\mathcal M_{t,\rho,R}^{\kappa|\nu|},\label{eq:dir Q est}
    \end{align}
    where we have used the monotonicity of $L^p$ norms in our second step and the inequality $r_\nu=\max\{1,\kappa|\nu|\}\le 1+\kappa(r+1)$ in our last step.
    
    For every summand in \eqref{eq:kir-score-expansion}, it follows from \eqref{eq:dir Q est} that
     \begin{align}
        \left|{Q_{\beta+e_i,t}(x)}\prod_{B\in\pi}{Q_{B,t}(x)}\right|
        &\leq (e^tH\mathcal M_{t,\rho,R}^\kappa)^{|\beta|+1}\prod_{B\in\pi}(e^tH\mathcal M_{t,\rho,R}^\kappa)^{|B|}\notag\\
        &\leq (e^tH\mathcal M_{t,\rho,R}^\kappa)^{|\gamma|+1}\notag.
    \end{align}
    Further, 
    \begin{align*}
        |\partial^\gamma s_{t,i}|
        &\leq \sum_{\beta\leq \gamma}\sum_{\pi\in \mathfrak P(\gamma-\beta)}\binom{\gamma}{\beta}|\pi|! \left|{Q_{\beta+e_i,t}(x)}\prod_{B\in\pi}{Q_{B,t}(x)}\right|\\
        &\leq C_{\gamma} (e^tH\mathcal M_{t,\rho,R}^\kappa)^{|\gamma|+1}.
    \end{align*}
    
    Hence, we conclude that 
    \begin{align*}
        \|s_{t,i}\|_{H^r(\Omega_R)}^2 
        &= \sum_{|\gamma|\leq r}\int_{\Omega_R} |\partial^\gamma s_{t,i}|^2\mathrm dx\\
        &\leq \omega_dR^d\sum_{|\gamma|\leq r}C_\gamma^2(e^tH\mathcal M_{t,\rho,R}^\kappa)^{2(|\gamma|+1)}\\
        &\leq C_{d,r}R^d(e^tH)^{2(r+1)}\mathcal{M}_{t,\rho,R}^{2\kappa(r+1)}.
    \end{align*}
    Taking the maximum over $1\le i\le d$ proves \eqref{eq:data-induced-regularity}.

\end{proof}

\paragraph{Proof of Corollary~\ref{cor:regularity envelope}:}
\begin{proof}
Under the assumptions of the corollary, Theorem~\ref{thm:kernel-induced-regularity} gives
\begin{equation*}
    \max_{1\leq i\leq d}
    \|s_{t,i}\|_{H^r(\Omega_R)}^2
    \leq
    C_{d,r}\mathcal K_{r,\rho,R}(t),
\end{equation*}
whereas Theorem~\ref{thm:data-induced-regularity} gives
\begin{equation*}
    \max_{1\leq i\leq d}
    \|s_{t,i}\|_{H^r(\Omega_R)}^2
    \leq
    C_{d,r}\mathcal D_{r,\rho,R}(t).
\end{equation*}
Taking the minimum of the two bounds proves \eqref{eq:regularity-envelope}.

It remains to prove the time-uniform assertion. 
Since $\sigma_t^2=1-e^{-2t}$, the kernel-induced bound can be written as
\begin{equation*}
    \mathcal K_{r,\rho,R}(t)
    =
    R^d
    \left[
        \frac{1+\sqrt{2\log (\rho^{-1})}}{\sigma_t}
        +
        \frac{R+e^{-t}B_\rho}{\sigma_t^2}
    \right]^{2(r+1)}.
\end{equation*}
Hence $t\mapsto\mathcal K_{r,\rho,R}(t)$ is continuous and strictly
decreasing.

On the other hand,
\begin{equation*}
    \mathcal M_{t,\rho,R}=1+B_\rho+2Re^t+\left( \sqrt{2\log(\rho^{-1})} +2\sqrt{1+\kappa(r+1)} \right)\sqrt{e^{2t}-1},
\end{equation*}
so $t\mapsto\mathcal M_{t,\rho,R}$ and hence $t\mapsto\mathcal D_{r,\rho,R}(t)$ are continuous and strictly increasing. Moreover,
\begin{equation*}
\begin{aligned}
    \lim_{t\downarrow0}
    \mathcal K_{r,\rho,R}(t)
    &=\infty,
    &
    \lim_{t\downarrow0}
    \mathcal D_{r,\rho,R}(t)
    &<\infty,
    \\
    \lim_{t\to\infty}
    \mathcal K_{r,\rho,R}(t)
    &<\infty,
    &
    \lim_{t\to\infty}
    \mathcal D_{r,\rho,R}(t)
    &=\infty.
\end{aligned}
\end{equation*}
The intermediate value theorem and strict monotonicity therefore give a unique $t_{r,\rho,R}^*>0$ such that
\begin{equation*}
    \mathcal K_{r,\rho,R}(t_{r,\rho,R}^*)
    =
    \mathcal D_{r,\rho,R}(t_{r,\rho,R}^*).
\end{equation*}

For $0<t\leq t_{r,\rho,R}^*$, we have
\begin{equation*}
    \mathcal E_{r,\rho,R}(t)
    =
    \mathcal D_{r,\rho,R}(t)
    \leq
    \mathcal D_{r,\rho,R}(t_{r,\rho,R}^*).
\end{equation*}
For $t\geq t_{r,\rho,R}^*$, we have
\begin{equation*}
    \mathcal E_{r,\rho,R}(t)
    =
    \mathcal K_{r,\rho,R}(t)
    \leq
    \mathcal K_{r,\rho,R}(t_{r,\rho,R}^*).
\end{equation*}
Therefore,
\begin{equation*}
\begin{aligned}
    \sup_{t>0}\mathcal E_{r,\rho,R}(t)
    &=
    \mathcal E_{r,\rho,R}(t_{r,\rho,R}^*)
    \\
    &=
    \mathcal K_{r,\rho,R}(t_{r,\rho,R}^*)
    \\
    &=
    \mathcal D_{r,\rho,R}(t_{r,\rho,R}^*).
\end{aligned}
\end{equation*}
Combining this identity with \eqref{eq:regularity-envelope} proves
\eqref{eq:uniform-regularity-envelope}.
\end{proof}

\paragraph{Proof of Corollary~\ref{cor:polynomial regularity envelope}:}
\begin{proof}
    Let
\begin{equation*}
    \bar t:=\frac{\log2}{2},
\end{equation*}
so that
\begin{equation*}
    \sigma_{\bar t}
    =
    e^{-\bar t}
    =
    \frac1{\sqrt2}.
\end{equation*}
At this reference time,
\begin{equation*}
    \mathcal K_{r,\rho,R}(\bar t)
    =
    2^{r+1}R^d
    \left(
        1+\sqrt{2\log(\rho^{-1})}
        +\sqrt2R+B_\rho
    \right)^{2(r+1)},
\end{equation*}
whereas
\begin{equation*}
    \mathcal M_{\bar t,\rho,R}
    =
    1+B_\rho+2\sqrt2R
    +\sqrt{2\log(\rho^{-1})}
    +2\sqrt{1+\kappa(r+1)}.
\end{equation*}
Since $\mathcal D_{r,\rho,R}$ is increasing and
$\mathcal K_{r,\rho,R}$ is decreasing,
\begin{equation*}
    \overline{\mathcal E}_{r,\rho,R}
    \leq
    \max\left\{
        \mathcal K_{r,\rho,R}(\bar t),
        \mathcal D_{r,\rho,R}(\bar t)
    \right\}.
\end{equation*}
For $R\geq1$, the kernel-induced term grows at most as
$R^{d+2(r+1)}$, whereas the data-induced term grows at most as
$R^{d+2\kappa(r+1)}$. Combining the two estimates gives
\begin{equation*}
    \overline{\mathcal E}_{r,\rho,R}
    \leq
    \mathcal C_{\mathrm{reg}}(H,\kappa;r,\rho)
    R^{d+2(r+1)(1\vee\kappa)}.
\end{equation*}
\end{proof}

\paragraph{Proof of Corollary~\ref{cor:uniform pointwise score bound}:}

\begin{proof}
Since $r>d/2$, the Sobolev embedding theorem on the unit ball
gives
\begin{equation*}
    \|g\|_{L^\infty(\Omega_1)}
    \leq
    C_{d,r}
    \|g\|_{H^r(\Omega_1)}
\end{equation*}
for every $g\in H^r(\Omega_1)$.

Fix $t>0$ and $i\in[d]$, and define
\begin{equation*}
    g_{t,i}(z)
    :=
    s_{t,i}(Rz),
    \qquad z\in\Omega_1.
\end{equation*}
For every multi-index $\alpha$ satisfying $|\alpha|\leq r$,
\begin{equation*}
    \partial^\alpha g_{t,i}(z)
    =
    R^{|\alpha|}
    (\partial^\alpha s_{t,i})(Rz).
\end{equation*}
Therefore, by the change of variables $x=Rz$,
\begin{align*}
    \|\partial^\alpha g_{t,i}\|_{L^2(\Omega_1)}^2
    &=
    R^{2|\alpha|}
    \int_{\Omega_1}
    |(\partial^\alpha s_{t,i})(Rz)|^2
    \,\mathrm dz
    \\
    &=
    R^{2|\alpha|-d}
    \|\partial^\alpha s_{t,i}\|_{L^2(\Omega_R)}^2.
\end{align*}
Since $R\geq1$ and $|\alpha|\leq r$, it follows that
\begin{align*}
    \|g_{t,i}\|_{H^r(\Omega_1)}^2
    &=
    \sum_{|\alpha|\leq r}
    R^{2|\alpha|-d}
    \|\partial^\alpha s_{t,i}\|_{L^2(\Omega_R)}^2
    \\
    &\leq
    R^{2r-d}
    \|s_{t,i}\|_{H^r(\Omega_R)}^2.
\end{align*}
Consequently,
\begin{align*}
    \|s_{t,i}\|_{L^\infty(\Omega_R)}
    &=
    \|g_{t,i}\|_{L^\infty(\Omega_1)}
    \\
    &\leq
    C_{d,r}
    \|g_{t,i}\|_{H^r(\Omega_1)}
    \\
    &\leq
    C_{d,r}
    R^{r-d/2}
    \|s_{t,i}\|_{H^r(\Omega_R)}.
\end{align*}
It follows that
\begin{align*}
    \sup_{x\in\Omega_R}\|s_t(x)\|_2
    &\leq
    \sqrt d\,
    \max_{1\leq i\leq d}
    \|s_{t,i}\|_{L^\infty(\Omega_R)}
    \\
    &\leq
    C_{d,r}
    R^{r-d/2}
    \max_{1\leq i\leq d}
    \|s_{t,i}\|_{H^r(\Omega_R)}.
\end{align*}
On the other hand, Corollary~\ref{cor:regularity envelope}
gives
\begin{equation*}
    \sup_{t>0}
    \max_{1\leq i\leq d}
    \|s_{t,i}\|_{H^r(\Omega_R)}^2
    \leq
    C_{d,r}
    \overline{\mathcal E}_{r,\rho,R}.
\end{equation*}
Combining the preceding estimates and taking the supremum over
$t>0$ yields
\begin{equation*}
    \sup_{t>0}
    \sup_{x\in\Omega_R}
    \|s_t(x)\|_2
    \leq
    C_{d,r}
    R^{r-d/2}
    \overline{\mathcal E}_{r,\rho,R}^{1/2}.
\end{equation*}
\end{proof}

\section{Proofs for Section~\ref{sec:shallow approx}}\label{appendix: shallow approx}

\paragraph{Proof of Lemma~\ref{lem:shallow approx}:}

Here we provide a detailed computation for calculating the explict depencies on $\Omega_R$ in Lemma~\ref{lem:shallow approx} by rescaling $\Omega_R$ to the unit ball.
\begin{proof}
    For a function $f:\RR^d\to \RR$, we introduce the scaling operator $T_R,$ defined by
    \begin{equation*}
        (T_Rf)(x):=f(Rx).
    \end{equation*}
    For any multi-index $\alpha$,
\begin{equation*}
    \partial^\alpha(T_Rf)(x)
    =
    R^{|\alpha|}
    (\partial^\alpha f)(Rx).
\end{equation*}
Consequently, by the change of variables $y=Rx$,
\begin{align*}
    \|\partial^\alpha(T_Rf)\|_{L^2(\Omega_1)}^2
    &=
    R^{2|\alpha|}
    \int_{\Omega_1}
    |(\partial^\alpha f)(Rx)|^2
    \,\mathrm dx
    \\
    &=
    R^{2|\alpha|-d}
    \|\partial^\alpha f\|_{L^2(\Omega_R)}^2.
\end{align*}
Therefore, for every $k\in\mathbb N_0$,
\begin{align*}
    \|T_Rf\|_{H^k(\Omega_1)}^2
    &=
    \sum_{|\alpha|\leq k}
    R^{2|\alpha|-d}
    \|\partial^\alpha f\|_{L^2(\Omega_R)}^2
    \\
    &\leq
    R^{-d}(1+R)^{2k}
    \|f\|_{H^k(\Omega_R)}^2.
\end{align*}
In particular, if $R\geq1$, then
\begin{equation*}
    \|T_Rf\|_{H^k(\Omega_1)}
    \lesssim_k
    R^{k-d/2}
    \|f\|_{H^k(\Omega_R)}.
\end{equation*}
By interpolation between consecutive integer-order Sobolev
spaces, the same scaling estimate holds for every real $s\geq0$:
\begin{equation*}
    \|T_Rf\|_{H^s(\Omega_1)}
    \lesssim_s
    R^{s-d/2}
    \|f\|_{H^s(\Omega_R)},
    \qquad R\geq1.
\end{equation*}

    We first state the approximation result on unit ball $\Omega_1$.

    From \cite{liu2025integral}, let $g=T_Rf\in H^{\frac{d+3}{2}}(\Omega_1)$, then we have 
        \begin{equation*}
        \inf_{g_m\in \Sigma_{m,M}^1}\|g-g_m\|_{L^2(\Omega_1)}\lesssim_d m^{-\frac{1}{2}-\frac{3}{2d}}\|g\|_{H^{\frac{d+3}{2}}(\Omega_1)},
    \end{equation*}
    with some $M\simeq_d \|g\|_{H^{\frac{d+3}{2}}(\Omega_1)}$.

    That is to say, there exist 
    \begin{equation*}
        g_m(x)=\sum_{j=1}^ma_j\sigma(w_j\cdot x+b_j)
    \end{equation*}
    with parameters satisfying
    \begin{equation*}
        \sum_{j=1}^m|a_j|\lesssim_d \|g\|_{H^{\frac{d+3}{2}}(\Omega_1)} , \qquad (w_j^\top,b_j)\in \mathbb S^{d}.
    \end{equation*}

    Define
    \begin{equation*}
        \lambda_j:= \left\|\left(\frac{w_j}{R},b_j\right)\right\|_2,\qquad \theta_j^\prime:=\lambda_j^{-1}\left(\frac{w_j}{R},b_j\right).
    \end{equation*}

    Then
    \begin{equation*}
        f_m(x):=g_m(x/R)=\sum_{j=1}^m a_j\sigma\left(\frac{w_j}{R}\cdot x+b_j\right)=\sum_{j=1}^m{a_j\lambda_j}\sigma\left(\theta_j^\prime\cdot\Tilde{x}\right).
    \end{equation*}
    Note that when $R\geq 1$,
    \begin{equation*}
        \sum_{j=1}^m|a_j\lambda_j|\leq \sum_{j=1}^m|a_j|\sqrt{\left(\frac{\|w_j\|^2}{R^2}+b_j^2\right)}\leq \sum_{j=1}^m|a_j|\lesssim_d\|g\|_{H^{\frac{d+3}{2}}(\Omega_1)}\lesssim_dR^{\frac{3}{2}}\|f\|_{H^{\frac{d+3}{2}}(\Omega_R)},
    \end{equation*}
    and
    \begin{equation*}
       \| \theta_j^\prime\|_2=1.
    \end{equation*}
    Thus $f_m\in \Sigma_{m,M}^1$ for some $M\simeq_d R^{\frac{3}{2}}\|f\|_{H^{\frac{d+3}{2}}(\Omega_R)}$.
    Moreover, 
    \begin{align*}
        \|f-f_m\|_{L^2(\Omega_R)}&=R^{\frac{d}{2}}\|g-g_m\|_{L^2(\Omega_1)}\\
        &\lesssim_d R^{\frac{d}{2}}m^{-\frac{1}{2}-\frac{3}{2d}}\|g\|_{H^{\frac{d+3}{2}}(\Omega_1)}\\
        &\lesssim_dR^{\frac{d+3}{2}}m^{-\frac{1}{2}-\frac{3}{2d}}\|f\|_{H^{\frac{d+3}{2}}(\Omega_R)}.
    \end{align*}
    
\end{proof}

\paragraph{Proof of Lemma~\ref{lem:density infty}}
\begin{proof}
    In fact, for O-U process, we have both
    \begin{equation*}
        \|p_t\|_{\infty}\leq \lp 2\pi\sigma_t^2\rp^{-d/2}
    \end{equation*}
    and 
    \begin{equation*}
         \|p_t\|_{\infty}\leq e^{dt}\|p_0\|_\infty. 
    \end{equation*}

    Set $A:=\|p_0\|_\infty^{2/d}$ and $b:=(2\pi)^{-1}$.
The two bounds above imply
\begin{equation*}
    \|p_t\|_\infty^{2/d}
    \leq
    \min\left\{
        Ae^{2t},\frac{b}{1-e^{-2t}}
    \right\}.
\end{equation*} 
Writing $u=e^{-2t}\in(0,1)$, we have
\begin{equation*}
    \min\left\{\frac{A}{u},\frac{b}{1-u}\right\}
    \leq A+b.
\end{equation*} 
Indeed, the first bound applies when $u\geq A/(A+b)$,
and the second applies otherwise.
Raising both sides to the power $d/2$ proves the claim.

\end{proof}

\paragraph{Proof of Theorem~\ref{thm:score approx}:}
\begin{proof}
    Since we assume $s_{t,i}\in H^{\frac{d+3}{2}}(\Omega_R)$, from Lemma~\ref{lem:shallow approx}, we have 
    \begin{equation*}
         \inf_{s_{m,i}\in \Sigma_{m,M_{t,i}}^1}\|s_{t,i}-s_{m,i}\|_{L^2(\Omega_R)}\lesssim_d R^{\frac{d+3}{2}}m^{-\frac{1}{2}-\frac{3}{2d}}\|s_{t,i}\|_{H^{\frac{d+3}{2}}(\Omega_R)},
    \end{equation*}
    with $M_{t,i}\simeq_{d}R^{\frac{3}{2}} \|s_{t,i}\|_{H^{\frac{d+3}{2}}(\Omega_R)}$. 

   Taking $M_t:=\max_{i\in[d]}M_{t,i}$, then 
   $$
   M_t\simeq_{d}R^{\frac{3}{2}}\max_{i\in[d]}\|s_{t,i}\|_{H^{\frac{d+3}{2}}(\Omega_R)}.
   $$
    In particular, $M_t\geq M_{t,i}$ for every $i\in[d]$. Further,
    $$
    \Sigma_{m,M_{t,i}}^1\subseteq\Sigma_{m,M_{t}}^1.
    $$ 
    Hence
    \begin{equation}\label{eq:shallow approx in thm proof}
         \inf_{s_{m,i}\in \Sigma_{m,M_{t}}^1}\|s_{t,i}-s_{m,i}\|_{L^2(\Omega_R)}\lesssim_d R^{\frac{d+3}{2}}m^{-\frac{1}{2}-\frac{3}{2d}}\|s_{t,i}\|_{H^{\frac{d+3}{2}}(\Omega_R)}.
    \end{equation}

    By Lemma~\ref{lem:density infty} and \eqref{eq:shallow approx in thm proof}, we have
    \begin{align*}
        &\inf_{s_m\in\mathcal F_{m,M_t}} 
    \mathbb E_{X_t\sim p_t}
    \left[
        \|s_m(X_t)-s_t(X_t)\|_2^2
        \mathbf 1_{\{X_t\in\Omega_R\}}
    \right]\\
    &=
     \sum_{i=1}^d \inf_{s_{m,i}\in\Sigma_{m,M_{t}}^1} \mathbb E_{X_t\sim p_t}
    \left[
        |s_{m,i}(X_t)-s_{t,i}(X_t)|^2
        \mathbf 1_{\{X_t\in\Omega_R\}}
    \right]\\
    &\leq P_\infty \sum_{i=1}^d\inf_{s_{m,i}\in\Sigma_{m,M_{t}}^1}\|s_{m,i}-s_{t,i}\|_{L^2(\Omega_R)}^2\\
    &\lesssim_d P_\infty R^{d+3}m^{-1-\frac{3}{d}}\max_{i\in[d]}\|s_{t,i}\|_{H^{\frac{d+3}{2}}(\Omega_R)}^2.
    \end{align*}
    Finally, taking square roots on both sides proves the desired result.
    
    If we further have $\sup_{t\in I}\|s_{t,i}\|_{H^{\frac{d+3}{2}}(\Omega_R)}<\infty$ for all $i\in[d]$, 
    we take $M:=\sup_{t\in I}M_t$, then $M\simeq_{d}R^{\frac{3}{2}}\sup_{t\in I}\max_{i\in[d]}\|s_{t,i}\|_{H^{\frac{d+3}{2}}(\Omega_R)}$, and $\Sigma_{m,M_t}^1\subseteq \Sigma_{m,M}^1$ for every $t\in I$. 
    Hence we can replace $M_t$ with $M$ in our theorem.
\end{proof}

\section{Proofs for Section~\ref{sec:stats}}

\subsection{Empirical Process Tools}

We first recall the empirical-process results used in the proof. Let $(\mathcal Z,P)$ be a probability space, let $\mathcal H$ be a class of measurable functions from $\mathcal Z$ to $\mathbb R$, and let $V_1,\ldots,V_n$ be i.i.d. observations with law $P$. For a measurable function $h$, write
\begin{equation*}
    Ph:=\mathbb E[h(V)],
    \qquad
    P_nh:=\frac1n\sum_{i=1}^n h(V_i).
\end{equation*}
Let $\xi_1,\ldots,\xi_n$ be i.i.d. Rademacher random variables, independent of the sample, and define
\begin{equation*}
    \mathfrak R_n h
    :=
    \frac1n\sum_{i=1}^n\xi_i h(V_i),
    \qquad
    \mathfrak R_n\mathcal H
    :=
    \sup_{h\in\mathcal H}\mathfrak R_nh.
\end{equation*}
The empirical Rademacher average is $\mathbb E_\xi\mathfrak R_n\mathcal H$, where the expectation is over the Rademacher variables. The population Rademacher average $\mathbb E\mathfrak R_n\mathcal H$ averages over both the sample and the Rademacher variables.

\paragraph{Sub-root functions.}
A function $\psi:[0,\infty)\to[0,\infty)$ is called sub-root if it is nonnegative and nondecreasing and if $r\mapsto\psi(r)/\sqrt r$ is nonincreasing on $(0,\infty)$.

\begin{lem}[\cite{bartlett2005local}, Lemma 3.2]
\label{lem:bartlett-fixed-point}
Let $\psi$ be a nontrivial sub-root function. Then $\psi$ is continuous on $[0,\infty)$ and has a unique positive fixed point $r^*$. Moreover, for every $r>0$,
\begin{equation*}
    r\geq\psi(r)
    \quad\Longleftrightarrow\quad
    r\geq r^*.
\end{equation*}
\end{lem}

\begin{lem}[\cite{bartlett2005local}, Lemma 3.4]
\label{lem:bartlett-subroot}
Let $\mathcal H$ be star-shaped around zero and let $T:\mathcal H\to[0,\infty)$ satisfy
\begin{equation*}
    T(\lambda h)\leq\lambda^2T(h)
    \qquad
    \text{for every }h\in\mathcal H
    \text{ and }\lambda\in[0,1].
\end{equation*}
Then
\begin{equation*}
    \psi(r)
    :=
    \mathbb E_\xi\mathfrak R_n
    \{h\in\mathcal H:T(h)\leq r\},
    \qquad r\geq0,
\end{equation*}
is sub-root. Its expectation with respect to the sample is also sub-root.
\end{lem}

\paragraph{Local Rademacher oracle inequality.}
The proof uses the following theorem of \cite{bartlett2005local}.

\begin{lem}[\cite{bartlett2005local}, Theorem 3.3]
\label{lem:bartlett local}
Let $\mathcal H$ be a class of functions taking values in $[a,b]$.
Suppose that a functional $T:\mathcal H\to[0,\infty)$ and a constant $B>0$ satisfy
\begin{equation*}
    \operatorname{Var}(h)
    \leq
    T(h)
    \leq
    BPh
    \qquad
    \text{for every }h\in\mathcal H.
\end{equation*}
Let $\psi$ be a sub-root function with fixed point $r^*$ and suppose that, for every $r\geq r^*$,
\begin{equation*}
    \psi(r)
    \geq
    B\mathbb E\mathfrak R_n
    \{h\in\mathcal H:T(h)\leq r\}.
\end{equation*}
Then, with $c_1=704$ and $c_2=26$, for every $\kappa>1$ and $u>0$, with probability at least $1-e^{-u}$,
\begin{equation*}
    Ph
    \leq
    \frac{\kappa}{\kappa-1}P_nh
    +
    \frac{c_1\kappa}{B}r^*
    +
    \frac{u\{11(b-a)+c_2B\kappa\}}{n}
\end{equation*}
holds simultaneously for all $h\in\mathcal H$.
\end{lem}

The following result compares population and empirical second-moment constraints.

\begin{lem}[\cite{bartlett2005local}, Corollary 2.2]
\label{lem:bartlett empirical localization}
Let $\mathcal H$ be a class of functions taking values in $[-b,b]$,
where $b>0$. If $u>0$ and $r>0$ satisfy
\begin{equation*}
    r
    \geq
    10b\mathbb E\mathfrak R_n
    \{h\in\mathcal H:Ph^2\leq r\}
    +
    \frac{11b^2u}{n},
\end{equation*}
then, with probability at least $1-e^{-u}$,
\begin{equation*}
    \{h\in\mathcal H:Ph^2\leq r\}
    \subseteq
    \{h\in\mathcal H:P_nh^2\leq2r\}.
\end{equation*}
\end{lem}

\paragraph{Entropy bound for empirical Rademacher averages.}
For a sample $S=(v_1,\ldots,v_n)$, let
$\mathcal N_2(\varepsilon,\mathcal H,S)$ be the external covering
number under the empirical metric
\begin{equation*}
    d_S(h,h')
    :=
    \left(
        \frac1n\sum_{i=1}^n|h(v_i)-h'(v_i)|^2
    \right)^{1/2}.
\end{equation*}

\begin{lem}[\cite{srebro2010optimistic}, Lemma A.1]
\label{lem:srebro dudley}
For every real-valued function class $\mathcal H$ and sample
$S=(v_1,\ldots,v_n)$,
then for any $\alpha\ge0$,
\begin{equation*}
    \mathbb E_\xi\mathfrak R_n\mathcal H
    \leq
        4\alpha
        +10\int_\alpha^{\sup_{h\in\mathcal H}\sqrt{P_nh^2}}
        \sqrt{
            \frac{
                \log\mathcal N_2(\varepsilon,\mathcal H,S)
            }{n}
        }
        \,\mathrm d\varepsilon.
\end{equation*}
In particular, taking infimum over $\alpha\ge0$ yields
\begin{equation*}
           \mathbb E_\xi\mathfrak R_n\mathcal H
    \leq \inf_{\alpha\ge0}
    \left\{
        4\alpha
        +10\int_\alpha^{\sup_{h\in\mathcal H}\sqrt{P_nh^2}}
        \sqrt{
            \frac{
                \log\mathcal N_2(\varepsilon,\mathcal H,S)
            }{n}
        }
        \,\mathrm d\varepsilon
    \right\}.
\end{equation*}
\end{lem}

\subsection{Proof of Theorem~\ref{thm:local oracle}}

\begin{proof}
For $f\in\mathcal F$, define the shifted, truncated excess loss
\begin{equation*}
    \bar h_{f,R}(x,y)
    :=
    h_f(x,y)\mathbf 1_{\mathcal A_R}(x,y)+\Delta_R
\end{equation*}
and the corresponding class
\begin{equation*}
    \mathcal H_{\mathrm{loss},R}
    :=
    \{\bar h_{f,R}:f\in\mathcal F\}.
\end{equation*}
The excess loss admits the decomposition
\begin{equation*}
    h_f(x,y)
    =
    \|f(x)-s_t(x)\|_2^2
    +
    2\langle f(x)-s_t(x),s_t(x)-y\rangle.
\end{equation*}
By Tweedie's formula, $s_t(x)=\mathbb E[Y\mid X=x]$, and therefore
\begin{equation*}
    Ph_f
    =
    \mathbb E_{X\sim P_X}\|f(X)-s_t(X)\|_2^2.
\end{equation*}
Moreover,
\begin{align*}
    P\bar h_{f,R}-Ph_f
    &=
    \Delta_R
    -
    \mathbb E\left[
        h_f(X,Y)\mathbf 1_{\mathcal A_R^c}(X,Y)
    \right],
\end{align*}
so the definition of $\Delta_R$ yields
\begin{equation*}
    Ph_f
    \leq
    P\bar h_{f,R}
    \leq
    Ph_f+2\Delta_R.
\end{equation*}
In particular,
\begin{equation*}
    \mathcal R_t(\widehat f)
    =
    Ph_{\widehat f}
    \leq
    P\bar h_{\widehat f,R}.
\end{equation*}

\paragraph{Step I: Boundedness and the variance condition.}
For $(x,y)\in\mathcal A_R$,
\begin{equation*}
    \|f(x)\|_2\leq M_R,
    \qquad
    \|s_t(x)\|_2\leq M_R,
    \qquad
    \|s_t(x)-y\|_2\leq N_R.
\end{equation*}
Consequently,
\begin{align*}
    h_f(x,y)^2
    &\leq
    2\|f(x)-s_t(x)\|_2^4
    +
    8|\langle f(x)-s_t(x),s_t(x)-y\rangle|^2
    \\
    &\leq
    (8M_R^2+8N_R^2)
    \|f(x)-s_t(x)\|_2^2
    \\
    &=
    B_R\|f(x)-s_t(x)\|_2^2.
\end{align*}
Since a constant shift does not change variance,
\begin{align*}
    \operatorname{Var}(\bar h_{f,R})
    &=
    \operatorname{Var}\bigl(
        h_f\mathbf 1_{\mathcal A_R}
    \bigr)
    \\
    &\leq
    \mathbb E\left[
        h_f(X,Y)^2\mathbf 1_{\mathcal A_R}(X,Y)
    \right]
    \\
    &\leq
    B_R\mathbb E\left[
        \|f(X)-s_t(X)\|_2^2
        \mathbf 1_{\mathcal A_R}(X,Y)
    \right]
    \\
    &\leq
    B_RPh_f
    \leq
    B_RP\bar h_{f,R}.
\end{align*}
This verifies the variance condition for the class $\mathcal H_{\mathrm{loss},R}$ in Lemma~\ref{lem:bartlett local}  with the functional
\begin{equation*}
    T(\bar h_{f,R})
    :=
    B_RP\bar h_{f,R}.
\end{equation*}
Also, for $(x,y)\in\mathcal A_R$,
\begin{equation*}
    |h_f(x,y)|
    \leq
    4M_R^2+4M_RN_R
    =
    U_R.
\end{equation*}
Because $h_f\mathbf 1_{\mathcal A_R}=0$ outside
$\mathcal A_R$, it follows that
\begin{equation*}
    \bar h_{f,R}
    \in
    [\Delta_R-U_R,\Delta_R+U_R]
\end{equation*}
pointwise. The range length is therefore $2U_R$.

\paragraph{Step II: Upper bounds for the local Rademacher complexity.}
Fix $r>0$. We begin with
\begin{equation*}
    \mathbb E\mathfrak R_n
    \left\{
        \bar h_{f,R}:
        P\bar h_{f,R}\leq\frac r{B_R}
    \right\}.
\end{equation*}
 The constant shift $\Delta_R$ does not contribute to the Rademacher complexity after taking expectation over the Rademacher variables. Hence
\begin{align*}
    &\mathbb E\mathfrak R_n
    \left\{
        \bar h_{f,R}:
        P\bar h_{f,R}\leq\frac r{B_R}
    \right\}
    =
    \mathbb E\mathfrak R_n
    \left\{
        h_f\mathbf 1_{\mathcal A_R}:
        P\bar h_{f,R}\leq\frac r{B_R}
    \right\}.
\end{align*}
Furthermore,
\begin{equation*}
    P\bar h_{f,R}
    \geq
    Ph_f
    =
    \mathbb E_{X\sim P_X}\|f(X)-s_t(X)\|_2^2
    \geq
    \mathbb E\left[
        \|f(X)-s_t(X)\|_2^2
        \mathbf 1_{\mathcal A_R}(X,Y)
    \right].
\end{equation*}
Define
\begin{equation*}
    \mathcal F_R(r)
    :=
    \left\{
        f\in\mathcal F:
        \mathbb E\left[
        \|f(X)-s_t(X)\|_2^2
        \mathbf 1_{\mathcal A_R}(X,Y)
    \right]\leq\frac{r}{B_R}
    \right\}.
\end{equation*}

Using the decomposition of
\begin{equation*}
    h_f(x,y)=\|f(x)-s_t(x)\|_2^2+2\langle f(x)-s_t(x),s_t(x)-y\rangle,
\end{equation*}
we thus obtain
\begin{equation*}
    \mathbb E\mathfrak R_n
    \left\{
        \bar h_{f,R}:
        P\bar h_{f,R}\leq\frac r{B_R}
    \right\}
    \leq
    \mathcal I_1(r)+\mathcal I_2(r),
\end{equation*}
where
\begin{align*}
    \mathcal I_1(r)
    &:=
    \mathbb E\mathfrak R_n
    \left\{
        (x,y)\mapsto
        \|f(x)-s_t(x)\|_2^2
        \mathbf 1_{\mathcal A_R}(x,y):
        f\in\mathcal F_R(r)
    \right\},
    \\
    \mathcal I_2(r)
    &:=
    2\mathbb E\mathfrak R_n
    \left\{
        (x,y)\mapsto
        \langle
            f(x)-s_t(x),
            s_t(x)-y
        \rangle
        \mathbf 1_{\mathcal A_R}(x,y):
        f\in\mathcal F_R(r)
    \right\}.
\end{align*}

For each coordinate $j$, recall
\begin{equation*}
    \mathcal G_{j,R}
    =
    \left\{
        (x,y)\mapsto
        (f_j(x)-s_{t,j}(x))
        \mathbf 1_{\mathcal A_R}(x,y):
        f\in\mathcal F
    \right\}
\end{equation*}
and $\mathcal G_{j,R}'=\operatorname{star}(\mathcal G_{j,R})$.
Every $g\in\mathcal G_{j,R}'$ satisfies $|g|\leq2M_R$, and
$0\in\mathcal G_{j,R}'$. Dropping the joint constraint across coordinates and allowing each coordinate function to be selected independently enlarges the supremum. The contraction principle, applied to $v\mapsto v^2$, which is $4M_R$-Lipschitz on $[-2M_R,2M_R]$, therefore gives
\begin{align*}
    \mathcal I_1(r)
    &\leq
    \sum_{j=1}^d
    \mathbb E\mathfrak R_n
    \left\{
        g^2:
        g\in\mathcal G_{j,R}',
        \ Pg^2\leq\frac r{B_R}
    \right\}
    \\
    &\leq
    4M_R\sum_{j=1}^d
    \mathbb E\mathfrak R_n
    \left\{
        g:
        g\in\mathcal G_{j,R}',
        \ Pg^2\leq\frac r{B_R}
    \right\}.
\end{align*}
For the multiplier term, define
\begin{equation*}
    m_{j,R}(x,y)
    :=
    (s_{t,j}(x)-y_j)\mathbf 1_{\mathcal A_R}(x,y).
\end{equation*}
Then $|m_{j,R}|\leq N_R$. Conditional on the sample, the maps $v\mapsto m_{j,R}(X_i,Y_i)v$ are $N_R$-Lipschitz and vanish at zero. Another application of the contraction principle yields
\begin{equation*}
    \mathcal I_2(r)
    \leq
    2N_R\sum_{j=1}^d
    \mathbb E\mathfrak R_n
    \left\{
        g:
        g\in\mathcal G_{j,R}',
        \ Pg^2\leq\frac r{B_R}
    \right\}.
\end{equation*}
Consequently,
\begin{align*}
    &\mathbb E\mathfrak R_n
    \left\{
        \bar h_{f,R}:
        P\bar h_{f,R}\leq\frac r{B_R}
    \right\}
    \leq
    (4M_R+2N_R)
    \sum_{j=1}^d\Psi_{j,R}(r),
\end{align*}
where
\begin{equation*}
    \Psi_{j,R}(r)
    :=
    \mathbb E\mathfrak R_n
    \left\{
        g:
        g\in\mathcal G_{j,R}',
        \ Pg^2\leq\frac r{B_R}
    \right\}.
\end{equation*}
Apply Lemma~\ref{lem:bartlett-subroot} with
$T_j(g):=B_RPg^2$. It follows that each $\Psi_{j,R}$ is sub-root.

Define
\begin{equation*}
    \Psi_R(r)
    :=
    B_RQ_R\sum_{j=1}^d\Psi_{j,R}(r)
    +
    \frac{44B_RM_R^2\log(dn)}{n}.
\end{equation*}
Positive linear combinations of sub-root functions remain sub-root, and adding a nonnegative constant preserves the sub-root property.
Thus, $\Psi_R$ is sub-root. Since $n\geq2$ and $M_R>0$, its constant term is strictly positive.
Let $r_{0,R}$ denote its unique positive fixed point.

For $r\geq r_{0,R}$, the fixed-point property gives $r\geq\Psi_R(r)$. Since $Q_R\geq20M_R$, for every coordinate $j$,
\begin{equation*}
    \frac r{B_R}
    \geq
    20M_R\Psi_{j,R}(r)
    +
    \frac{44M_R^2\log(dn)}{n}.
\end{equation*}
Fix such an $r$ and a coordinate $j$. Because $n\geq2$, we have $\log(dn)>0$. Lemma~\ref{lem:bartlett empirical localization}, with $b=2M_R$ and $u=\log(dn)$, shows that the event
\begin{align*}
    \mathcal E_{j,R}(r)
    :=
    \Biggl\{
        &\left\{
            g\in\mathcal G_{j,R}':
            Pg^2\leq\frac r{B_R}
        \right\}
        \subseteq
        \left\{
            g\in\mathcal G_{j,R}':
            P_ng^2\leq\frac{2r}{B_R}
        \right\}
    \Biggr\}
\end{align*}
has probability at least $1-(dn)^{-1}$. No simultaneity over $r$ is needed here; $r$ is fixed in this application.

On $\mathcal E_{j,R}(r)$, every function satisfying the population second-moment bound also satisfies the corresponding empirical bound. On the complement, the conditional Rademacher average is at most $2M_R$. Splitting the expectation over these two events gives
  \begin{align*}
        \Psi_{j,R}(r)
        &=
        \mathbb E_S
        \mathbb E_\xi \mathfrak R_n
        \left\{
        g\in\mathcal G_{j,R}':
        Pg^2\le \frac r{B_R}
        \right\} \\
        &=
        \mathbb E_S
        \left[
        \mathbb E_\xi \mathfrak R_n
        \left\{
        g\in\mathcal G_{j,R}':
        Pg^2\le \frac r{B_R}
        \right\}
        \mathbf 1_{\mathcal E_{j,R}(r)}
        \right] \\
        &\quad+
        \mathbb E_S
        \left[
        \mathbb E_\xi \mathfrak R_n
        \left\{
        g\in\mathcal G_{j,R}':
        Pg^2\le \frac r{B_R}
        \right\}
        \mathbf 1_{\mathcal E_{j,R}(r)^c}
        \right] \\
        &\le
        \mathbb E_S
        \left[
        \mathbb E_\xi \mathfrak R_n
        \left\{
        g\in\mathcal G_{j,R}':
        P_ng^2\le \frac{2r}{B_R}
        \right\}
        \right]
        +
        2M_R\mathbb P(\mathcal E_{j,R}(r)^c)\\
        &\le
        \mathbb E_S
        \mathbb E_\xi \mathfrak R_n
        \left\{
        g\in\mathcal G_{j,R}':
        P_ng^2\le \frac{2r}{B_R}
        \right\}
        +
        \frac{2M_R}{dn}.
    \end{align*}

Applying Lemma~\ref{lem:srebro dudley} with $\alpha=0$,
and using the monotonicity of external covering numbers
under set inclusion, yields
\begin{align*}
    &\mathbb E_\xi \mathfrak R_n
        \left\{
        g\in\mathcal G_{j,R}':
        P_ng^2\le \frac{2r}{B_R}
        \right\}
    \leq
       10\int_0^{\sqrt{2r/B_R}}
        \sqrt{
            \frac{
                \log\mathcal N_2(
                    \varepsilon,\mathcal G_{j,R}',S
                )
            }{n}
        }
        \,\mathrm d\varepsilon.
\end{align*}

Taking expectation with respect to $S$ gives
\begin{equation*}
    \Psi_{j,R}(r)
    \leq
    \mathfrak D_{j,R}(r)
    +
    \frac{2M_R}{dn}.
\end{equation*}
Combining this estimate with the coordinate reduction above yields, for every $r\geq r_{0,R}$,
\begin{align*}
    B_R\mathbb E\mathfrak R_n
    \left\{
        \bar h_{f,R}:
        P\bar h_{f,R}\leq\frac r{B_R}
    \right\}
    &\leq
    B_RQ_R\sum_{j=1}^d\mathfrak D_{j,R}(r)
    +
    \frac{2M_RB_RQ_R}{n}
    \\
    &\leq
    \Phi_R(r)\\
    &\leq
    \psi_R(r).
\end{align*}
The preceding bound on each $\Psi_{j,R}$ also shows that
\begin{equation*}
    r_{0,R}
    =
    \Psi_R(r_{0,R})
    \leq
    \Phi_R(r_{0,R})
    \leq
    \psi_R(r_{0,R}),
\end{equation*}
Lemma~\ref{lem:bartlett-fixed-point} therefore implies $r_{0,R}\leq r_R^*$. Consequently, for every $r\geq r_R^*$,
\begin{equation*}
    \psi_R(r)
    \geq
    B_R\mathbb E\mathfrak R_n
    \left\{
        \bar h_{f,R}:
        P\bar h_{f,R}\leq\frac r{B_R}
    \right\}.
\end{equation*}
This is precisely the  condition in Lemma~\ref{lem:bartlett local} for $\mathcal H_{\mathrm{loss},R}$ and $T(\bar h_{f,R})=B_RP\bar h_{f,R}$.

\paragraph{Step III: The oracle inequality.}
Lemma~\ref{lem:bartlett local} shows that, for every $\kappa>1$ and $u>0$, with probability at least $1-e^{-u}$,
\begin{equation*}
    P\bar h_{\widehat f,R}
    \leq
    \frac{\kappa}{\kappa-1}P_n\bar h_{\widehat f,R}
    +
    \frac{c_1\kappa}{B_R}r_R^*
    +
    \frac{u(22U_R+c_2B_R\kappa)}{n}.
\end{equation*}

Fix $\zeta>0$ and choose $f_\zeta\in\mathcal F$ such that
\begin{equation*}
    Ph_{f_\zeta}
    \leq
    \varepsilon_{\mathrm{app}}+\zeta.
\end{equation*}
Bernstein's inequality and the bounds from Step I imply that, for every $\delta\in(0,1)$ and $\lambda>0$, with probability at least $1-\delta$,
\begin{align*}
    P_n\bar h_{f_\zeta,R}
     &\leq
    P\bar h_{f_\zeta,R}
    +
    \sqrt{
        \frac{
            2\operatorname{Var}(\bar h_{f_\zeta,R})\log(1/\delta)
        }{n}
    }
    +
    \frac{2U_R\log(1/\delta)}{3n}
    \\
    &\leq
    P\bar h_{f_\zeta,R}
    +
    \sqrt{
        \frac{
            2B_RP\bar h_{f_\zeta,R}\log(1/\delta)
        }{n}
    }
    +
    \frac{2U_R\log(1/\delta)}{3n}
    \\
    &\leq
    (1+\lambda)P\bar h_{f_\zeta,R}
    +
    \left(
        \frac{B_R}{2\lambda}
        +\frac{2U_R}{3}
    \right)
    \frac{\log(1/\delta)}{n}
    \\
    &\leq
    (1+\lambda)
    (\varepsilon_{\mathrm{app}}+\zeta+2\Delta_R)
    +
    \left(
        \frac{B_R}{2\lambda}
        +\frac{2U_R}{3}
    \right)
    \frac{\log(1/\delta)}{n}.
\end{align*}

Let $\mathcal E_R$ be the event that all training observations lie in $\mathcal A_R$, formally,
\begin{equation*}
    \mathcal E_R
    :=
    \bigcap_{i=1}^n\{(X_i,Y_i)\in\mathcal A_R\}.
\end{equation*}
Since $\mathbb P((X,Y)\notin\mathcal A_R)=\eta_R$, the union bound gives
\begin{equation*}
    \mathbb P(\mathcal E_R)
    \geq
    1-n\eta_R.
\end{equation*}
On $\mathcal E_R$,
\begin{equation*}
    P_n\bar h_{f,R}
    =
    P_nh_f+\Delta_R
    \qquad
    \text{for every }f\in\mathcal F.
\end{equation*}
Because $\widehat f$ minimizes the empirical DSM loss, $P_nh_{\widehat f}\leq P_nh_{f_\zeta}$. Hence, on $\mathcal E_R$,
\begin{equation*}
    P_n\bar h_{\widehat f,R}
    \leq
    P_n\bar h_{f_\zeta,R}.
\end{equation*}

Combining the two concentration events with $\mathcal E_R$, we obtain, with probability at least
$1-e^{-u}-\delta-n\eta_R$,
\begin{align*}
    \mathcal R_t(\widehat f)
    \leq\;&
    \frac{\kappa}{\kappa-1}
    \Biggl[
        (1+\lambda)
        (\varepsilon_{\mathrm{app}}+\zeta+2\Delta_R)
        +
        \left(
            \frac{B_R}{2\lambda}
            +\frac{2U_R}{3}
        \right)
        \frac{\log(1/\delta)}{n}
    \Biggr]
    \\
    &+
    \frac{c_1\kappa}{B_R}r_R^*
    +
    \frac{u(22U_R+c_2B_R\kappa)}{n}.
\end{align*}
Finally, take $\kappa=2$, $\lambda=1$, and $\delta=e^{-u}$, and let
$\zeta\downarrow0$. Since $c_1=704$ and $c_2=26$, this gives
\begin{align*}
    \mathbb E_{X\sim P_X}
    \|\widehat f(X)-s_t(X)\|_2^2
    \leq\;&
    4\varepsilon_{\mathrm{app}}
    +8\Delta_R
    +\frac{1408}{B_R}r_R^*
    +
    \left(
        53B_R+\frac{70}{3}U_R
    \right)
    \frac{u}{n}
\end{align*}
with probability at least $1-2e^{-u}-n\eta_R$. This proves the
claim.
\end{proof}

\section{Proofs in Section~\ref{sec:stats shallow}}\label{app:stats shallow}
\paragraph{Proof of Lemma~\ref{lem:localized shallow bound}}
\begin{proof}
For each component $f_j$, using the $1$-Lipschitzness and positive homogeneity of the ReLU activation, we have
\begin{align*}
    |f_j(x)|
    &\le
    \sum_{\ell=1}^m
    |a_{\ell,j}|
    |\sigma(\theta_{\ell,j}\cdot\tilde x)| \\
    &\le
    \sum_{\ell=1}^m
    |a_{\ell,j}|
    \|\theta_{\ell,j}\|_2
    \|\tilde x\|_2 \\
    &\le
    M\sqrt{1+\|x\|_2^2}.
\end{align*}
Taking the Euclidean norm over $d$ coordinates gives
\begin{equation*}
    \|f(x)\|_2
    \le
    \sqrt d\,M\sqrt{1+\|x\|_2^2}.
\end{equation*}
The bound on the ball follows by restricting to $\|x\|_2\le R_X$.
\end{proof}

\paragraph{Proof of Proposition~\ref{prop:shallow fixed point}:}
First, we state an auxiliary lemma.
\begin{lem}[\cite{mendelson2002improving}, Lemma 4.5]\label{lem: star covering}
Let $(E,\|\cdot\|)$ be a normed space, and let $A$ be a totally bounded subset of the unit ball of $E$. Then, for every $0<\varepsilon\le 1$,
\begin{equation*}
    \log \mathcal N(2\varepsilon,\operatorname{star}(A),\|\cdot\|)
    \le
    \log\frac{2}{\varepsilon}
    +
    \log\mathcal N(\varepsilon,A,\|\cdot\|).
\end{equation*}
\end{lem}
With this result, we are able to give an estimate for the fixed point.
\begin{proof}
Set
\begin{equation*}
    W:=m(d+2),
    \qquad
    R_\Theta:=\sqrt{M^2+m},
    \qquad
    R_X':=\sqrt{1+R_X^2},
    \qquad
    L_R:=R_X'\sqrt{m+M^2}.
\end{equation*}
We further introduce the auxiliary quantities
\begin{equation*}
    A_{m,R}
    :=
    \max\left\{
        8M_R,\,
        2M_R+4L_RR_\Theta
    \right\}
    =
    \max\left\{
        8M_R,\,
        2M_R+4R_X'(m+M^2)
    \right\}
\end{equation*}
and
\begin{equation*}
    D_R
    :=
    2M_RB_RQ_R
    +
    44B_RM_R^2\log(dn).
\end{equation*}

We first estimate the empirical metric entropy of $\mathcal G_{j,R}'$.  Write each $f_\Theta\in\Sigma_{m,M}^1$ as
\begin{equation*}
    f_\Theta(x)
    =
    \sum_{\ell=1}^m
    a_\ell
    \sigma(\theta_\ell^\top\widetilde x),
    \qquad
    \widetilde x=(x^\top,1)^\top.
\end{equation*}
Let $\mathcal P_{m,M}$ denote the parameter set
\begin{equation*}
    \mathcal P_{m,M}
    :=
    \left\{
        \Theta
        =
        (a_1,\ldots,a_m,
        \theta_1^\top,\ldots,\theta_m^\top)^\top:
        \sum_{\ell=1}^m|a_\ell|\leq M,\ 
        \|\theta_\ell\|_2=1
    \right\}.
\end{equation*}
For every $\Theta\in\mathcal P_{m,M}$,
\begin{equation*}
    \|\Theta\|_2^2
    =
    \sum_{\ell=1}^m|a_\ell|^2
    +
    \sum_{\ell=1}^m\|\theta_\ell\|_2^2
    \leq
    M^2+m
    =
    R_\Theta^2.
\end{equation*}
Thus,
\begin{equation*}
    \mathcal P_{m,M}
    \subseteq
    B_2(0,R_\Theta)\subseteq \RR^W.
\end{equation*}

For $\|x\|_2\leq R_X$, we have $\|\widetilde x\|_2\leq R_X'$. Hence, for any $\Theta,\Theta'\in\mathcal P_{m,M}$, the $1$-Lipschitz property of the ReLU activation gives
\begin{equation*}
\begin{aligned}
    |f_\Theta(x)-f_{\Theta'}(x)|
    \leq&
    \sum_{\ell=1}^m
    |a_\ell-a_\ell'|
    |\sigma(\theta_\ell^\top\widetilde x)|
    +
    \sum_{\ell=1}^m
    |a_\ell'|
    \left|
        \sigma(\theta_\ell^\top\widetilde x)
        -
        \sigma((\theta_\ell')^\top\widetilde x)
    \right|
    \\
    \le&
    R_X'
    \sum_{\ell=1}^m
    |a_\ell-a_\ell'|
    +
    R_X'
    \sum_{\ell=1}^m
    |a_\ell'|
    \|\theta_\ell-\theta_\ell'\|_2 \\
    \leq&
    R_X'\sqrt m
    \left(
        \sum_{\ell=1}^m|a_\ell-a_\ell'|^2
    \right)^{1/2}+R_X'M\left(    \sum_{\ell=1}^m\|\theta_\ell-\theta_\ell'\|_2^2
    \right)^{1/2}
    \\
    \leq\;&
    R_X'\sqrt{m+M^2}
    \|\Theta-\Theta'\|_2
    \\
    =\;&
    L_R\|\Theta-\Theta'\|_2.
\end{aligned}
\end{equation*}

For the sample $S=((X_1,Y_1),\ldots,(X_n,Y_n))$, define
\begin{equation*}
    g_{\Theta,j}(x,y)
    :=
    \bigl(f_\Theta(x)-s_{t,j}(x)\bigr)
    \mathbf 1_{\mathcal A_R}(x,y)\in\mathcal G_{j,R}.
\end{equation*}
Then
\begin{equation*}
\begin{aligned}
    d_S(g_{\Theta,j},g_{\Theta',j})^2
    &=
    \frac1n\sum_{i=1}^n
    |f_\Theta(X_i)-f_{\Theta'}(X_i)|^2
    \mathbf 1_{\mathcal A_R}(X_i,Y_i)
    \\
    &\leq
    L_R^2\|\Theta-\Theta'\|_2^2.
\end{aligned}
\end{equation*}
Since
$\mathcal P_{m,M}\subseteq B_2(0,R_\Theta)$, the standard volumetric bound gives
\begin{equation*}
    \mathcal N\left(
        \frac{\varepsilon}{L_R},
        \mathcal P_{m,M},
        \|\cdot\|_2
    \right)
    \leq
    \left(
        1+\frac{2L_RR_\Theta}{\varepsilon}
    \right)^W.
\end{equation*}
By the preceding Lipschitz estimate, this parameter-space cover induces an empirical $L_2$ $\varepsilon$-cover of $\mathcal G_{j,R}$. Hence
\begin{equation*}
    \mathcal N_2(
        \varepsilon,\mathcal G_{j,R},S
    )
    \leq
    \left(
        1+\frac{2L_RR_\Theta}{\varepsilon}
    \right)^W.
\end{equation*}

On $\mathcal A_R$, every $g\in\mathcal G_{j,R}$ satisfies $|g|\leq2M_R$. Define
\begin{equation*}
    \widetilde{\mathcal G}_{j,R}
    :=
    \frac{1}{2M_R}\mathcal G_{j,R}.
\end{equation*}
Then $\widetilde{\mathcal G}_{j,R}$ is contained in the unit ball of the empirical $L_2$ space, and
\begin{equation*}
    \mathcal G_{j,R}'
    =
    2M_R
    \operatorname{star}
    \bigl(\widetilde{\mathcal G}_{j,R}\bigr).
\end{equation*}
Applying Lemma~\ref{lem: star covering} with $\delta=\varepsilon/(4M_R)$, for $0<\varepsilon\leq2M_R$ we obtain
\begin{align*}
    \log\mathcal N_2(\varepsilon,\mathcal G_{j,R}',S)
    &=
    \log\mathcal N_2
    \left(
    \frac{\varepsilon}{2M_R},
    \operatorname{star}(\widetilde{\mathcal G}_{j,R}),
    S
    \right) \\
    &\le
    \log\left(\frac{8M_R}{\varepsilon}\right)
    +
    \log\mathcal N_2
    \left(
    \frac{\varepsilon}{2},
    \mathcal G_{j,R},
    S
    \right) \\
    &\le
    \log\left(\frac{8M_R}{\varepsilon}\right)
    +
    W\log\left(
    1+\frac{4L_RR_\Theta}{\varepsilon}
    \right).
\end{align*}

For $r\geq0$, define
\begin{equation*}
    K_r
    :=
    \min\left\{
        \sqrt{\frac{2r}{B_R}},
        2M_R
    \right\}.
\end{equation*}
Since every element of $\mathcal G_{j,R}'$ is bounded by $2M_R$, the zero function is a one-element empirical cover whenever $\varepsilon\geq2M_R$. Therefore,
\begin{equation*}
    \log\mathcal N_2(
        \varepsilon,\mathcal G_{j,R}',S
    )
    =
    0,
    \qquad
    \varepsilon\geq2M_R.
\end{equation*}
Moreover, for $0<\varepsilon\leq K_r$, the definition of $A_{m,R}$ gives
\begin{equation*}
\begin{aligned}
    \log\left(\frac{8M_R}{\varepsilon}\right)
    &\leq
    \log\left(\frac{A_{m,R}}{\varepsilon}\right) \quad\text{and}\quad
    \log\left(
        1+\frac{4L_RR_\Theta}{\varepsilon}
    \right)
    &\leq
    \log\left(\frac{A_{m,R}}{\varepsilon}\right).
\end{aligned}
\end{equation*}
Consequently,
\begin{equation*}
    \log\mathcal N_2(
        \varepsilon,\mathcal G_{j,R}',S
    )
    \leq
    (W+1)
    \log\left(\frac{A_{m,R}}{\varepsilon}\right),
    \qquad
    0<\varepsilon\leq K_r.
\end{equation*}

Combining \eqref{eq:D_j,R} with the preceding entropy bound and truncating the integral at $K_r$, we obtain
\begin{equation*}
\begin{aligned}
    \mathfrak D_{j,R}(r)
    &=
    10\mathbb E_S
    \int_0^{\sqrt{2r/B_R}}
    \sqrt{
        \frac{
            \log\mathcal N_2(
                \varepsilon,\mathcal G_{j,R}',S
            )
        }{n}
    }
    \,\mathrm d\varepsilon
    \\
    &\leq
    10\sqrt{\frac{W+1}{n}}
    \int_0^{K_r}
    \sqrt{
        \log\left(\frac{A_{m,R}}{\varepsilon}\right)
    }
    \,\mathrm d\varepsilon.
\end{aligned}
\end{equation*}
By the Cauchy--Schwarz inequality,
\begin{equation*}
\begin{aligned}
    \int_0^{K_r}
    \sqrt{
        \log\left(\frac{A_{m,R}}{\varepsilon}\right)
    }
    \,\mathrm d\varepsilon
    &\leq
    K_r^{1/2}
    \left(
        \int_0^{K_r}
        \log\left(\frac{A_{m,R}}{\varepsilon}\right)
        \,\mathrm d\varepsilon
    \right)^{1/2}
    \\
    &=
    K_r
    \sqrt{
        \log\left(\frac{A_{m,R}}{K_r}\right)+1
    }.
\end{aligned}
\end{equation*}
Thus,
\begin{equation*}
    \mathfrak D_{j,R}(r)
    \leq
    10
    \sqrt{\frac{W+1}{n}}
    K_r
    \sqrt{
        \log\left(\frac{A_{m,R}}{K_r}\right)+1
    }.
\end{equation*}

Define
\begin{equation*}
    q_{m,R}(u):=
    \begin{cases}
        0, & u=0,\\[1ex]
        u\sqrt{
            1+\log\left(\dfrac{A_{m,R}}{u}\right)
        }, & 0<u\leq2M_R.
    \end{cases}
\end{equation*}
and set
\begin{equation*}
    \psi_{m,M,R}(r)
    :=
    10B_RQ_Rd
    \sqrt{\frac{W+1}{n}}
    q_{m,R}(K_r)
    +
    \frac{D_R}{n}.
\end{equation*}
The definition of $\Phi_R$ and the coordinatewise entropy estimate imply
\begin{equation*}
\begin{aligned}
    \Phi_R(r)
    &=
    B_RQ_R
    \sum_{j=1}^d
    \mathfrak D_{j,R}(r)
    +
    \frac{D_R}{n}
    \\
    &\leq
    10B_RQ_Rd
    \sqrt{\frac{W+1}{n}}
    q_{m,R}(K_r)
    +
    \frac{D_R}{n}
    \\
    &=
    \psi_{m,M,R}(r).
\end{aligned}
\end{equation*}

We next verify that $\psi_{m,M,R}$ is a sub-root function. For $0<u\leq2M_R$, since $A_{m,R}\geq8M_R$, we have $A_{m,R}/u\geq4$, and hence
\begin{equation*}
\begin{aligned}
    q_{m,R}'(u)
    &=
    \sqrt{
        \log\left(\frac{A_{m,R}}{u}\right)+1
    }
    -
    \frac{
        1
    }{
        2\sqrt{
            \log\left(\frac{A_{m,R}}{u}\right)+1
        }
    }>0.
\end{aligned}
\end{equation*}
Thus $q_{m,R}(K_r)$ is nondecreasing in $r$. Furthermore,
\begin{equation*}
    \frac{q_{m,R}(K_r)}{\sqrt r}
    =
    \begin{cases}
        \displaystyle
        \sqrt{\frac{2}{B_R}}
        \sqrt{
            1+
            \log\left(
                A_{m,R}\sqrt{\frac{B_R}{2r}}
            \right)
        },
        &0<r\leq2B_RM_R^2,
        \\[2ex]
        \displaystyle
        \frac{q_{m,R}(2M_R)}{\sqrt r},
        &r\geq2B_RM_R^2.
    \end{cases}
\end{equation*}
The right-hand side is nonincreasing in $r$. Since $D_R/(n\sqrt r)$ is also nonincreasing, $\psi_{m,M,R}(r)/\sqrt r$ is nonincreasing. Therefore $\psi_{m,M,R}$ is a nontrivial sub-root function satisfying $\psi_{m,M,R}\geq\Phi_R$.

Let $r_{m,M,R}^*$ be its fixed point. Define
\begin{equation*}
    k_R
    :=
    \min\left\{
        \sqrt{\frac{2D_R}{B_Rn}},
        2M_R
    \right\}.
\end{equation*}
Since
\begin{equation*}
    r_{m,M,R}^*
    =
    \psi_{m,M,R}(r_{m,M,R}^*)
    \geq
    \frac{D_R}{n},
\end{equation*}
the monotonicity of $r\mapsto K_r$ gives
\begin{equation*}
    K_{r_{m,M,R}^*}
    \geq
    k_R.
\end{equation*}
We also have
\begin{equation*}
    K_{r_{m,M,R}^*}
    \leq
    \sqrt{\frac{2r_{m,M,R}^*}{B_R}}.
\end{equation*}
It follows that
\begin{equation*}
\begin{aligned}
    r_{m,M,R}^*
    &=
    \psi_{m,M,R}(r_{m,M,R}^*)
    \\
    &\leq
    10B_RQ_Rd
    \sqrt{\frac{W+1}{n}}
    \sqrt{\frac{2r_{m,M,R}^*}{B_R}}
    \sqrt{
        \log\left(\frac{A_{m,R}}{k_R}\right)+1
    }
    +
    \frac{D_R}{n}.
\end{aligned}
\end{equation*}
Young's inequality therefore yields
\begin{equation*}
\begin{aligned}
    r_{m,M,R}^*
    \leq\;&
    \frac12r_{m,M,R}^*
    +
    \frac{
        100B_RQ_R^2d^2(W+1)
    }{n}
    \left[
        \log\left(\frac{A_{m,R}}{k_R}\right)+1
    \right]
    +
    \frac{D_R}{n}.
\end{aligned}
\end{equation*}
Rearranging gives
\begin{equation*}
    r_{m,M,R}^*
    \leq
    \frac{
        200B_RQ_R^2d^2(W+1)
    }{n}
    \left[
        \log\left(\frac{A_{m,R}}{k_R}\right)+1
    \right]
    +
    \frac{2D_R}{n}.
\end{equation*}

It remains to simplify the logarithmic factor. Define
\begin{equation*}
    A_{n,R}
    :=
    \max\left\{
        8M_R,\,
        2M_R+4R_X'(n+M^2)
    \right\}.
\end{equation*}
Since $m\leq n$,
\begin{equation*}
    A_{m,R}\leq A_{n,R}.
\end{equation*}
Moreover,
\begin{equation*}
    D_R
    \geq
    44B_RM_R^2\log(dn).
\end{equation*}
Because $dn\geq2$,
\begin{equation*}
\begin{aligned}
    \sqrt{\frac{2D_R}{B_Rn}}
    &\geq
    M_R\sqrt{\frac{88\log(dn)}{n}}
\geq
    \frac{2M_R}{\sqrt n}.
\end{aligned}
\end{equation*}
Since $2M_R\geq2M_R/\sqrt n$, it follows that
\begin{equation*}
    k_R
    \geq
    \frac{2M_R}{\sqrt n}.
\end{equation*}
Consequently,
\begin{equation*}
\begin{aligned}
    \log\left(\frac{A_{m,R}}{k_R}\right)+1
    &\leq
    1+
    \log\left(
        \frac{\sqrt n\,A_{n,R}}{2M_R}
    \right)
    \\
    &=
    1+
    \log\left[
        \sqrt n
        \max\left\{
            4,\,
            1+
            \frac{
                2\sqrt{1+R_X^2}(n+M^2)
            }{M_R}
        \right\}
    \right]
    \\
    &=
    \Lambda_{n,R}.
\end{aligned}
\end{equation*}

Finally, since $m\geq1$,
\begin{equation*}
    W+1
    =
    m(d+2)+1
    \leq
    m(d+3),
\end{equation*}
and
\begin{equation*}
    \frac{2D_R}{n}
    =
    \frac{
        4M_RB_RQ_R
        +
        88B_RM_R^2\log(dn)
    }{n}.
\end{equation*}
Substituting these estimates into the preceding fixed-point bound gives
\begin{equation*}
    r_{m,M,R}^*
    \leq
    \frac{
        200Q_R^2B_Rd^2(d+3)m
    }{n}
    \Lambda_{n,R}
    +
    \frac{
        4M_RB_RQ_R
        +
        88B_RM_R^2\log(dn)
    }{n},
\end{equation*}
which proves the claim.
\end{proof}

\paragraph{Proof of Theorem~\ref{thm:approx-est tradeoff}:}
\begin{proof}
By \eqref{eq:approximation tail decomposition},
\begin{equation*}
    \varepsilon_{\mathrm{app}}
    \leq
    C_d\Gamma_{\mathrm{app}}(R_X)m^{-\frac{d+3}{d}}
    +
    \tau_{t,R_X}(M).
\end{equation*}
On the other hand, Proposition~\ref{prop:shallow fixed point} gives
\begin{align*}
    \frac{1408}{B_R}r_{m,M,R}^*
    \leq
    281600Q_R^2d^2(d+3)\Lambda_{n,R}
    \frac{m}{n} +
    \frac{
        5632M_RQ_R
        +
        123904M_R^2\log(dn)
    }{n}.
\end{align*}
Substituting these two estimates into Theorem~\ref{thm:local oracle} proves
\eqref{eq:pre balanced rate}.

It remains to balance the two width-dependent terms. Set
\begin{equation*}
    \alpha:=\frac{d+3}{d}.
\end{equation*}
Then
\begin{equation*}
    \frac{1}{\alpha+1}
    =
    \frac{d}{2d+3},
    \qquad
    \frac{\alpha}{\alpha+1}
    =
    \frac{d+3}{2d+3},
\end{equation*}
and the continuous balancing scale is
\begin{equation*}
    \widetilde m
    =
    \left(
        \frac{\Gamma_{\mathrm{app}}(\rho,R_X)n}{\Gamma_{\mathrm{est}}(n,R)}
    \right)^{\frac{1}{\alpha+1}}.
\end{equation*}
At this scale,
\begin{align*}
    \Gamma_{\mathrm{app}}(\rho,R_X)\widetilde m^{-\alpha}
    &=
    \Gamma_{\mathrm{app}}(\rho,R_X)^{\frac{1}{\alpha+1}}
    \left(
        \frac{\Gamma_{\mathrm{est}}(n,R)}{n}
    \right)^{\frac{\alpha}{\alpha+1}},
    \\
    \frac{\Gamma_{\mathrm{est}}(n,R)}{n}\widetilde m
    &=
    \Gamma_{\mathrm{app}}(\rho,R_X)^{\frac{1}{\alpha+1}}
    \left(
        \frac{\Gamma_{\mathrm{est}}(n,R)}{n}
    \right)^{\frac{\alpha}{\alpha+1}}.
\end{align*}
For $m=\lceil\widetilde m\rceil$,
\begin{equation*}
    m^{-\alpha}\leq\widetilde m^{-\alpha},
    \qquad
    m\leq\widetilde m+1.
\end{equation*}
Thus integer rounding contributes at most $\Gamma_{\mathrm{est}}(n,R)/n$, and
\eqref{eq:balanced rate} follows.
\end{proof}

\paragraph{Proof of Proposition~\ref{prop:subG}:}

\begin{proof}
For every $u\in\mathbb S^{d-1}$, the OU representation gives
\begin{equation*}
    \langle X_t,u\rangle
    =
    e^{-t}\langle X_0,u\rangle
    +
    \sigma_t\langle Z,u\rangle.
\end{equation*}
By the triangle inequality for the sub-Gaussian norm,
\begin{align*}
    \|\langle X_t,u\rangle\|_{\psi_2}
    &\leq
    e^{-t}\|\langle X_0,u\rangle\|_{\psi_2}
    +
    \sigma_t\|\langle Z,u\rangle\|_{\psi_2}
    \\
    &\leq
    e^{-t}K_0+\sigma_tK_Z
    \leq
    K_0+K_Z
    =
    K.
\end{align*}
It follows that, for every $a>0$,
\begin{equation*}
    \mathbb P
    \left(
        |\langle X_t,u\rangle|\geq a
    \right)
    \leq
    2\exp\left(-\frac{a^2}{K^2}\right).
\end{equation*}

Let $\mathcal N$ be a $1/2$-net of $\mathbb S^{d-1}$ satisfying
\begin{equation*}
    |\mathcal N|\leq5^d.
\end{equation*}
For every $x\in\mathbb R^d$,
\begin{equation*}
    \|x\|_2
    \leq
    2\max_{v\in\mathcal N}|\langle x,v\rangle|.
\end{equation*}
Therefore, by the union bound,
\begin{align*}
    \mathbb P\bigl(\|X_t\|_2\geq R_X\bigr)
    &\leq
    \mathbb P
    \left(
        \max_{v\in\mathcal N}
        |\langle X_t,v\rangle|
        \geq
        \frac{R_X}{2}
    \right)
    \\
    &\leq
    2\cdot5^d
    \exp\left(-\frac{R_X^2}{4K^2}\right).
\end{align*}
If $R_X\geq K\sqrt{8d\log5}$, then
\begin{equation*}
    5^d
    \leq
    \exp\left(\frac{R_X^2}{8K^2}\right),
\end{equation*}
and hence
\begin{equation*}
    \mathbb P\bigl(\|X_t\|_2\geq R_X\bigr)
    \leq
    2\exp\left(-\frac{R_X^2}{8K^2}\right).
\end{equation*}

We next control the DSM label
\begin{equation*}
    Y=-\frac{Z}{\sigma_t}.
\end{equation*}
Since $z\mapsto\|z\|_2$ is $1$-Lipschitz and
\begin{equation*}
    \mathbb E\|Z\|_2
    \leq
    \sqrt{\mathbb E\|Z\|_2^2}
    =
    \sqrt d,
\end{equation*}
Gaussian concentration yields, for every $a\geq0$,
\begin{equation*}
    \mathbb P
    \left(
        \|Z\|_2-\sqrt d\geq a
    \right)
    \leq
    \exp\left(-\frac{a^2}{2}\right).
\end{equation*}
Consequently,
\begin{align*}
    \mathbb P\bigl(\|Y\|_2\geq R_Y\bigr)
    &=
    \mathbb P\bigl(\|Z\|_2\geq\sigma_tR_Y\bigr)
    \\
    &\leq
    \exp\left(
        -\frac{(\sigma_tR_Y-\sqrt d)^2}{2}
    \right).
\end{align*}
When $R_Y\geq2\sqrt d/\sigma_t$, we have
\begin{equation*}
    \sigma_tR_Y-\sqrt d
    \geq
    \frac{\sigma_tR_Y}{2},
\end{equation*}
so that
\begin{equation*}
    \mathbb P\bigl(\|Y\|_2\geq R_Y\bigr)
    \leq
    \exp\left(-\frac{\sigma_t^2R_Y^2}{8}\right).
\end{equation*}

Finally, since
\begin{equation*}
    \mathcal A_R^c
    \subseteq
    \{\|X_t\|_2>R_X\}
    \cup
    \{\|Y\|_2>R_Y\},
\end{equation*}
another union bound gives
\begin{align*}
    \eta_R
     =
    \mathbb P\bigl((X,Y)\notin\mathcal A_R\bigr)
    &\leq
    \mathbb P\bigl(\|X_t\|_2>R_X\bigr)
    +
    \mathbb P\bigl(\|Y\|_2>R_Y\bigr)
    \\
    &\leq
    2\exp\left(-\frac{R_X^2}{8K^2}\right)
    +
    \exp\left(-\frac{\sigma_t^2R_Y^2}{8}\right).
\end{align*}
\end{proof}

\paragraph{Proof of Propostion~\ref{prop:tail control}}

Before proving the upper bound for the tails. We first present an auxiliary lemma.
\begin{lem}[Fourth moment of the diffused data]
\label{lem:4th moment}
Under Assumption~\ref{asm:subg}, define
\begin{equation*}
     \mu_{t,4}:=
        \left[\mathbb E(1+\|X_t\|_2^2)^2\right]^{1/2}.
\end{equation*} 
Then
\begin{equation*}
    \mu_{t,4}^2
    \leq
    2+64d^2e^{-4t}K_0^4+16d(d+2)\sigma_t^4.
\end{equation*}
In particular,
\begin{equation*}
    \sup_{t>0}\mu_{t,4}^2
    \leq
    2+64d^2K_0^4+16d(d+2)\leq C_d(1+K_0)^4
    <\infty.
\end{equation*}
\end{lem}

\begin{proof}
For every coordinate $i\in[d]$,
\begin{equation*}
    \|X_{0,i}\|_{\psi_2}\leq K_0,
\end{equation*}
and hence
\begin{equation*}
    \mathbb P(|X_{0,i}|>a)
    \leq
    2\exp\left(-\frac{a^2}{K_0^2}\right),
    \qquad a>0.
\end{equation*}
Therefore,
\begin{align*}
    \mathbb E|X_{0,i}|^4
    &=
    \int_0^\infty
    4a^3\mathbb P(|X_{0,i}|>a)\,\mathrm da
    \\
    &\leq
    8K_0^4\int_0^\infty u^3e^{-u^2}\,\mathrm du
    =
    4K_0^4.
\end{align*}
By Cauchy--Schwarz,
\begin{align*}
    \mathbb E\|X_0\|_2^4
    &=
    \sum_{i=1}^d\mathbb E X_{0,i}^4
    +
    2\sum_{1\leq i<j\leq d}
    \mathbb E(X_{0,i}^2X_{0,j}^2)
\leq
    4dK_0^4+4(d^2-d)K_0^4
    =
    4d^2K_0^4.
\end{align*}
Since $X_t=e^{-t}X_0+\sigma_tZ$,
\begin{align*}
    \mathbb E\|X_t\|_2^4
    &\leq
    8e^{-4t}\mathbb E\|X_0\|_2^4
    +
    8\sigma_t^4\mathbb E\|Z\|_2^4
     \leq
    32d^2e^{-4t}K_0^4
    +
    8d(d+2)\sigma_t^4.
\end{align*}
Finally,
\begin{align*}
    \mu_{t,4}^2
    &=
    \mathbb E(1+\|X_t\|_2^2)^2
    \\
    &\leq
    2+2\mathbb E\|X_t\|_2^4
    \\
    &\leq
    2+64d^2e^{-4t}K_0^4
    +16d(d+2)\sigma_t^4.
\end{align*}
The time-uniform assertion follows from $e^{-t}\leq1$ and
$\sigma_t\leq1$.
\end{proof}

\begin{proof}[proof of Proposition~\ref{prop:tail control}]
By Lemma~\ref{lem:localized shallow bound} and Cauchy--Schwarz, for
every $k\geq1$ and $f\in\mathcal F_{k,M}$,
\begin{align*}
    \mathbb E\left[
        \|f(X_t)\|_2^2
        \mathbf 1_{\{\|X_t\|_2>R_X\}}
    \right]
    &\leq
    dM^2
    \mathbb E\left[
        (1+\|X_t\|_2^2)
        \mathbf 1_{\{\|X_t\|_2>R_X\}}
    \right]
    \\
    &\leq
    dM^2\mu_{t,4}
    \mathbb P(\|X_t\|_2>R_X)^{1/2}
    \\
    &\leq
    dM^2\mu_{t,4}\eta_R^{1/2}.
\end{align*}
Moreover, the DSM identity
\begin{equation*}
    s_t(X_t)=\mathbb E[Y\mid X_t]
\end{equation*}
and conditional Jensen's inequality give
\begin{equation*}
    \mathbb E\|s_t(X_t)\|_2^4
    \leq
    \mathbb E\|Y\|_2^4
    =
    \frac{d(d+2)}{\sigma_t^4},
\end{equation*}
where $Y=Y_t\sim\mathcal N(0,\frac{1}{\sigma_t^{2}}I_d)$.

Another application of Cauchy--Schwarz therefore yields
\begin{equation*}
    \mathbb E\left[
        \|s_t(X_t)\|_2^2
        \mathbf 1_{\{\|X_t\|_2>R_X\}}
    \right]
    \leq
    \frac{\sqrt{d(d+2)}}{\sigma_t^2}\eta_R^{1/2}.
\end{equation*}
Using \eqref{eq:approximation tail} and
$\|a-b\|_2^2\leq2\|a\|_2^2+2\|b\|_2^2$ proves
\begin{equation*}
    \tau_{t,R_X}(M)
    \leq
    \left(
        2dM^2\mu_{t,4}
        +
        \frac{2\sqrt{d(d+2)}}{\sigma_t^2}
    \right)\eta_R^{1/2}.
\end{equation*}

We next control $\Delta_R$. For every $f\in\mathcal F_{m,M}$, recall that
\begin{equation*}
     h_f(x,y)
    :=
    \|f(x)-y\|_2^2-\|s_t(x)-y\|_2^2.
\end{equation*}
We have
\begin{align*}
    \left|
        \mathbb E\left[
            h_f(X_t,Y)\mathbf 1_{\mathcal A_R^c}
        \right]
    \right|
    \leq{}&
    2\mathbb E\left[
        \|f(X_t)\|_2^2\mathbf 1_{\mathcal A_R^c}
    \right] +
    2\mathbb E\left[
        \|s_t(X_t)\|_2^2\mathbf 1_{\mathcal A_R^c}
    \right] +
    4\mathbb E\left[
        \|Y\|_2^2\mathbf 1_{\mathcal A_R^c}
    \right].
\end{align*}
The same moment bounds give
\begin{align*}
    \mathbb E\left[
        \|f(X_t)\|_2^2\mathbf 1_{\mathcal A_R^c}
    \right]
    &\leq
    dM^2\mu_{t,4}\eta_R^{1/2},
    \\
    \mathbb E\left[
        \|s_t(X_t)\|_2^2\mathbf 1_{\mathcal A_R^c}
    \right]
    &\leq
    \frac{\sqrt{d(d+2)}}{\sigma_t^2}\eta_R^{1/2},
    \\
    \mathbb E\left[
        \|Y\|_2^2\mathbf 1_{\mathcal A_R^c}
    \right]
    &\leq
    \frac{\sqrt{d(d+2)}}{\sigma_t^2}\eta_R^{1/2}.
\end{align*}
Taking the supremum over $f\in\mathcal F_{m,M}$ proves
\begin{equation*}
    \Delta_R
    \leq
    \left(
        2dM^2\mu_{t,4}
        +
        \frac{6\sqrt{d(d+2)}}{\sigma_t^2}
    \right)\eta_R^{1/2}.
\end{equation*}
Finally, the chosen capacity satisfies
\begin{equation*}
    M^2
    \leq
    C_dR_X^3\overline{\mathcal E}_{q_d,\rho,R_X},
\end{equation*}
and it together with Lemma~\ref{lem:4th moment} gives the last assertion.
\end{proof}

\paragraph{Proof of Theorem~\ref{thm:concrete subgaussian rate}}
\begin{proof}
Use the choices of $M_R$ in \eqref{eq:shallow truncation constants}, and define
\begin{equation*}
    L_{t,R}:
    =M_R+R_Y.
\end{equation*}
Then by definition of the $Q_R,B_R, \text{ and } U_R$ in \eqref{eq:truncation const},
\begin{equation*}
    Q_R\leq22L_{t,R},\qquad
    B_R\leq4L_{t,R}^2,\qquad
    U_R\leq4L_{t,R}^2.
\end{equation*}
The definitions in \eqref{eq:app est coeff} and \eqref{eq:width-dependent remainder} therefore give
\begin{align*}
    \Gamma_{\mathrm{est}}(n,R)
    &\leq C_dL_{t,R}^2\Lambda_{n,R},\\
    \operatorname{Rem}_{n,R}(u)
    &\leq C_d\frac{L_{t,R}^2(1+\log(dn)+u)}{n}.
\end{align*}

Under the capacity choice \eqref{eq:cap choice},
\begin{equation*}
    M=C_{\mathrm{cap},d}R_X^{3/2}
        \overline{\mathcal E}_{q_d,1/2,R_X}^{1/2}.
\end{equation*}
Moreover, Corollary~\ref{cor:uniform pointwise score bound} gives
\begin{equation*}
   \sup_{\|x\|\leq R_X} \|s_t(x)\|_2
    \leq C_dR_X^{q_d-d/2}
        \overline{\mathcal E}_{q_d,1/2,R_X}^{1/2}.
\end{equation*}
Since $q_d-d/2\in\{3/2,2\}$ and $R_X\geq1$, it follows that
\begin{equation*}
    L_{t,R}^2
    \leq C_d\left(
        R_X^5\overline{\mathcal E}_{q_d,1/2,R_X}+R_Y^2
    \right).
\end{equation*}

We now take $R=R_{n,t}$ and $M=M_n$. For brevity, set
\begin{equation*}
    A_n:=P_\infty R_{X,n}^{d+3}E_n,
    \qquad
    D_n:=R_{X,n}^5E_n+R_{Y,n,t}^2.
\end{equation*}
The preceding estimates imply
\begin{align*}
    \Gamma_{\mathrm{app}}(R_{X,n})&=A_n,\\
    \Gamma_{\mathrm{est}}(n,R_{n,t})&\leq C_dD_n\Lambda_n,\\
    \operatorname{Rem}_{n,R_{n,t}}(u)
    &\leq C_d\frac{D_n(1+\log(dn)+u)}{n}.
\end{align*}

Let $\alpha_d:=(d+3)/d$. By the definition of the balancing width,
\begin{equation*}
    \widetilde m_n=
    \left(\frac{A_nn}{D_n\Lambda_n}\right)^{\gamma_d},
    \qquad
    \widetilde m_n\leq m_n\leq\widetilde m_n+1.
\end{equation*}
Using $\alpha_d\gamma_d=\beta_d$ and $\gamma_d+\beta_d=1$, we obtain
\begin{align*}
    A_nm_n^{-\alpha_d}
    &\leq A_n^{\gamma_d}
        \left(\frac{D_n\Lambda_n}{n}\right)^{\beta_d},\\
    D_n\Lambda_n\frac{m_n}{n}
    &\leq A_n^{\gamma_d}
        \left(\frac{D_n\Lambda_n}{n}\right)^{\beta_d}
        +\frac{D_n\Lambda_n}{n}.
\end{align*}

Set $u:=\log(4/\delta)$. Proposition~\ref{prop:subG} and the prescribed radii give
\begin{equation*}
    \eta_{R_{n,t}}
    \leq\frac{\delta}{2n^2}+\frac{\delta}{2n^2}
    =\frac{\delta}{n^2}.
\end{equation*}
Since $n\geq2$,
\begin{equation*}
    2e^{-u}+n\eta_{R_{n,t}}
    \leq\frac{\delta}{2}+\frac{\delta}{n}
    \leq\delta.
\end{equation*}
Applying \eqref{eq:pre balanced rate} with $R=R_{n,t}$ and $m=m_n\leq n$, we therefore obtain, with probability at least $1-\delta$,
\begin{align*}
    \mathcal R_t(\widehat f_{m_n})
    \leq{}& C_dA_n^{\gamma_d}
        \left(\frac{D_n\Lambda_n}{n}\right)^{\beta_d}
        +4\tau_{t,R_{X,n}}(M_n)+8\Delta_{R_{n,t}} +C_d\frac{D_n(\Lambda_n+1+\log(dn)+u)}{n}.
\end{align*}
Here $\Delta_{R_{n,t}}$ is evaluated over $\mathcal F_{m_n,M_n}$.

It remains to control the tail terms. Lemma~\ref{lem:4th moment} gives $\mu_{t,4}\leq C_dS_0^2$.
The definitions of $K$ and $R_{X,n}$ imply $S_0\lesssim R_{X,n}$, while $R_{Y,n,t}^2\geq4d\sigma_t^{-2}$.
Consequently,
\begin{equation*}
    R_{X,n}^3E_n\mu_{t,4}+\sigma_t^{-2}
    \leq C_d\left(R_{X,n}^5E_n+R_{Y,n,t}^2\right)
    =C_dD_n.
\end{equation*}
By Proposition~\ref{prop:tail control},
\begin{align*}
    4\tau_{t,R_{X,n}}(M_n)+8\Delta_{R_{n,t}}
    &\leq C_d\left(
        R_{X,n}^3E_n\mu_{t,4}+\sigma_t^{-2}
    \right)\eta_{R_{n,t}}^{1/2} 
    \leq C_dD_n\frac{\sqrt\delta}{n}
    \leq C_d\frac{D_n}{n}.
\end{align*}
Finally, $1+\log(dn)+u\leq C_d\ell_{n,\delta}$, so on the
same event,
\begin{equation*}
    \mathcal R_t(\widehat f_{m_n})
    \lesssim_d
    A_n^{\gamma_d}\left(\frac{D_n\Lambda_n}{n}\right)^{\beta_d}
    +\frac{D_n(\Lambda_n+\ell_{n,\delta})}{n}.
\end{equation*}
Substituting the definitions of $A_n$ and $D_n$ proves
\eqref{eq:compact subgaussian rate}.
\end{proof}

\paragraph{Proof of Corollary~\ref{cor:explicit distributional rate}}
\begin{proof}
We first verify the radius bound in \eqref{eq:explicit radius envelope}.
Under the stated convention for the sub-Gaussian norm, $K_Z=\sqrt{8/3}$, so $K=K_0+K_Z\leq K_ZS_0$.
Since $d\geq1$ and $\ell_{n,\delta}\geq1$, the definition of $R_{X,n}$ gives
\begin{align*}
    R_{X,n}
    &\leq1\vee\left(
        \sqrt8\,K_ZS_0\sqrt{d\log5\,\ell_{n,\delta}}
    \right)\\
    &\leq6\sqrt d\,S_0\sqrt{\ell_{n,\delta}}
    =L_{n,\delta},
\end{align*}
where we used $\sqrt8\,K_Z\sqrt{\log5}<6$.
Also,
\begin{equation*}
    R_{Y,n,t}^2
    =\sigma_t^{-2}\max\left\{4d,8\log\frac{2n^2}{\delta}\right\}
    \leq C_d\sigma_t^{-2}\ell_{n,\delta}.
\end{equation*}
By \eqref{eq:regularity complexity explicit},
\begin{equation*}
    E_n
    \leq\mathcal C_{\mathrm{reg}}(H,\kappa)
        R_{X,n}^{e_{d,\kappa}}
    \leq\mathcal C_{\mathrm{reg}}(H,\kappa)
        L_{n,\delta}^{e_{d,\kappa}}.
\end{equation*}

As in the preceding proof, write
\begin{equation*}
    A_n:=P_\infty R_{X,n}^{d+3}E_n,
    \qquad
    D_n:=R_{X,n}^5E_n+R_{Y,n,t}^2.
\end{equation*}
The preceding estimates yield
\begin{align*}
    A_n
    &\leq P_\infty\mathcal C_{\mathrm{reg}}(H,\kappa)
        L_{n,\delta}^{e_{d,\kappa}+d+3},\\
    D_n
    &\leq\mathcal C_{\mathrm{reg}}(H,\kappa)
        L_{n,\delta}^{e_{d,\kappa}+5}
        +C_d\sigma_t^{-2}\ell_{n,\delta}.
\end{align*}
Since $0<\beta_d<1$, we have $(a+b)^{\beta_d}\leq a^{\beta_d}+b^{\beta_d}$ for $a,b\geq0$.
It follows that
\begin{align*}
    &A_n^{\gamma_d}
        \left(\frac{D_n\Lambda_n}{n}\right)^{\beta_d}\lesssim_d
        P_\infty^{\gamma_d}
        \left(\frac{\Lambda_n}{n}\right)^{\beta_d}
        \Bigl[
            \mathcal C_{\mathrm{reg}}(H,\kappa)
            L_{n,\delta}^{e_{d,\kappa}+c_d} 
            +\mathcal C_{\mathrm{reg}}(H,\kappa)^{\gamma_d}
            L_{n,\delta}^{\gamma_d(e_{d,\kappa}+d+3)}
            (\sigma_t^{-2}\ell_{n,\delta})^{\beta_d}
        \Bigr],
\end{align*}
where we used $\gamma_d+\beta_d=1$ and
\begin{equation*}
    \gamma_d(e_{d,\kappa}+d+3)
    +\beta_d(e_{d,\kappa}+5)
    =e_{d,\kappa}+c_d.
\end{equation*}
Similarly,
\begin{align*}
    &\frac{D_n(\Lambda_n+\ell_{n,\delta})}{n}\lesssim_d
        \frac{\Lambda_n+\ell_{n,\delta}}{n}
        \left[
            \mathcal C_{\mathrm{reg}}(H,\kappa)
            L_{n,\delta}^{e_{d,\kappa}+5}
            +\sigma_t^{-2}\ell_{n,\delta}
        \right].
\end{align*}
Substituting these two bounds into
\eqref{eq:compact subgaussian rate} gives
\eqref{eq:concrete subgaussian rate} on the same event of
probability at least $1-\delta$.

\end{proof}

\section{Proofs in Section~\ref{sec:sampling}}\label{appendix:sampling}
\paragraph{Proof of Proposition~\ref{prop:accumulated score error}:}
\begin{proof}
Throughout the proof, all quantities in Corollary~\ref{cor:explicit distributional rate} are evaluated with the confidence parameter $\delta/N$. In particular,
\begin{equation*}
    \ell_{n,\delta/N}
    =
    1+\log\frac{4n^2N}{\delta}\asymp 1+\log\frac{nN}{\delta}.
\end{equation*}

For each $k=0,\ldots,N-1$, let $\mathcal E_k$ denote the event on which Corollary~\ref{cor:explicit distributional rate}, applied at time $t=\tau_k$ with confidence parameter $\delta/N$, holds. Since $m_{n,k}\leq n$, the theorem gives
\begin{equation*}
    \mathbb P(\mathcal E_k)
    \geq
    1-\frac{\delta}{N}.
\end{equation*}
Therefore, by the union bound,
\begin{equation*}
    \mathbb P\left(
        \bigcap_{k=0}^{N-1}\mathcal E_k
    \right)
    \geq
    1-\sum_{k=0}^{N-1}\frac{\delta}{N}
    =
    1-\delta.
\end{equation*}
No independence among the estimators at different time points is
required for this argument.

We work on the event
\begin{equation*}
    \mathcal E
    :=
    \bigcap_{k=0}^{N-1}\mathcal E_k.
\end{equation*}

By Remark~\ref{rmk:Lambda n log term} and the parameters choices in Theorem~\ref{thm:concrete subgaussian rate}, we have 
\begin{equation*}
\Lambda_{n}\lesssim_{d,p_0}1+\log(nN/\delta).
\end{equation*}

Further, from Corollary~\ref{cor:explicit distributional rate}, there exist
constants $C>0$ and $A\geq0$, depending only on $d$ and the initial distribution $p_0$, such that
on $\mathcal E$,
\begin{equation*}
    \mathcal R_{\tau_k}(\widehat s_{\tau_k})
    \leq
    C\ell_{n,\delta/N}^{A}
    \left(
        \frac{\sigma_{\tau_k}^{-2\beta_d}}{n^{\beta_d}}
        +\frac{\sigma_{\tau_k}^{-2}}{n}
    \right),
    \qquad k=0,\ldots,N-1.
\end{equation*}

Recall that
\begin{equation*}
    h_k
    =
    \tau_k-\tau_{k+1},
    \qquad
    \tau_0=T,
    \qquad
    \tau_N=\gamma.
\end{equation*}
Consequently,
\begin{equation*}
    \sum_{k=0}^{N-1}h_k
    =
    T-\gamma.
\end{equation*}

For any $a>0$, the function $u\mapsto\sigma_u^{-2a}$ is decreasing, and hence
\begin{align*} 
    \sum_{k=0}^{N-1}
    h_k\sigma_{\tau_k}^{-2a}
    &\leq
    \int_\gamma^T
    \sigma_u^{-2a}\,\mathrm du
    :=
    J_a(\gamma,T).
\end{align*}
Since $0<\beta_d<1$ and $T\geq1$, we obtain
\begin{equation*}
    \sum_{k=0}^{N-1}
    h_k\sigma_{\tau_k}^{-2\beta_d}
    \leq
    J_{\beta_d}(\gamma,T)\lesssim T
\end{equation*}
and
\begin{equation*}
    \sum_{k=0}^{N-1}
    h_k\sigma_{\tau_k}^{-2}
    \leq
    J_1(\gamma,T)\lesssim T+\log (\gamma^{-1}).
\end{equation*}

Finally, by the definition of the accumulated score error,
\begin{align*}
    \varepsilon_{\mathrm{score}}^2
    &=
    \sum_{k=0}^{N-1}
    h_k
    \mathbb E_{X_{\tau_k}\sim p_{\tau_k}}
    \left[
        \left\|
            \widehat s_{\tau_k}(X_{\tau_k})
            -
            s_{\tau_k}(X_{\tau_k})
        \right\|_2^2
    \right]=
    \sum_{k=0}^{N-1}
    h_k
    \mathcal R_{\tau_k}
    \bigl(\widehat s_{\tau_k}\bigr).
\end{align*}
Multiplying the preceding pointwise risk bound by $h_k$, summing over
$k$, and applying the preceding integral bounds above gives
\begin{align*}
    \varepsilon_{\mathrm{score}}^2
    &= \widetilde O_{d,p_0}\!\left(
        \frac{T}{n^{\beta_d}}
        +\frac{T+\log(\gamma^{-1})}{n}
        \right).
\end{align*}
\end{proof}

\paragraph{Proof of Corollary~\ref{cor:sampling consequence}:}
\begin{proof}
By Proposition~\ref{prop:accumulated score error},
with probability $1-\delta$, we have 
\begin{equation*}
    \varepsilon_{\mathrm{score}}^2=\widetilde{O}_{d,P_0}\left(\frac{T}{n^{\beta_d}}+\frac{T+\log(\gamma^{-1})}{n}\right)
\end{equation*}

Applying Lemma~\ref{lem:score to sampling} gives
\begin{align*}
    \operatorname{KL}
    \left(
        p_\gamma
        \,\middle\|\,
        \widehat p_{\mathrm{out}}
    \right)
    \lesssim
    \varepsilon_{\mathrm{score}}^2
    +
    \overline h_N^2 dN
    +
    \overline h_N(dT+M_2)
    +
    (d+M_2)e^{-2T},
\end{align*}
where $\overline h_N$ is the mesh parameter associated with the
reverse-time grid.

By the grid construction in
Lemma~\ref{lem:score to sampling},
\begin{equation*}
    \overline h_N
    \lesssim
    \frac{
        T+\log(\gamma^{-1})
    }{N}.
\end{equation*}
Consequently, combining these bounds with the estimate for
$\varepsilon_{\mathrm{score}}^2$ gives
\begin{align*}
    \operatorname{KL}
    \left(
        p_\gamma
        \,\middle\|\,
        \widehat p_{\mathrm{out}}
    \right)=\widetilde O_{d,P_0}\!\left(
\frac{ T}{n^{\beta_d}}
+{\frac{T+\log(\gamma^{-1})}{n}}
+\left(\frac{T+\log(\gamma^{-1})}{\sqrt N}\right)^2
+e^{-2T}
\right)
\end{align*}
with probability $1-\delta$.
\end{proof}

\paragraph{Proof of Proposition~\ref{prop: early stopping error}:}
\begin{proof}
Throughout the proof, all $L^1$-norms are taken over $\RR^d$.
Since $r\geq1$, Assumption~\ref{asm:p0 reg} gives
$p_0\in W^{2,1}(\RR^d)$.
By Assumption~\ref{asm:poly growth}, for every $i\in[d]$,
\begin{align*}
    |\partial_i p_0(x)|
    &\leq H(1+\|x\|_2)^\kappa p_0(x),\\
    |\partial_{ii}p_0(x)|
    &\leq H^2(1+\|x\|_2)^{2\kappa}p_0(x)
\end{align*}
for almost every $x\in\RR^d$.
Consequently,
\begin{equation*}
    \|\nabla p_0(x)\|_2
    =
    \left(
        \sum_{i=1}^d|\partial_i p_0(x)|^2
    \right)^{1/2}
    \leq
    \sqrt d\,H(1+\|x\|_2)^\kappa p_0(x),
\end{equation*}
and
\begin{equation*}
    \|\Delta p_0\|_{L^1}
    \leq
    \sum_{i=1}^d\|\partial_{ii}p_0\|_{L^1}
    \leq
    dH^2\mathbb E(1+\|X_0\|_2)^{2\kappa}.
\end{equation*}

We next bound the moments appearing above.
By Assumption~\ref{asm:subg} and convexity of the exponential,
\begin{equation*}
    \mathbb E\exp\left(
        \frac{\|X_0\|_2^2}{dK_0^2}
    \right)
    \leq
    \frac1d\sum_{i=1}^d
    \mathbb E\exp\left(
        \frac{X_{0,i}^2}{K_0^2}
    \right)
    \leq2.
\end{equation*}
Markov's inequality therefore gives
\begin{equation*}
    \mathbb P(\|X_0\|_2>u)
    \leq
    2\exp\left(-\frac{u^2}{dK_0^2}\right),
    \qquad u>0.
\end{equation*}
For every $q>0$, integration of this tail bound yields
\begin{align*}
    \mathbb E\|X_0\|_2^q
    &=
    \int_0^\infty
        q u^{q-1}\mathbb P(\|X_0\|_2>u)\,\mathrm du\leq
    2\Gamma\left(1+\frac q2\right)d^{q/2}K_0^q.
\end{align*}
Using $(1+u)^q\leq2^{\max\{q-1,0\}}(1+u^q)$, we obtain, for every $q\geq0$,
\begin{equation*}
    \mathbb E(1+\|X_0\|_2)^q
    \leq
    2^{\max\{q-1,0\}}
    \left[
        1+
        2\Gamma\left(1+\frac q2\right)d^{q/2}K_0^q
    \right],
\end{equation*}
where the case $q=0$ is immediate.
In particular,
\begin{align*}
    \int_{\RR^d}
        \|x\|_2\,\|\nabla p_0(x)\|_2\,\mathrm dx
    &\leq
    \sqrt d\,H\,
    \mathbb E\left[
        \|X_0\|_2(1+\|X_0\|_2)^\kappa
    \right]\\
    &\leq
    \sqrt d\,H\,
    \mathbb E(1+\|X_0\|_2)^{\kappa+1}
    <\infty.
\end{align*}

Then we have
\begin{equation*}
    \nabla\cdot(xp_0)
    =
    \sum_{i=1}^d\partial_i(x_i p_0)
    =
    dp_0+x\cdot\nabla p_0.
\end{equation*}
Since $\int_{\RR^d}p_0(x)\,\mathrm dx=1$, it follows that
\begin{align*}
    \|\Delta p_0+\nabla\cdot(xp_0)\|_{L^1}
    &\leq
    \|\Delta p_0\|_{L^1}
    +d
    +\int_{\RR^d}
        \|x\|_2\,\|\nabla p_0(x)\|_2\,\mathrm dx\\
    &\leq
    d+dH^2\mathbb E(1+\|X_0\|_2)^{2\kappa}
    +\sqrt d\,H\,
        \mathbb E(1+\|X_0\|_2)^{\kappa+1}\\
    &\leq
    d
    +2^{\max\{2\kappa-1,0\}}dH^2
    \left[
        1+2\Gamma(1+\kappa)d^\kappa K_0^{2\kappa}
    \right]\\
    &\quad
    +2^\kappa\sqrt d\,H
    \left[
        1+
        2\Gamma\left(\frac{\kappa+3}{2}\right)
        d^{(\kappa+1)/2}K_0^{\kappa+1}
    \right]
    <\infty.
\end{align*}

For $t>0$, the Ornstein--Uhlenbeck transition density is
\begin{equation*}
    k_t(x,y)
    =
    \frac{1}{(2\pi\sigma_t^2)^{d/2}}
    \exp\left(
        -\frac{\|x-e^{-t}y\|_2^2}{2\sigma_t^2}
    \right),
    \qquad
    \sigma_t^2=1-e^{-2t}.
\end{equation*}

By the Kolmogorov backward equation \cite[Section~1.7]{kloeden1992numerical},  the Ornstein--Uhlenbeck transition density satisfies
\begin{equation*}
    \partial_t k_t(x,y)
    =
    \Delta_y k_t(x,y)
    -
    y\cdot\nabla_y k_t(x,y).
\end{equation*}
Differentiating
$p_t(x)=\int_{\RR^d}k_t(x,y)p_0(y)\,\mathrm dy$
and integrating by parts in the weak sense yield
\begin{align*}
    \partial_t p_t(x)
    &=
    \int_{\RR^d}
        \partial_t k_t(x,y)p_0(y)\,\mathrm dy\\
    &=
    \int_{\RR^d}
        k_t(x,y)
        \left[
            \Delta p_0(y)
            +\nabla_y\cdot\bigl(yp_0(y)\bigr)
        \right]
        \,\mathrm dy.
\end{align*}
These operations are justified by the integrability established
above and the Gaussian decay of the kernel and its derivatives.

Since $k_t(x,y)\geq0$ and
$\int_{\RR^d}k_t(x,y)\,\mathrm dx=1$,
Tonelli's theorem gives
\begin{align*}
    \|\partial_t p_t\|_{L^1}
    &\leq
    \int_{\RR^d}\int_{\RR^d}
        k_t(x,y)
        \left|
            \Delta p_0(y)
            +\nabla_y\cdot\bigl(yp_0(y)\bigr)
        \right|
        \,\mathrm dy\,\mathrm dx\\
    &=
    \|\Delta p_0+\nabla\cdot(xp_0)\|_{L^1}.
\end{align*}
Hence, 
\begin{align*}
   \mathrm{TV}(p_0,p_\gamma)
    = \frac12\|p_\gamma-p_0\|_{L^1}
    &\leq\frac12
    \int_0^\gamma
        \|\partial_t p_t\|_{L^1}\,\mathrm dt\leq
    \frac\gamma2
    \|\Delta p_0+\nabla\cdot(xp_0)\|_{L^1}.
\end{align*}

\end{proof}

\paragraph{Proof of Corollary~\ref{cor:endtoend TV bound}:}
\begin{proof}

The triangle inequality and Pinsker's inequality give that
\begin{equation*}
    \mathrm{TV}(p_0,\widehat p_{\mathrm{out}})
    \leq \mathrm{TV}(p_0,p_\gamma)
    +\sqrt{\frac12\operatorname{KL}\!\left(
        p_\gamma\,\middle\|\,\widehat p_{\mathrm{out}}
    \right)}.
\end{equation*} 
Substituting the bounds from Proposition~\ref{prop: early stopping error} and Corollary~\ref{cor:sampling consequence}, and using the subadditivity of the square root, proves the first assertion.
For the prescribed choices of $\gamma,T,N$ and fixed $\delta$, both $T+\log(\gamma^{-1})$ and $\log(nN/\delta)$ are $O(\log n)$.
Since $0<\beta_d<1$, every term in the resulting TV bound is $\widetilde O(n^{-\beta_d/2})$, which proves the second assertion.
\end{proof}

\end{document}